\documentclass[11pt,a4paper]{article}
\usepackage[T1]{fontenc}
\usepackage[utf8]{inputenc}
\usepackage[a4paper,margin=22mm]{geometry}
\usepackage{lmodern,microtype}
\usepackage[numbers,super,sort&compress]{natbib}
\usepackage{graphicx,adjustbox,booktabs,multirow,array,tabularx}
\newcolumntype{Y}{>{\raggedright\arraybackslash}X}
\usepackage{xcolor,amsmath,amssymb,subcaption}
\usepackage{placeins}
\newwrite\displaypositions
\AtBeginDocument{\immediate\openout\displaypositions=\jobname.displays}
\newcommand{\DisplayMark}[1]{\pdfsavepos\write\displaypositions{#1|\thepage|\the\pdflastxpos|\the\pdflastypos}}
\usepackage{xr-hyper}
\usepackage[hidelinks]{hyperref}
\usepackage{fontawesome5}
\newcommand{\GitHubMark}{\raisebox{-0.15ex}{\faGithub}\,}
\newcommand{\HFMark}{\raisebox{-0.35ex}{\includegraphics[height=1.15em]{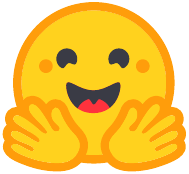}}\,}
\usepackage{xcolor}
\definecolor{resourcelink}{RGB}{0,84,147}
\newcommand{\ResourceLinks}{\begin{center}\small
\href{https://github.com/mahmoudibrahim98/CompDiff}{\GitHubMark\textcolor{resourcelink}{Code}}\qquad
\href{https://huggingface.co/collections/mahmoudibra98/compdiff-fair-intersectional-medical-image-generation-6a54b3ce9a66aa70a8ad534d}{\HFMark\textcolor{resourcelink}{Models}}\qquad
\href{https://huggingface.co/datasets/mahmoudibra98/compdiff-chest-xray-synthetic}{\HFMark\textcolor{resourcelink}{Dataset}}
\end{center}}
\usepackage{caption}
\newcommand{\AppendixName}{Supplementary Note}
\newcommand{\ManuscriptTitle}{CompDiff enables fair and zero shot medical image generation across demographic intersections through compositional diffusion}

\newcommand{\chestBaseAUROC}{0.704}

\newcommand{\chestBaseAge}{6.50}

\newcommand{\chestBaseBioViL}{0.287}

\newcommand{\chestBaseESFIDIntRN}{14.2}

\newcommand{\chestBaseFIDRN}{8.62}
\newcommand{\chestBaseFIDSD}{4.7}
\newcommand{\chestBaseFID}{88.4}

\newcommand{\chestBaseRace}{0.980}

\newcommand{\chestCDAUROC}{0.754}

\newcommand{\chestCDAge}{8.66}

\newcommand{\chestCDBioViL}{0.326}

\newcommand{\chestCDESFIDIntRN}{12.1}

\newcommand{\chestCDFIDRN}{6.64}
\newcommand{\chestCDFIDSD}{5.7}
\newcommand{\chestCDFID}{74.7}

\newcommand{\chestCDRace}{0.949}

\newcommand{\chestFairDAUROC}{0.720}

\newcommand{\chestFairDAge}{8.04}

\newcommand{\chestFairDBioViL}{0.334}

\newcommand{\chestFairDESFIDIntRN}{14.3}

\newcommand{\chestFairDFIDRN}{8.85}
\newcommand{\chestFairDFIDSD}{1.7}
\newcommand{\chestFairDFID}{85.6}

\newcommand{\chestFairDRace}{0.974}

\newcommand{\fundusBaseAUROC}{0.937}
\newcommand{\fundusBaseCDR}{0.957}

\newcommand{\fundusBaseFIDSD}{3.7}
\newcommand{\fundusBaseFID}{72.7}
\newcommand{\fundusBaseGlaucoma}{0.916}

\newcommand{\fundusCDAUROC}{0.975}
\newcommand{\fundusCDCDR}{0.994}

\newcommand{\fundusCDFIDRN}{5.76}
\newcommand{\fundusCDFIDSD}{4.3}
\newcommand{\fundusCDFID}{60.1}
\newcommand{\fundusCDGlaucoma}{0.957}

\newcommand{\fundusFairDAUROC}{0.917}
\newcommand{\fundusFairDCDR}{0.904}

\newcommand{\fundusFairDFIDRN}{5.51}
\newcommand{\fundusFairDFIDSD}{1.8}
\newcommand{\fundusFairDFID}{64.2}
\newcommand{\fundusFairDGlaucoma}{0.930}

\newcommand{\ckptN}{24}
\newcommand{\ckptFloorSex}{0.98}
\newcommand{\ckptFloorRace}{0.92}
\newcommand{\ckptFloorAge}{10}

\newcommand{\ablParamDeltaPct}{0.069}
\newcommand{\ablComposerParamsHier}{2{,}896{,}640}
\newcommand{\ablComposerParamsFlat}{2{,}898{,}632}
\newcommand{\ablHierFloorLo}{7{,}500}
\newcommand{\ablHierFloorHi}{10{,}000}
\newcommand{\ablFlatFloorStep}{11{,}000}

\newcommand{\robSizeAnchor}{600}                  
\newcommand{\robSizeCombos}{32}                   

\newcommand{\robValN}{150}                        

\newcommand{\robCells}{42}                        
\newcommand{\robHOCells}{12}                      
\newcommand{\robWinnerInfJ}{41}                   
\newcommand{\robWinnerKID}{38}                    
\newcommand{\robHOOrderInfJ}{12}                  
\newcommand{\robHOOrderKID}{10}                   

\newcommand{\robSelBudget}{4\times}               
\newcommand{\robSelCondWins}{12}                  
\newcommand{\robSelFloor}{120}                    
\newcommand{\robSelDropped}{4}                    

\newcommand{\figbWinsTotal}{28}

\newcommand{\rsNInstances}{96}
\newcommand{\rsNSets}{92}
\newcommand{\rsNOverlap}{4}
\newcommand{\rsNRatings}{288}

\newcommand{\rsNCompDiff}{91}
\newcommand{\rsNFairD}{95}
\newcommand{\rsNBase}{90}
\newcommand{\rsNReal}{12}
\newcommand{\rsReplacedBase}{6}
\newcommand{\rsReplacedCompDiff}{5}
\newcommand{\rsReplacedFairD}{1}

\newcommand{\rsNNotAssessable}{10}
\newcommand{\rsRealismN}{287}
\newcommand{\rsCorrN}{277}

\newcommand{\rsTOSTMargin}{0.5}

\newcommand{\rsTOSTN}{89}

\newcommand{\rsFamilySize}{8}
\newcommand{\rsNPerm}{10{,}000}
\newcommand{\rsFCCompDiff}{48.4\%}
\newcommand{\rsFCFairD}{36.8\%}
\newcommand{\rsFCBase}{6.7\%}
\newcommand{\rsFCRealWins}{11}
\newcommand{\rsFCRealN}{12}

\newcommand{\readerCorrBaseBH}{0.005}
\newcommand{\readerCorrFairBH}{0.031}

\newcommand{\readerCorrSets}{89}

\newcommand{\numIntroductionA}{47}

\newcommand{\numIntroductionB}{92}

\newcommand{\numIntroductionC}{26}

\newcommand{\numIntroductionD}{48}

\newcommand{\numIntroductionE}{80}

\newcommand{\numIntroductionF}{29}

\newcommand{\numResultsA}{15}

\newcommand{\numResultsB}{17}

\newcommand{\numResultsC}{0.25}

\newcommand{\numResultsD}{0.75}

\newcommand{\numResultsF}{0.11}

\newcommand{\numResultsG}{0.15}

\newcommand{\numResultsK}{0.76}

\newcommand{\numResultsL}{29}

\newcommand{\numResultsQ}{1.1}

\newcommand{\numResultsV}{16}

\newcommand{\numResultsW}{14.9}

\newcommand{\numResultsX}{18.1}

\newcommand{\numResultsY}{0.0001}

\newcommand{\numResultsAB}{80}

\newcommand{\numResultsAC}{2.36}

\newcommand{\numResultsAD}{1}

\newcommand{\numResultsAE}{5}

\newcommand{\numResultsAF}{4.17}

\newcommand{\numResultsAG}{4.13}

\newcommand{\numResultsAH}{0.47}

\newcommand{\numResultsAI}{0.01}

\newcommand{\numResultsAJ}{0.05}

\newcommand{\numResultsAK}{44}

\newcommand{\numResultsAL}{35}

\newcommand{\numResultsAM}{0.37}

\newcommand{\numResultsAN}{4.92}

\newcommand{\numResultsAO}{0.72}

\newcommand{\numResultsAP}{0.69}

\newcommand{\numResultsAQ}{0.68}

\newcommand{\numResultsAU}{0.40}

\newcommand{\numResultsAV}{0.46}

\newcommand{\numResultsAW}{0.44}

\newcommand{\numResultsBA}{3}

\newcommand{\numResultsBB}{4}

\newcommand{\numResultsBE}{0.66}

\newcommand{\numResultsBF}{0.61}

\newcommand{\numResultsBG}{0.28}

\newcommand{\numResultsBH}{0.39}

\newcommand{\numResultsBI}{0.47}

\newcommand{\numResultsCT}{121}

\newcommand{\numResultsCY}{0.097}

\newcommand{\numResultsCZ}{0.235}

\newcommand{\numResultsDA}{0.284}

\newcommand{\numResultsDD}{0.010}

\newcommand{\numResultsDE}{0.059}

\newcommand{\numResultsDF}{0.066}

\newcommand{\numResultsDG}{0.027}

\newcommand{\numResultsDH}{0.077}

\newcommand{\numResultsDI}{0.085}

\newcommand{\numResultsDJ}{14}

\newcommand{\numResultsDK}{0.027}

\newcommand{\numResultsDL}{0.069}

\newcommand{\numResultsDM}{0.083}

\newcommand{\numResultsDN}{0.075}

\newcommand{\numResultsDO}{0.128}

\newcommand{\numResultsDP}{0.135}

\newcommand{\numResultsDQ}{0.077}

\newcommand{\numResultsDR}{0.077}

\newcommand{\numResultsDS}{27}

\newcommand{\numResultsDT}{9}

\newcommand{\numResultsDU}{0.023}

\newcommand{\numResultsDV}{0.030}

\newcommand{\numResultsDW}{0.037}

\newcommand{\numResultsDX}{0.074}

\newcommand{\numResultsDY}{69}

\newcommand{\numResultsDZ}{6}

\newcommand{\numResultsEA}{47}

\newcommand{\numResultsEB}{9}

\newcommand{\numResultsEC}{0.010}

\newcommand{\numResultsED}{0.038}

\newcommand{\numResultsEE}{0.016}

\newcommand{\numResultsEF}{0.028}

\newcommand{\numResultsEG}{1000}

\newcommand{\numResultsEH}{95}

\newcommand{\numResultsEI}{0.167}

\newcommand{\numResultsEJ}{0.110}

\newcommand{\numResultsEK}{0.122}

\newcommand{\numResultsEL}{0.127}

\newcommand{\numResultsEN}{0.095}

\newcommand{\numResultsEO}{60}

\newcommand{\numResultsEP}{80}

\newcommand{\numResultsEQ}{211}

\newcommand{\numResultsER}{51}

\newcommand{\numResultsES}{0.878}

\newcommand{\numResultsET}{0.82}

\newcommand{\numResultsEU}{0.93}

\newcommand{\numResultsEV}{0.783}

\newcommand{\numResultsEW}{0.777}

\newcommand{\numResultsEX}{0.74}

\newcommand{\numResultsEY}{0.82}

\newcommand{\numResultsFA}{0.99}

\newcommand{\numResultsFB}{0.99}

\newcommand{\numResultsFC}{94.5}

\newcommand{\numResultsFD}{70.9}

\newcommand{\numResultsFE}{0.52}

\newcommand{\numResultsFF}{0.02}

\newcommand{\numResultsFG}{77}

\newcommand{\numResultsFH}{140.0}

\newcommand{\numResultsFI}{0.75}

\newcommand{\numResultsFJ}{72.0}

\newcommand{\numResultsFK}{71.6}

\newcommand{\numResultsFL}{0.51}

\newcommand{\numResultsFM}{0.51}

\newcommand{\numResultsFN}{0.54}

\newcommand{\numResultsFO}{12.41}

\newcommand{\numResultsFP}{0.90}

\newcommand{\numResultsFQ}{0.50}

\newcommand{\numResultsFR}{0.87}

\newcommand{\numResultsFS}{15.31}

\newcommand{\numResultsFZ}{11.33}

\newcommand{\numResultsGA}{8.76}

\newcommand{\numResultsGE}{77.6}

\newcommand{\numResultsGF}{10.1}

\newcommand{\numResultsGH}{88.0}

\newcommand{\numResultsGJ}{97.1}

\newcommand{\numResultsGK}{0.1}

\newcommand{\numMethodsC}{4}

\newcommand{\numMethodsD}{3}

\newcommand{\numMethodsE}{2}

\newcommand{\numMethodsF}{4}

\newcommand{\numMethodsI}{2}

\newcommand{\numMethodsJ}{0}

\newcommand{\numMethodsK}{2}

\newcommand{\numMethodsL}{1}

\newcommand{\numMethodsM}{10}

\newcommand{\numMethodsO}{2}

\newcommand{\numMethodsR}{1}

\newcommand{\numMethodsS}{0}

\newcommand{\numMethodsT}{1}

\newcommand{\numMethodsU}{18}

\newcommand{\numMethodsV}{40}

\newcommand{\numMethodsW}{40}

\newcommand{\numMethodsX}{60}

\newcommand{\numMethodsY}{60}

\newcommand{\numMethodsZ}{80}

\newcommand{\numMethodsAA}{80}

\newcommand{\numMethodsAE}{32}

\newcommand{\numMethodsAF}{6.32}

\newcommand{\numMethodsAG}{1.21}

\newcommand{\numMethodsAH}{0.52}

\newcommand{\numMethodsAI}{2.13}

\newcommand{\numMethodsAJ}{0.35}

\newcommand{\numMethodsAK}{256}

\newcommand{\numMethodsAL}{1024}

\newcommand{\numMethodsAM}{128}

\newcommand{\numMethodsAN}{0}

\newcommand{\numMethodsAO}{100}

\newcommand{\numMethodsAP}{0.1}

\newcommand{\numMethodsAQ}{5}

\newcommand{\numMethodsAR}{1}

\newcommand{\numMethodsAS}{512}

\newcommand{\numMethodsAT}{1}

\newcommand{\numMethodsAU}{0.005}

\newcommand{\numMethodsAV}{5{,}000}

\newcommand{\numMethodsAW}{0.1}

\newcommand{\numMethodsAX}{1.0}

\newcommand{\numMethodsAY}{0.5}

\newcommand{\numMethodsAZ}{10}

\newcommand{\numMethodsBA}{0}

\newcommand{\numMethodsBB}{18}

\newcommand{\numMethodsBF}{40}

\newcommand{\numMethodsBK}{7{,}039}

\newcommand{\numMethodsBL}{62{,}094}

\newcommand{\numMethodsBM}{1{,}300}

\newcommand{\numMethodsBN}{6,000}

\newcommand{\numMethodsBO}{1,000}

\newcommand{\numMethodsBP}{3,000}

\newcommand{\numMethodsBQ}{28{,}174}

\newcommand{\numMethodsBR}{31{,}694}

\newcommand{\numMethodsBU}{512}

\newcommand{\numMethodsBV}{1}

\newcommand{\numMethodsBW}{10}

\newcommand{\numMethodsBX}{5}

\newcommand{\numMethodsBY}{500}

\newcommand{\numMethodsBZ}{16}

\newcommand{\numMethodsCA}{96}

\newcommand{\numMethodsCB}{30{,}000}

\newcommand{\numMethodsCC}{2.88}

\newcommand{\numMethodsCD}{62{,}094}

\newcommand{\numMethodsCE}{46}

\newcommand{\numMethodsCK}{1{,}000}

\newcommand{\numMethodsCL}{30{,}000}

\newcommand{\numMethodsCM}{75}

\newcommand{\numMethodsCN}{7.5}

\newcommand{\numMethodsCO}{512}

\newcommand{\numMethodsCP}{1}

\newcommand{\numMethodsCQ}{2}

\newcommand{\numMethodsCR}{3}

\newcommand{\numMethodsCS}{62{,}094}

\newcommand{\numMethodsCT}{124{,}188}

\newcommand{\numMethodsCU}{186{,}282}

\newcommand{\numMethodsCW}{7{,}039}

\newcommand{\numMethodsCX}{28{,}174}

\newcommand{\numMethodsDB}{0.25}

\newcommand{\numMethodsDC}{0.75}

\newcommand{\numMethodsDE}{50}

\newcommand{\numMethodsDH}{121}

\newcommand{\numMethodsDI}{5}

\newcommand{\numMethodsDJ}{224}

\newcommand{\numMethodsDK}{512}

\newcommand{\numMethodsDL}{80}

\newcommand{\numMethodsDM}{20}

\newcommand{\numMethodsDN}{380}

\newcommand{\numMethodsDO}{18}

\newcommand{\numMethodsDP}{40}

\newcommand{\numMethodsDQ}{80}

\newcommand{\numMethodsDR}{381}

\newcommand{\numMethodsDS}{782}

\newcommand{\numMethodsDT}{40}

\newcommand{\numMethodsDU}{60}

\newcommand{\numMethodsDV}{60}

\newcommand{\numMethodsDW}{80}

\newcommand{\numMethodsDX}{18}

\newcommand{\numMethodsDY}{40}

\newcommand{\numMethodsDZ}{60}

\newcommand{\numMethodsEA}{80}

\newcommand{\numMethodsEB}{62{,}094}

\newcommand{\numMethodsEC}{1{,}300}

\newcommand{\numMethodsED}{7{,}039}

\newcommand{\numMethodsEE}{60{,}830}

\newcommand{\numMethodsEF}{1{,}564}

\newcommand{\numMethodsEG}{8{,}039}

\newcommand{\numMethodsEH}{57{,}737}

\newcommand{\numMethodsEI}{2{,}199}

\newcommand{\numMethodsEJ}{10{,}497}

\newcommand{\numMethodsEK}{16}

\newcommand{\numMethodsEL}{1}

\newcommand{\numMethodsEM}{5}

\newcommand{\numMethodsEN}{2}

\newcommand{\numMethodsEO}{2}

\newcommand{\numMethodsES}{1}

\newcommand{\numMethodsET}{10}

\newcommand{\numMethodsEU}{30}

\newcommand{\numMethodsEV}{50}

\newcommand{\numMethodsEW}{70}

\newcommand{\numMethodsEX}{100}

\newcommand{\numMethodsEY}{48}

\newcommand{\numMethodsFC}{14}

\newcommand{\numMethodsFD}{14}

\newcommand{\numMethodsFE}{121}

\newcommand{\numMethodsFH}{100}

\newcommand{\numMethodsFJ}{121}

\newcommand{\numMethodsFO}{6}

\newcommand{\numMethodsFP}{18}

\newcommand{\numMethodsFQ}{9}

\newcommand{\numMethodsFR}{9}

\newcommand{\numMethodsFS}{9}

\newcommand{\numMethodsFW}{10}

\newcommand{\numMethodsFX}{7{,}000}

\newcommand{\numMethodsFY}{20}

\newcommand{\numMethodsFZ}{43}

\newcommand{\numMethodsGA}{663}

\newcommand{\numMethodsGE}{62{,}094}

\newcommand{\numMethodsGF}{32{,}319}

\newcommand{\numMethodsGG}{77}

\newcommand{\numMethodsGH}{36{,}562}

\newcommand{\numMethodsGI}{24}

\newcommand{\numMethodsGK}{32}

\newcommand{\numMethodsGL}{0.20}

\newcommand{\numMethodsGM}{95}

\newcommand{\numMethodsGN}{500}

\newcommand{\numMethodsGR}{2}

\newcommand{\numMethodsGS}{95}

\newcommand{\numMethodsGT}{1{,}000}

\newcommand{\numMethodsGU}{0}

\newcommand{\numMethodsGV}{2}

\newcommand{\numMethodsGW}{0}

\newcommand{\numMethodsGX}{2}

\newcommand{\numMainFiguresA}{0}

\newcommand{\numMainFiguresB}{80}

\newcommand{\numMainFiguresC}{20}

\newcommand{\numMainFiguresD}{8}

\newcommand{\numMainFiguresE}{380}

\newcommand{\numMainFiguresF}{8}

\newcommand{\numMainFiguresG}{381}

\newcommand{\numMainFiguresH}{782}

\newcommand{\numMainFiguresI}{4}

\newcommand{\numMainFiguresK}{0.25}

\newcommand{\numMainFiguresL}{0.75}

\newcommand{\numMainFiguresM}{50}

\newcommand{\numMainFiguresU}{60}

\newcommand{\numMainFiguresV}{80}

\newcommand{\numMainFiguresW}{16}

\newcommand{\numMainFiguresX}{40}

\newcommand{\numMainFiguresY}{60}

\newcommand{\numMainFiguresZ}{0.5}

\newcommand{\numMainFiguresAA}{80}

\newcommand{\numMainFiguresAB}{60}

\newcommand{\numMainFiguresAC}{80}

\newcommand{\numMainFiguresAD}{80}

\newcommand{\numMainFiguresAE}{80}

\newcommand{\numMainFiguresAF}{7}

\newcommand{\numMainFiguresAG}{8}

\newcommand{\numMainFiguresAP}{121}

\newcommand{\numMainFiguresAQ}{14}

\newcommand{\numMainFiguresAR}{121}

\newcommand{\numMainFiguresAS}{14}

\newcommand{\numMainFiguresAT}{1000}

\newcommand{\numMainFiguresAU}{95}

\newcommand{\numMainFiguresAV}{9}

\newcommand{\numMainFiguresAW}{9}

\newcommand{\numMainFiguresAX}{9}

\newcommand{\numMainFiguresAY}{95}

\newcommand{\numMainFiguresBA}{9}

\newcommand{\numSupplementFiguresE}{29}

\newcommand{\numSupplementFiguresF}{50}

\newcommand{\numSupplementFiguresG}{8{,}039}

\newcommand{\numSupplementFiguresH}{10{,}497}

\newcommand{\numSupplementFiguresI}{0}

\newcommand{\numSupplementFiguresJ}{8}

\newcommand{\numSupplementFiguresK}{8}

\newcommand{\numSupplementFiguresL}{8}

\newcommand{\numSupplementFiguresM}{36{,}562}

\newcommand{\numSupplementFiguresN}{24}

\newcommand{\numSupplementFiguresO}{24}

\newcommand{\numSupplementFiguresP}{0.20}

\newcommand{\numSupplementFiguresQ}{0.67}

\newcommand{\numSupplementFiguresR}{0.77}

\newcommand{\numSupplementFiguresS}{95}

\newcommand{\numSupplementFiguresT}{500}

\newcommand{\numSupplementFiguresU}{2.8}

\newcommand{\numSupplementFiguresV}{1}

\newcommand{\numSupplementFiguresW}{5}

\newcommand{\numSupplementFiguresX}{2}

\newcommand{\numSupplementFiguresY}{2}

\newcommand{\scModels}{18}
\newcommand{\scCells}{16}

\newcommand{\scStop}{20{,}000}

\newcommand{\scSelectFloor}{5{,}000}
\newcommand{\scLevel}{95}
\newcommand{\scBootstrapDraws}{531{,}441}
\newcommand{\scPermutations}{400}
\newcommand{\scSeedFirst}{42}
\newcommand{\scSeedSecond}{43}
\newcommand{\scSeedThird}{44}

\newcommand{\scRNWins}{16}
\newcommand{\scRNBaseEffect}{-2.364}
\newcommand{\scRNBaseLo}{-3.710}
\newcommand{\scRNBaseHi}{-1.018}

\newcommand{\scRNFairEffect}{-1.246}
\newcommand{\scRNFairLo}{-2.569}
\newcommand{\scRNFairHi}{+0.078}

\newcommand{\scFIDWins}{12}

\newcommand{\dsConfigurations}{48}

\newcommand{\dsLabels}{8}
\newcommand{\dsGroups}{10}

\newcommand{\dsSweepResolution}{224}

\newcommand{\dsBalancedRG}{414\,980}
\newcommand{\dsBalancedCD}{415\,072}
\newcommand{\dsSynthMimicBase}{0.7849}
\newcommand{\dsSynthMimicComp}{0.7962}
\newcommand{\dsSynthMimicGain}{0.0113}
\newcommand{\dsSynthMimicLo}{0.0067}
\newcommand{\dsSynthMimicHi}{0.0158}
\newcommand{\dsWinsMimic}{40}
\newcommand{\dsScratchWinsMimic}{23}
\newcommand{\dsImageNetWinsMimic}{17}

\newcommand{\dsPtxSensitivityMimicBase}{67.6}
\newcommand{\dsPtxSensitivityMimicComp}{70.7}

\newcommand{\dsSynthChexBase}{0.7275}
\newcommand{\dsSynthChexComp}{0.7363}
\newcommand{\dsSynthChexGain}{0.0088}
\newcommand{\dsSynthChexLo}{0.0039}
\newcommand{\dsSynthChexHi}{0.0137}
\newcommand{\dsWinsChex}{40}

\newcommand{\dsPtxAurocChex}{0.0140}
\newcommand{\dsPtxSensitivityChexBase}{39.8}
\newcommand{\dsPtxSensitivityChexComp}{42.6}

\newcommand{\dsWorstPercentSex}{84.4}

\newcommand{\dsWorstPercentRace}{87.5}

\newcommand{\dsWorstPercentAge}{71.9}

\IfFileExists{supplement.aux}{\externaldocument[][nocite]{supplement}[supplement.pdf]}{\IfFileExists{supplement-labels.tex}{\makeatletter
\newlabel{fig:zeroshot-design}{{1}{2}{Two zero-shot withheld splits of the MIMIC-CXR distribution. Rows: age-bin $\times $ sex. Columns: race (outer) $\times $ split (inner). \textbf {a}, Natural distribution: image counts per age $\times $ sex $\times $ race $\times $ split intersection before any manipulation. \textbf {b}, Distribution after reallocation under the two zero-shot splits: each split removes eight disjoint three-way intersectional cells from train (count shown as $\numMainFiguresA {}$) and reallocates the affected patients, $\numMainFiguresB {}\%$ to test and $\numMainFiguresC {}\%$ to validation, inflating the val/test counts for the withheld cells. Red outlines mark the eight cells of Holdout~A (the rarest \numMainFiguresD {}, natural training count below $\numMainFiguresE {}$); violet outlines mark the eight cells of Holdout~B (the next rarest \numMainFiguresF {}, natural training count $\numMainFiguresG {}$--$\numMainFiguresH {}$). The two splits are disjoint, so each cell falls into exactly one category}{figure.caption.1}{}}
\newlabel{app:zeroshot-inception}{{1}{3}{Zero-shot results across repeated training runs and sensitivity analyses}{section.1}{}}
\newlabel{app:seed-campaign}{{1}{3}{Zero-shot results across repeated training runs and sensitivity analyses}{section.1}{}}
\newlabel{fig:zeroshot-cells}{{2}{3}{FID-RadImageNet within each withheld intersection across repeated training runs. Panels a and b show the eight target intersections in Holdouts A and B, respectively. Colours identify the generators. Open circles show individual trained models and filled diamonds show the mean across three models for that method and intersection. The eight cell observations contributed by one model share that fitted generator and are not independent training replicates. Lower distances indicate closer real--synthetic feature distributions. CompDiff has the lowest mean across three runs in all $\scRNWins $ cells; these descriptive cell means do not imply that every individual run wins. No cell confidence interval is shown. Numerical means and standard deviations are in Supplementary Table~\ref {tab:seed-cells}}{figure.caption.2}{}}
\newlabel{tab:seed-cells}{{1}{4}{Fidelity at each withheld intersection across independently trained models}{table.caption.3}{}}
\newlabel{fig:zeroshot-inception}{{3}{5}{Zero-shot comparison across repeated training runs using Inception-v3 features. This uses the same models and generated cohorts as Figure~\ref {fig:zeroshot}, changing only the feature backbone. \textbf {a,b}, Each open circle is one model's mean FID across its eight withheld intersections; filled diamonds show means across three independently trained models. \textbf {c,d}, CompDiff minus comparator differences averaged equally across cells and holdouts, with pointwise $\scLevel \%$ Welch--Satterthwaite confidence intervals. Negative differences favour CompDiff. Each difference has its own zero reference, and both panels use the same scale. Both intervals for the average differences include zero. Results for individual cells appear in Supplementary Figure~\ref {fig:zeroshot-cells-inception}}{figure.caption.4}{}}
\newlabel{fig:zeroshot-cells-inception}{{4}{6}{Inception FID within each withheld intersection across repeated training runs. This figure uses the same generated cohorts and conventions as Supplementary Figure~\ref {fig:zeroshot-cells}. Open circles show individual model scores and filled diamonds their means across three runs, separated by method within each cell. CompDiff has the lowest mean in $\scFIDWins $ of $\scCells $ cells. The corresponding average effects and intervals across trained models are shown in Supplementary Figure~\ref {fig:zeroshot-inception}; the cell observations are repeated outcomes from those same models, not additional independent replicates}{figure.caption.5}{}}
\newlabel{tab:seed-runs}{{2}{7}{Selected checkpoints for the zero-shot comparison across repeated training runs}{table.caption.6}{}}
\newlabel{app:estimator-robustness}{{2}{7}{Robustness of the zero-shot ranking to the choice of fidelity estimator}{section.2}{}}
\newlabel{fig:estimator-size-curve}{{5}{8}{The fidelity estimator is size-dependent; the kernel estimator is not. Both metrics are read on the \emph {same} subsamples: each cell's candidate and reference sets are reduced to a common size and the metric is expressed as a percentage change against that cell's own reading at $\robSizeAnchor $ images. The generator, the cell and the image pool are fixed, so any vertical movement is attributable to the estimator alone. Lines are medians over the $\robSizeCombos $ cell $\times $ arm curves with the interquartile range shaded; both series share one axis, so the difference in magnitude can be read directly. FID-RadImageNet inflates as the cell shrinks, by a median $\numSupplementFiguresE {}\%$ at $\numSupplementFiguresF {}$ images, while KID-RadImageNet~\cite {binkowski2018mmdgans}, which is unbiased, stays at zero on the identical images. The ranking consequences of this bias are given in Supplementary Table~\ref {tab:estimator-robustness}}{figure.caption.8}{}}
\newlabel{tab:estimator-robustness}{{3}{9}{Robustness of the zero-shot arm ranking to the fidelity estimator. Each row rescores every cell with a different estimator and compares the resulting cell arm ordering against the ordering under the primary metric. Agreement is reported over all cells and over the withheld intersectional cells alone. Estimators: FID$_\infty $ extrapolates to infinite sample size in the candidate dimension ($1/k$) or jointly in candidate and reference ($1/k,1/m$); FID at common $n$ rescores every cell at a shared sample size; KID is an unbiased kernel estimator. Ordering agreement counts require the complete arm ordering to match}{table.caption.11}{}}
\newlabel{app:selection-control}{{3}{9}{Post hoc sample selection does not substitute for demographic conditioning}{section.3}{}}
\newlabel{app:flat-composer}{{4}{10}{Zero-shot generalisation with an unstructured composer of matched size}{section.4}{}}
\newlabel{tab:composer-paired-test}{{4}{11}{Paired composer control on the full data corpus, evaluated on the withheld \emph {test} split at a matched step of $10{,}000$ (all $7{,}039$ test images, mean $\pm $ SD over three generation seeds). The two arms were trained together under an identical recipe and differ only in the composer. The last row is the spread across three independent training runs of the final configuration, the scale against which these differences should be read}{table.caption.13}{}}
\newlabel{tab:steps-to-floor}{{5}{11}{First checkpoint at which each full data run satisfies all three controllability floors of Eq.~\ref {eq:ckpt-rule}. The three factorised runs are independent training runs of the final configuration}{table.caption.14}{}}
\newlabel{fig:composer-zeroshot}{{6}{11}{Fidelity of the two composers at unseen intersections. \textbf {a}, Holdout~A, the eight rarest intersections. \textbf {b}, Holdout~B, the eight next-rarest intersections. Both panels show FID-RadImageNet within each cell (lower is better) for the factorised composer (the seed-$\scSeedFirst $ CompDiff models in Supplementary Table~\ref {tab:seed-runs}), the unstructured composer with a matched parameter count, and the RoentGen-v2 baseline as a scale reference; FairDiffusion is omitted because this figure answers a composer question. Every arm is scored at full coverage of its split ($\numSupplementFiguresG {}$ and $\numSupplementFiguresH {}$ images). Checkpoint selection is specified in Section~\ref {subsec:composer-control}. The footer counts wins for the \emph {factorised} arm against the other two arms shown, so $\numSupplementFiguresI {}/\numSupplementFiguresJ {}$ on Holdout~A means the unstructured control attains the lowest value in all eight of those cells, and $\numSupplementFiguresK {}/\numSupplementFiguresL {}$ on Holdout~B the reverse; the sign of the composer difference does not reproduce across the two splits}{figure.caption.15}{}}
\newlabel{app:reader-results}{{5}{12}{Reader study contrasts and sensitivity analyses}{section.5}{}}
\newlabel{tab:reader-contrasts}{{6}{12}{Mixed-effects contrasts on the blinded reader study. Rating estimates are differences in mean score from Equation~\eqref {eq:reader-lmm}, with two-sided Wald tests and 95\% intervals. Realism uses $\rsRealismN $ ratings from $\rsNSets $ sets; correspondence uses $\rsCorrN $ ratings from $\readerCorrSets $ sets. Both models include the two readers. $p_{\mathrm {BH}}$ is Benjamini--Hochberg-adjusted across the $\rsFamilySize $ comparisons. Forced-choice rows give selections for each source among assessments containing both. Two-sided exact binomial tests condition on selection of either compared source, giving 50 and 79 trials, respectively; preferences for a third source are excluded. Numerical $p$-values are shown to three significant digits}{table.caption.16}{}}
\newlabel{tab:reader-robustness}{{7}{12}{Robustness of the reader study contrasts. $\Delta $ and the Wald column come from the mixed-effects model of Equation~\eqref {eq:reader-lmm}. The paired sample size counts reader--set assessments with both source ratings and applies to the permutation and Wilcoxon tests. The permutation column uses $\rsNPerm $ sign-flip draws; the Wilcoxon column uses signed ranks. The Cluster column gives Wald tests from ordinary least squares with standard errors clustered by image set. The mixed and ordinary least-squares models use 287 realism ratings from 92 sets or 277 correspondence ratings from 89 sets. All tests are two-sided; $p$-values are unadjusted and shown to three significant digits}{table.caption.17}{}}
\newlabel{app:reader-interface}{{6}{13}{Reader study rating interface}{section.6}{}}
\newlabel{fig:reader-interface}{{7}{13}{The blinded reader study rating interface (Section~\ref {subsec:reader-study}). The clinical impression is displayed at the top, followed by three chest X-rays displayed side by side in randomised order; each image carries an anatomical realism control ($\numSupplementFiguresV {}$--$\numSupplementFiguresW {}$) and a prompt correspondence control ($-\numSupplementFiguresX {}$ to $+\numSupplementFiguresY {}$), with a forced ``most realistic'' choice below the row and unreadable and N/A options for each image. Raters are blinded to which images are synthetic or real and in what proportion}{figure.caption.18}{}}
\newlabel{app:checkpoint-results}{{7}{13}{Validation trajectories}{section.7}{}}
\newlabel{tab:ckpt-baseline}{{8}{14}{Validation trajectory for the \textbf {Baseline SD} generator over the shared $30{,}000$-step budget (training run with seed~11). This run is one of the three independently trained Baseline SD models whose selected checkpoints are averaged into the Baseline SD bars of the main text; the values tabulated here are those of this single run, not the means across three runs, and its selected checkpoint is marked $\star $. All three generators are scored at the identical 24 checkpoints. The selected checkpoint (step 8,500) is the checkpoint with the lowest FID among those meeting the demographic control floors of Section~\ref {app:checkpoint-selection}; rows marked $\dagger $ fail at least one floor and are therefore not eligible, regardless of their FID. Metrics follow Section~\ref {subsec:eval-synth} (FID-RIN = FID-RadImageNet; MS-SSIM as defined there, read against the $0.25$--$0.75$ band; Sex/Race = prompt alignment accuracy; Age RMSE in years) and are computed on the validation split}{table.caption.20}{}}
\newlabel{tab:ckpt-fairdiffusion}{{9}{15}{Validation trajectory for the \textbf {FairDiffusion} generator over the shared $30{,}000$-step budget (training run with seed~42). This run is one of the three independently trained FairDiffusion models whose selected checkpoints are averaged into the FairDiffusion bars of the main text; the values tabulated here are those of this single run, not the means across three runs, and its selected checkpoint is marked $\star $. All three generators are scored at the identical 24 checkpoints. The selected checkpoint (step 11,000) is the checkpoint with the lowest FID among those meeting the demographic control floors of Section~\ref {app:checkpoint-selection}; rows marked $\dagger $ fail at least one floor and are therefore not eligible, regardless of their FID. Metrics follow Section~\ref {subsec:eval-synth} (FID-RIN = FID-RadImageNet; MS-SSIM as defined there, read against the $0.25$--$0.75$ band; Sex/Race = prompt alignment accuracy; Age RMSE in years) and are computed on the validation split}{table.caption.21}{}}
\newlabel{tab:ckpt-compdiff}{{10}{16}{Validation trajectory for the \textbf {CompDiff} generator over the shared $30{,}000$-step budget (training run with seed~42). This run is one of the three independently trained CompDiff models whose selected checkpoints are averaged into the CompDiff bars of the main text; the values tabulated here are those of this single run, not the means across three runs, and its selected checkpoint is marked $\star $. All three generators are scored at the identical 24 checkpoints. The selected checkpoint (step 10,000) is the checkpoint with the lowest FID among those meeting the demographic control floors of Section~\ref {app:checkpoint-selection}; rows marked $\dagger $ fail at least one floor and are therefore not eligible, regardless of their FID. Metrics follow Section~\ref {subsec:eval-synth} (FID-RIN = FID-RadImageNet; MS-SSIM as defined there, read against the $0.25$--$0.75$ band; Sex/Race = prompt alignment accuracy; Age RMSE in years) and are computed on the validation split}{table.caption.22}{}}
\newlabel{app:downstream-evidence}{{8}{17}{Downstream diagnostic and subgroup evidence}{section.8}{}}
\newlabel{tab:downstream-runs}{{11}{17}{Repeated classifier training using AdamW and a consistent set of eight diagnostic labels}{table.caption.24}{}}
\newlabel{tab:downstream-all-models}{{12}{17}{Sensitivity to including all available classifiers trained on synthetic images alone within the AdamW family}{table.caption.26}{}}
\newlabel{fig:downstream-synthetic-runs}{{8}{18}{Independent classifier runs using synthetic chest images alone. \textbf {a}--\textbf {d}, Evaluation on MIMIC-CXR for the natural 1$\times $, 2$\times $ and 3$\times $ corpora and the balanced corpus. \textbf {e}--\textbf {h}, The same classifiers evaluated on CheXpert, in the same order. Each panel shows four classifier runs per generator for one fixed corpus regime under the AdamW recipe. Hollow circles are individual runs and filled diamonds are their means. RG denotes RoentGen-v2 and CD denotes CompDiff. AUROC$_8$ uses the same eight diagnostic labels as Figure~\ref {fig:downstream-d3deep}. The same 32 trained classifiers are evaluated on MIMIC-CXR and CheXpert. Natural corpora contain one, two or three times the real training set size; the balanced corpus samples demographic intersections uniformly}{figure.caption.25}{}}
\newlabel{fig:downstream-sweep-mimic}{{9}{19}{Complete sweep of downstream training configurations on MIMIC-CXR. Panels show the four synthetic corpus regimes; lines distinguish initialisation from scratch and from ImageNet. Every point is a CompDiff minus RoentGen-v2 AUROC$_8$ difference at one fraction of real data used for fine-tuning, giving 48 matched configurations. Positive values favour CompDiff. All scores average the same eight diagnostic labels. Axes are shared with the corresponding figure for the other test dataset. Lines connect settings in the sweep and do not imply independent replications or interpolation between evaluated fractions}{figure.caption.27}{}}
\newlabel{fig:downstream-sweep-chexpert}{{10}{20}{Complete sweep of downstream training configurations on CheXpert. Panels show the four synthetic corpus regimes; lines distinguish initialisation from scratch and from ImageNet. Every point is a CompDiff minus RoentGen-v2 AUROC$_8$ difference at one fraction of real data used for fine-tuning, giving 48 matched configurations. Positive values favour CompDiff. All scores average the same eight diagnostic labels. Axes are shared with the corresponding figure for the other test dataset. Lines connect settings in the sweep and do not imply independent replications or interpolation between evaluated fractions}{figure.caption.28}{}}
\newlabel{tab:downstream-labels}{{13}{21}{Diagnostic gains across the complete fine-tuning sweep}{table.caption.29}{}}
\newlabel{tab:downstream-groups}{{14}{22}{Subgroup sample sizes and gains across training configurations}{table.caption.31}{}}
\newlabel{tab:downstream-uncertainty}{{15}{22}{Patient resampling in the three sparsely sampled MIMIC-CXR groups}{table.caption.32}{}}
\newlabel{app:clip-sensitivity}{{9}{23}{Demographic tokens in the prompt: truncation statistics and probe of the encoder}{section.9}{}}
\newlabel{fig:clip-sensitivity}{{11}{24}{CLIP's text encoder is sensitive to demographic tokens, and this sensitivity falls steeply with clinical prompt length. For each of $\numSupplementFiguresM {}$ MIMIC-CXR clinical impressions, the clinical text and the same text with each of $\numSupplementFiguresN {}$ demographic prefixes (three age descriptors $\times $ two sexes $\times $ four race labels) are passed through Stable Diffusion 2.1's frozen CLIP text encoder; we report the cosine perturbation between embeddings with and without demographic text, averaged over the $\numSupplementFiguresO {}$ demographic variants. \textbf {a}, Perturbation for each report against clinical prompt length on linear axes. Each point is one report, coloured by its clinical density (fraction of words that are recognised clinical entities); the dark line is a LOWESS smoother (span $\numSupplementFiguresP {}$). The same relationship is well described by a power law in log--log space, $y \propto L^{-\numSupplementFiguresQ {}}$ ($R^2_{\log \log }=\numSupplementFiguresR {}$). \textbf {b}, Mean perturbation within quartiles of word count with bootstrap $\numSupplementFiguresS {}\%$ CIs (\numSupplementFiguresT {} resamples); impressions in the shortest quartile are perturbed $\numSupplementFiguresU {}\times $ more than impressions in the longest quartile}{figure.caption.35}{}}
\makeatother}{}}
\title{\ManuscriptTitle}
\author{\begin{minipage}{\dimexpr\textwidth-2\tabcolsep\relax}\centering
Mahmoud K. Ibrahim\textsuperscript{1,2,3,*}, Bart Elen\textsuperscript{3}, Chang Sun\textsuperscript{1,2},
Ahmad Jiblawi\textsuperscript{4}, Mohammed M. Saleh\textsuperscript{5}, Maryam K. Ibrahim\textsuperscript{6},
G\"okhan Ertaylan\textsuperscript{3} and Michel Dumontier\textsuperscript{1,2}\\[1em]
\small
\textsuperscript{1}Institute of Data Science, Faculty of Science and Engineering, Maastricht University, Maastricht, The Netherlands\\[0.35em]
\textsuperscript{2}Department of Advanced Computing Sciences, Faculty of Science and Engineering, Maastricht University, Maastricht, The Netherlands\\[0.35em]
\textsuperscript{3}VITO, Belgium\\[0.35em]
\textsuperscript{4}Icahn School of Medicine at Mount Sinai, The Mount Sinai Hospital, New York City, USA\\[0.35em]
\textsuperscript{5}Department of Radiology, Division of Abdominal Imaging, Massachusetts General Hospital, Harvard Medical School, Boston, Massachusetts, USA\\[0.35em]
\textsuperscript{6}Division of Gastroenterology, Massachusetts General Hospital, Boston, Massachusetts, USA\\[1em]
\textsuperscript{*}Correspondence: \href{mailto:mahmoud.ibrahim@maastrichtuniversity.nl}{mahmoud.ibrahim@maastrichtuniversity.nl}\\
ORCID: \href{https://orcid.org/0000-0002-5099-7609}{0000-0002-5099-7609}
\end{minipage}}
\date{}

\date{}
\begin{document}
\maketitle
\ResourceLinks
\begin{abstract}
Medical image generators trained on imbalanced data can fail at demographic intersections absent from training. We introduce CompDiff, which encodes age, sex and race separately and composes supervised demographic tokens alongside clinical text. Across chest radiographs and fundus images, CompDiff improves overall and subgroup fidelity relative to prompt conditioning (RoentGen-v2) and loss reweighting (FairDiffusion). It generalises in zero-shot generation to $\scCells$ chest X-ray intersections excluded from training, achieving the lowest mean FID-RadImageNet in every intersection. In a blinded reader study of these unseen intersections, two radiologists gave CompDiff the highest mean scores among generators for anatomical realism and agreement with the clinical impression, and selected its images most often as the most realistic. Pretraining with CompDiff images improved downstream classification, while CompDiff audit cohorts reduced estimation error on rare intersections. These findings support compositional demographic conditioning for extending medical image synthesis to underserved populations.

\end{abstract}
\clearpage
\section{Introduction}\label{sec:intro}\label{sec:related-work}\label{subsec:rw-fair-medical}

Audit studies of classifiers for chest radiographs have documented that models trained on demographically imbalanced imaging data can systematically underserve underrepresented patient subgroups, with the largest disparities affecting intersections such as Black women~\cite{seyyed2021underdiagnosis,glocker2023algorithmic}. We use \emph{cell} to denote a subgroup defined by a specific combination of demographic attributes, such as an age $\times$ sex $\times$ race group. Synthetic augmentation with text-to-image diffusion models is a widely proposed remedy: if a generator can synthesise realistic samples of an underrepresented subgroup, it can rebalance the training corpus of a downstream classifier~\cite{chambon2022roentgen,bluethgen2024vision,moroianu2025improvingperformancerobustnessfairness,ktena2024generative}. This remedy, however, depends on the generator itself synthesising equally well across demographic groups.

\begin{figure*}[tp]
\centering
\DisplayMark{fig:problem/plot-top}
\begin{adjustbox}{max totalsize={\textwidth}{0.665\textheight},center}
\begin{minipage}{\textwidth}
\centering

\includegraphics[width=\textwidth]{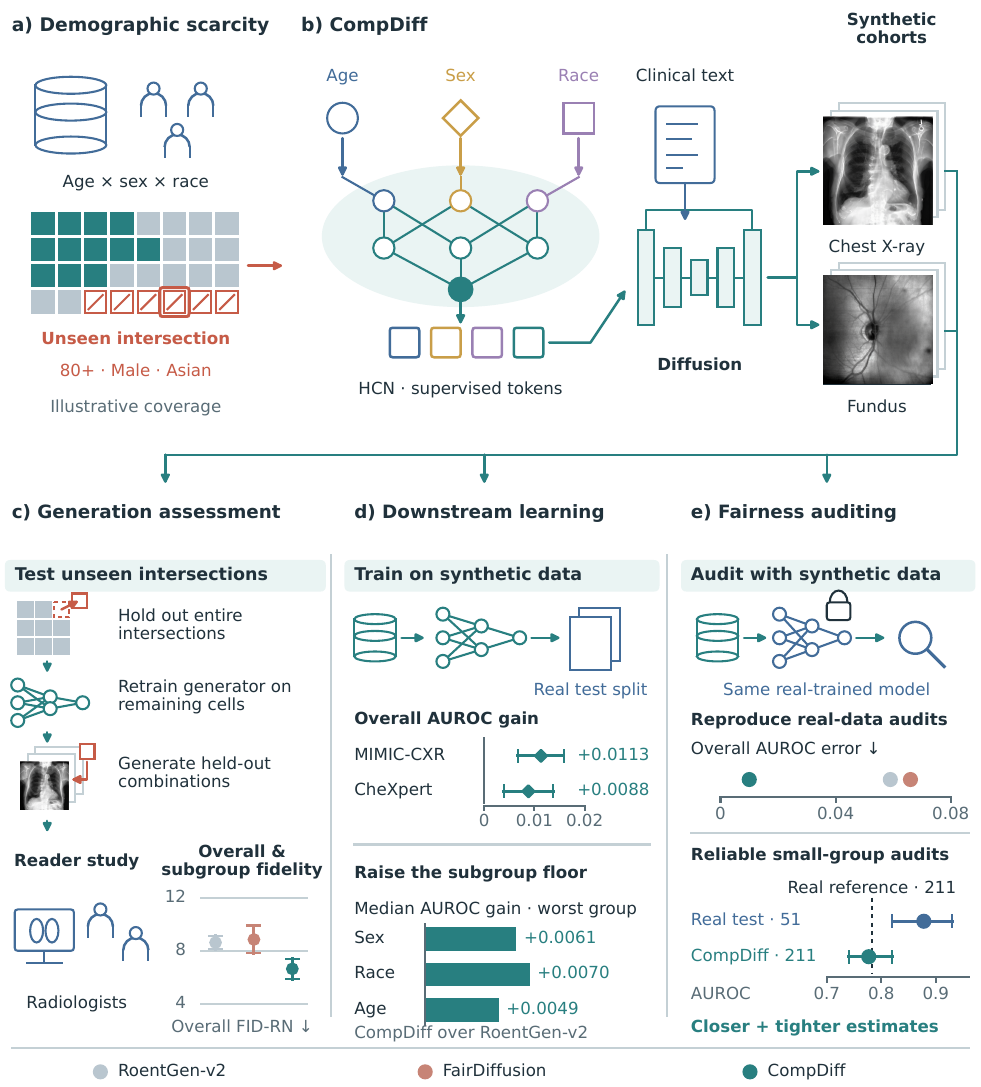}

\end{minipage}
\end{adjustbox}
\par\DisplayMark{fig:problem/caption-top}
\caption{CompDiff: generating across demographic intersections to support learning and fairness auditing.
\textbf{(a)} Uneven demographic coverage leaves some combinations of age, sex and race absent from training.
\textbf{(b)} The Hierarchical Conditioner Network (HCN) composes supervised demographic tokens, which guide diffusion alongside separately encoded clinical text. Coverage and model diagrams are schematic.
\textbf{(c)} Generation is assessed through a blinded radiologist study of intersections withheld from training and comparisons of overall and subgroup fidelity. The example plot shows overall chest X-ray Fr\'echet inception distance with RadImageNet features (FID-RN), with standard deviations across three training runs.
\textbf{(d)} Synthetic training is evaluated for overall classification and for the weakest demographic groups. Diamonds show CompDiff's overall gains over RoentGen-v2 after training on synthetic images alone; bars show gains for the baseline's weakest groups after synthetic pretraining and fine-tuning on real images. Diamonds have pointwise 95\% Welch confidence intervals from four classifier runs per generator and corpus, averaging four corpora equally. Bars show medians across 48 configurations evaluated on both test datasets.
\textbf{(e)} Synthetic cohorts audit a fixed classifier in two ways: reproducing evaluations on real data overall and improving the accuracy and precision of estimates for small subgroups. Upper points show absolute error relative to real-data evaluation. Below, real and CompDiff estimates for Asian women aged $\numResultsEO$--$\numResultsEP$ are compared with the dashed real reference; intervals are 95\% bootstrap confidence intervals from 1,000 resamples.
AUROC denotes area under the receiver operating characteristic curve; panel (d) averages eight diagnostic labels. Grey, orange and green denote RoentGen-v2, FairDiffusion and CompDiff; blue denotes real data.}
\label{fig:problem}
\par\DisplayMark{fig:problem/caption-bottom}
\end{figure*}

When trained on imbalanced data, a diffusion model can achieve strong average fidelity while its quality on individual subgroups is left to chance (Figure~\ref{fig:problem}); on the MIMIC-CXR training distribution used in this work, the rarest age $\times$ sex $\times$ race intersections contain only $\numIntroductionA{}$--$\numIntroductionB{}$ training images, from $\numIntroductionC{}$--$\numIntroductionD{}$ patients (Supplementary Figure~\ref{fig:zeroshot-design}a), and combined with rare pathologies these intersections are frequently absent altogether (Figure~\ref{fig:problem}a). A training corpus may contain elderly patients, Asian patients, and female patients, yet have zero examples at the intersection of all three with a specific pathology. No amount of oversampling or reweighting can address groups that do not exist in the training data, and prompt conditioning supplies no mechanism that supervises how such a group is composed from the groups that do. We refer to this gap as the \emph{imbalanced generator problem}. The stakes are not merely aesthetic: when synthetic images systematically have lower fidelity for a subgroup, using them to augment or rebalance a training set teaches the downstream classifier a distorted picture of exactly the patients the intervention was meant to protect, reintroducing or even amplifying the disparities the augmentation was intended to remove. 

Existing generation methods address unequal data in several ways. FairDiffusion uses Fair Bayesian Perturbation to adapt the diffusion training loss across demographic groups~\cite{friedrich2024fairdiffusion}. FairGen combines resampling, regularisation for class diversity and preference optimisation aligned with physician judgements to improve demographic equity and clinically relevant synthesis~\cite{li2026fairgen}. RoentMod edits pathology in existing chest radiographs while preserving unrelated anatomy, using the resulting counterfactual images to identify and reduce shortcut learning~\cite{cooke2026roentmod}. ChexGen is a foundation model for chest radiography that supports synthesis guided by text, masks and bounding boxes, including demographic control and downstream fairness applications~\cite{ji2026chexgen}. These studies establish complementary uses of synthetic medical images. Here we examine a specific design question: whether a dedicated demographic conditioner can generate intersections deliberately removed from training, and whether those images support downstream learning and fairness audits. FairGen, RoentMod and ChexGen provide context for this question; they were not retrained as experimental comparators in this study.

We test whether demographic attributes can be composed from components learned in observed cells, whether this representation improves fidelity and subgroup equity relative to prompt conditioning (RoentGen-v2) and loss reweighting (FairDiffusion), and whether the resulting cohorts improve downstream classification and the estimation of performance on rare intersections. The distinction between conditioning mechanisms is summarised in Table~\ref{tab:methods-comparison}.

\begin{table*}[!tp]
\centering
\caption{Conditioning and fairness mechanisms relevant to CompDiff. The table distinguishes the intervention and the task it supports; it does not report a matched performance comparison with every cited system. The fused demographic embedding and second text branch are controls from this study. The unstructured composer with a matched parameter count retains CompDiff's separate attribute encoders and supervised output tokens.}
\label{tab:methods-comparison}
\small
\begin{tabularx}{\textwidth}{@{}p{0.22\textwidth}YY@{}}
\toprule
Method & Intervention & Evaluation focus \\
\midrule
Standard latent diffusion~\cite{rombach2022stablediffusion} & Text-conditioned denoising & Image synthesis from prompts \\
FairDiffusion~\cite{friedrich2024fairdiffusion} & Adaptive weighting of the diffusion loss & Demographic equity in medical synthesis \\
FairGen~\cite{li2026fairgen} & Resampling, diversity regularisation and physician preferences & Equitable, clinically relevant medical synthesis \\
RoentMod~\cite{cooke2026roentmod} & Image-to-image modification of pathology & Counterfactual shortcut testing and correction \\
ChexGen~\cite{ji2026chexgen} & Foundation model with text and spatial controls & Chest radiograph synthesis and downstream model development \\
Composable Diffusion~\cite{liu2022composable} & Composition of conditional scores at sampling & Object and scene combinations \\
Fused demographic embedding & Demographic embeddings fused into one token, with auxiliary supervision & Conditioning control (Section~\ref{subsec:ablations}) \\
Second text branch & Additional demographic text pathway & Conditioning control (Section~\ref{subsec:ablations}) \\
CompDiff & Separate encoders and supervised tokens for age, sex and race, plus a composed token & Unseen demographic intersections, two imaging modalities, blinded chest X-ray reader study, downstream learning and auditing \\
\bottomrule
\end{tabularx}
\end{table*}

\begin{figure*}[tp]
    \centering
\DisplayMark{fig:architecture/plot-top}
\begin{adjustbox}{max totalsize={\textwidth}{0.76\textheight},center}
\begin{minipage}{\textwidth}
\centering

    \includegraphics[width=\textwidth]{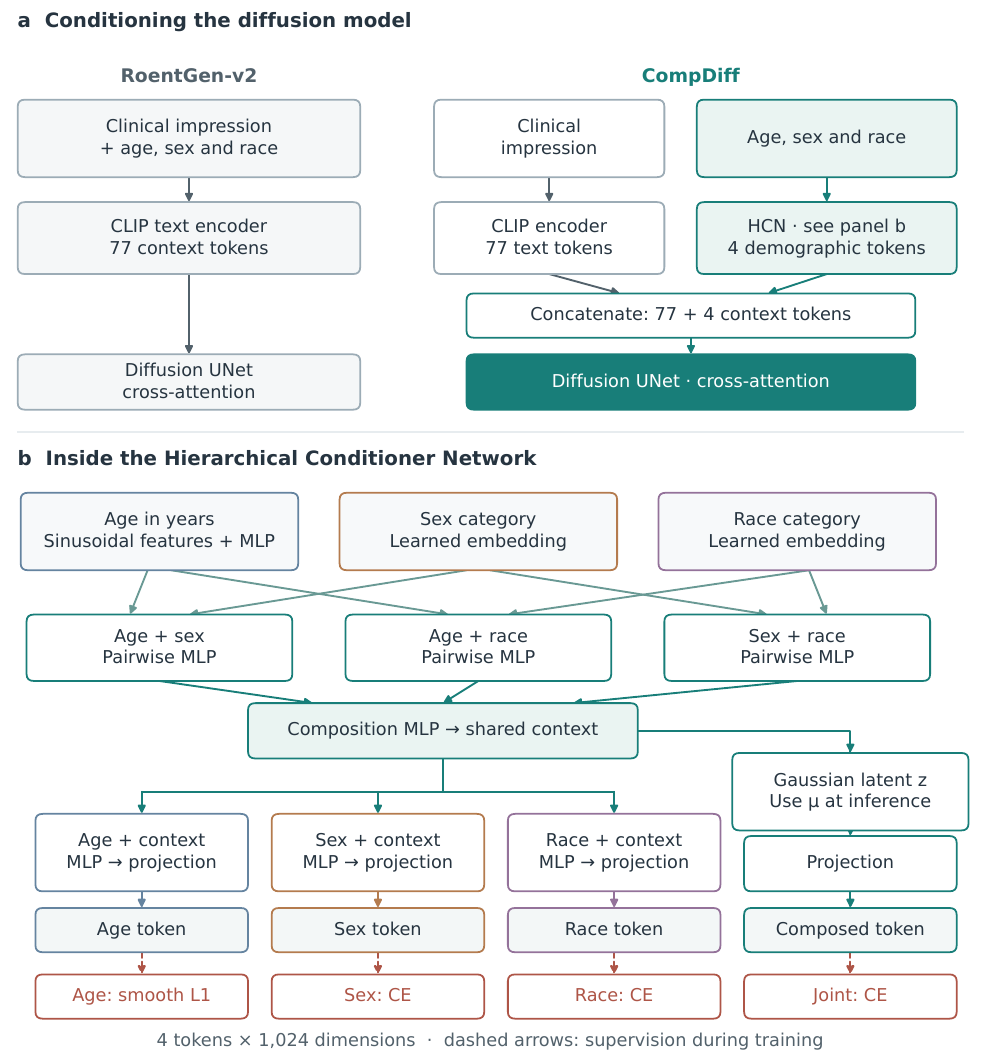}
    
\end{minipage}
\end{adjustbox}
\par\DisplayMark{fig:architecture/caption-top}
\caption{\textbf{CompDiff architecture and demographic conditioning.}
\textbf{a,} RoentGen-v2 encodes clinical and demographic text together through CLIP's
77-token context. CompDiff encodes the clinical impression through CLIP and age, sex
and race through the Hierarchical Conditioner Network (HCN). HCN supplies
$\numMainFiguresI{}$ additional context tokens, concatenated with the text tokens
before diffusion UNet cross-attention.
\textbf{b,} HCN encodes age continuously and sex and race with separate categorical
embeddings. Three dedicated multilayer perceptrons (MLPs) combine the attribute pairs;
a composition MLP fuses their outputs into a shared demographic context. Each
attribute embedding is concatenated with this context and projected to its own token.
A Gaussian latent supplies only the composed token: it is sampled during training
and set to its mean at inference. The four projected tokens each have 1,024 dimensions.
Auxiliary heads supervise age, sex and race from their respective tokens and the joint
demographic label from the composed token. Dashed arrows mark this training-only
supervision; CE denotes cross-entropy.}
    \label{fig:architecture}
\par\DisplayMark{fig:architecture/caption-bottom}
\end{figure*}

We propose \textbf{CompDiff}, which addresses the imbalanced generator problem at the \emph{representation} level (Figure~\ref{fig:problem}b). Our key idea is that demographic identity can be modelled compositionally: a rare intersection such as ``\numIntroductionE{}+ Asian female'' can be \emph{composed} from separate representations of age, sex and race learned from more abundant cells. CompDiff instantiates this idea in a dedicated Hierarchical Conditioner Network (HCN). It encodes age as a continuous value and sex and race as categories, then produces one demographic token per attribute plus a composed token, concatenated to the CLIP~\cite{clip} text embeddings as cross-attention context for the diffusion UNet. HCN is named for the pairwise hierarchy of the composer we instantiate. The resulting interface separates demographic control from the clinical text pathway (Figure~\ref{fig:architecture}).

We evaluate CompDiff on chest X-rays from MIMIC-CXR~\cite{johnson2019mimiccxr} and fundus images from FairGenMed~\cite{friedrich2024fairdiffusion}, with matched RoentGen-v2 and FairDiffusion baselines. CompDiff attains the best value on $\figbWinsTotal$ of $\numIntroductionF{}$ subgroup comparisons and the lowest mean FID-RadImageNet across three training runs in all $\scCells$ unseen chest X-ray intersections. A separate blinded radiologist study tests realism and clinical correspondence at these intersections. Downstream experiments assess synthetic training data on MIMIC-CXR and CheXpert~\cite{irvin2019chexpert}, while separate audit experiments test how closely synthetic cohorts reproduce estimates from real data overall and on rare intersections. Ablations distinguish the contribution of separate encoding and supervision for each attribute from the network used to combine them. Figure~\ref{fig:problem}c--e outlines the protocol for withholding intersections and distinguishes overall and subgroup benefits of synthetic cohorts for classifier training and fairness auditing.

\clearpage
\section{Results}\label{sec:results}

We first compared the fidelity and conditioning alignment of CompDiff, RoentGen-v2 and FairDiffusion on chest radiographs and fundus images. Chest X-ray experiments then tested generation at unseen demographic intersections, radiologist assessment, classifier training and classifier auditing using synthetic cohorts. These experiments distinguish three uses of classifiers: measuring whether generated images express the requested labels, learning from synthetic training data, and evaluating a fixed model on synthetic test data.

\subsection{Generation fidelity and alignment}\label{subsec:overall}

\begin{figure*}[tp]
\centering
\DisplayMark{fig:overall/plot-top}
\begin{adjustbox}{max totalsize={\textwidth}{0.67\textheight},center}
\begin{minipage}{\textwidth}
\centering

\includegraphics[width=\textwidth]{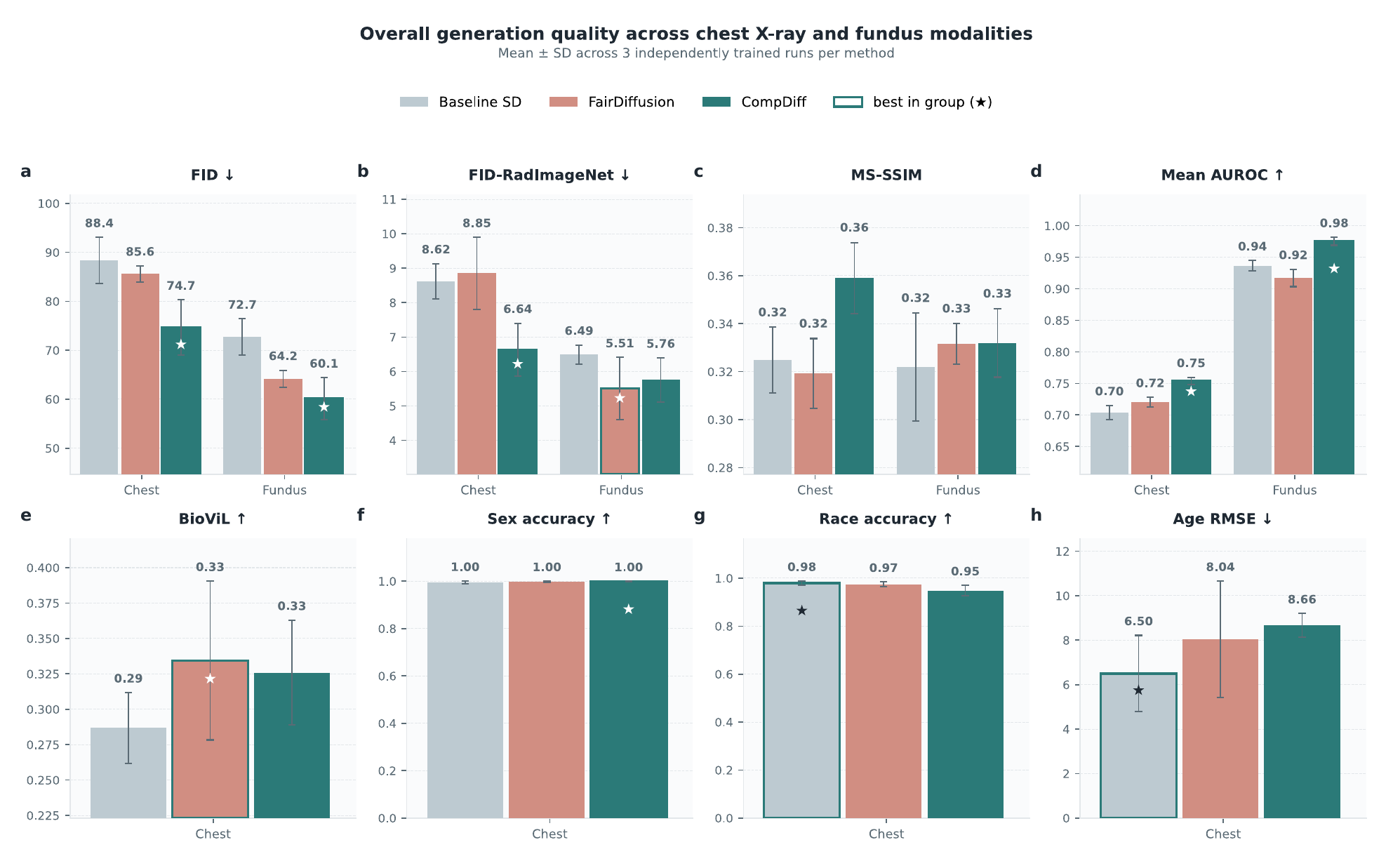}

\end{minipage}
\end{adjustbox}
\par\DisplayMark{fig:overall/caption-top}
\caption{Overall generation quality across chest X-ray and fundus modalities. \textbf{a}--\textbf{d}, Metrics computed for both modalities, each panel showing a chest group and a fundus group: \textbf{a}, FID; \textbf{b}, FID-RadImageNet; \textbf{c}, MS-SSIM; \textbf{d}, mean AUROC. \textbf{e}--\textbf{h}, Agreement with the requested findings and demographic attributes for chest images: \textbf{e}, BioViL; \textbf{f}, sex accuracy; \textbf{g}, race accuracy; \textbf{h}, age RMSE. Each group contains the three methods (Baseline SD, FairDiffusion, CompDiff). A star and a green outline mark the best method within a group, and an upward arrow in a panel title indicates that higher is better and a downward arrow that lower is better; MS-SSIM carries neither, because neither tail is desirable and values are read against the defensible band $\numMainFiguresK{}$--$\numMainFiguresL{}$.  Chest and Fundus bars are the mean over three independently trained runs per method, evaluated on the corresponding evaluation split, with error bars giving the standard deviation across those runs. Demographic and vision--language classifiers trained on chest images are out of domain on fundus images and are therefore reported for chest only.}
\label{fig:overall}
\par\DisplayMark{fig:overall/caption-bottom}
\end{figure*}

CompDiff achieved the lowest mean Fr\'echet Inception Distance (FID) on both modalities (Figure~\ref{fig:overall}). FID measures the distance between real and synthetic image feature distributions, with lower values indicating closer agreement. Values are means $\pm$ standard deviations across three independently trained runs per method and modality. On chest X-rays, CompDiff reached $\chestCDFID \pm \chestCDFIDSD$, compared with $\chestBaseFID \pm \chestBaseFIDSD$ for RoentGen-v2 and $\chestFairDFID \pm \chestFairDFIDSD$ for FairDiffusion, a $\numResultsA{}\%$ reduction relative to RoentGen-v2. On fundus images, the corresponding values were $\fundusCDFID \pm \fundusCDFIDSD$, $\fundusBaseFID \pm \fundusBaseFIDSD$ and $\fundusFairDFID \pm \fundusFairDFIDSD$, a $\numResultsB{}\%$ reduction relative to RoentGen-v2.

\begin{figure*}[tp]
\centering
\DisplayMark{fig:umap/plot-top}
\begin{adjustbox}{max totalsize={\textwidth}{0.67\textheight},center}
\begin{minipage}{\textwidth}
\centering

\includegraphics[width=\textwidth]{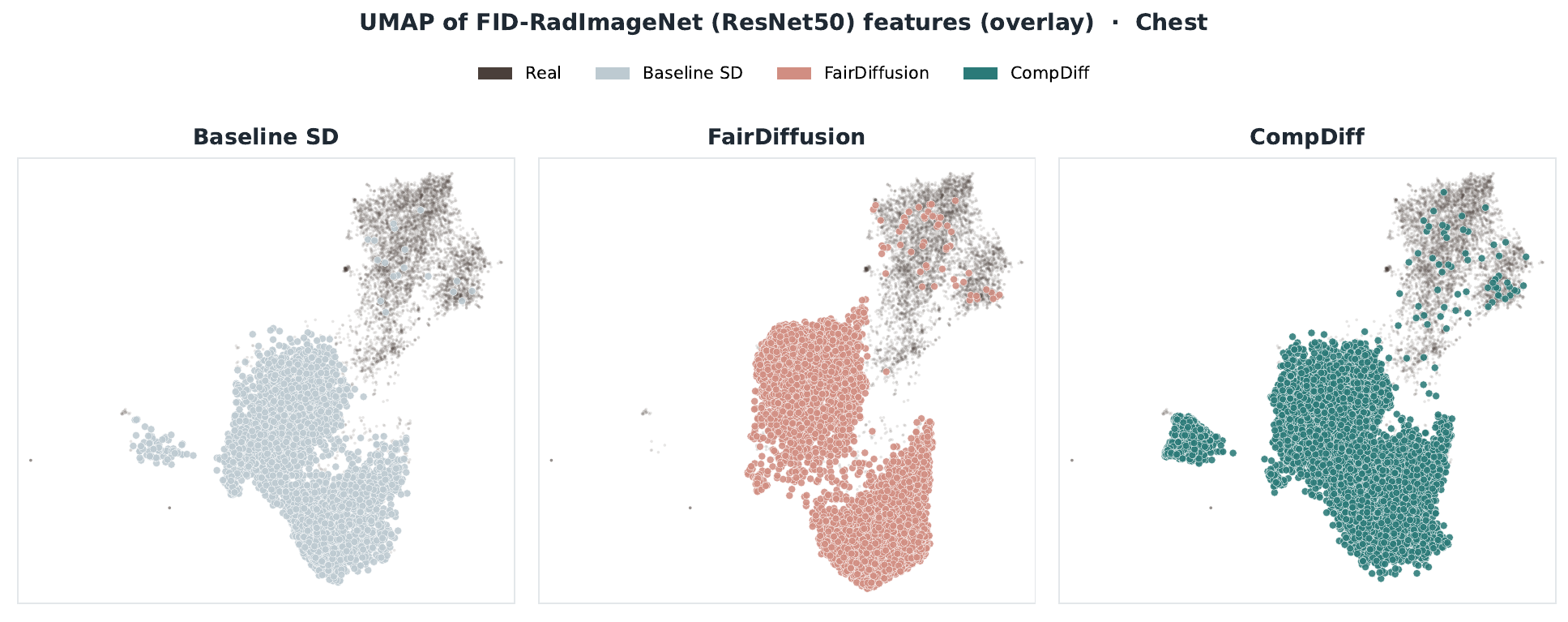}

\end{minipage}
\end{adjustbox}
\par\DisplayMark{fig:umap/caption-top}
\caption{Joint UMAP projection of chest X-ray RadImageNet features. One UMAP was fitted to pooled real and synthetic ResNet-\numMainFiguresM{} features, giving all panels shared axes. The same real cohort appears in warm grey beneath each generator's samples: RoentGen-v2 (Baseline SD, light grey), FairDiffusion (red) and CompDiff (green). CompDiff shows greater overlap with the cloud of real images. This projection is qualitative; Precision, Recall, Density and Coverage in Table~\ref{tab:overall-pr} quantify the distributions in the original feature space.}
\label{fig:umap}
\par\DisplayMark{fig:umap/caption-bottom}
\end{figure*}

The chest ordering persisted with features from RadImageNet, pretrained on radiological images~\cite{mei2022radimagenet}: FID-RadImageNet was $\chestCDFIDRN$ for CompDiff, $\chestBaseFIDRN$ for RoentGen-v2 and $\chestFairDFIDRN$ for FairDiffusion. On fundus, FairDiffusion had the lowest FID-RadImageNet ($\fundusFairDFIDRN$ versus $\fundusCDFIDRN$ for CompDiff). We therefore distinguish the two feature spaces and use Inception FID as the primary fundus measure, since RadImageNet is out of domain for fundus photography.

\paragraph{Agreement with the requested findings and attributes.}
Pretrained classifiers assessed whether generated images expressed their conditioning labels. Chest disease alignment was the mean AUROC over five prompted findings: atelectasis, cardiomegaly, oedema, pneumothorax and pleural effusion. CompDiff scored $\chestCDAUROC$, compared with $\chestFairDAUROC$ for FairDiffusion and $\chestBaseAUROC$ for RoentGen-v2. Fundus alignment used separate classifiers for glaucoma and cup-disc ratio. CompDiff had the highest average score ($\fundusCDAUROC$ versus $\fundusFairDAUROC$ and $\fundusBaseAUROC$), with the best glaucoma AUROC in every training run ($\fundusCDGlaucoma$ versus $\fundusFairDGlaucoma$ and $\fundusBaseGlaucoma$) and a larger margin in cup-disc alignment ($\fundusCDCDR$ versus $\fundusFairDCDR$ and $\fundusBaseCDR$).

Sex accuracy was near ceiling for all methods. CompDiff had lower race accuracy ($\chestCDRace$ versus $\chestBaseRace$ and $\chestFairDRace$) and higher age error ($\chestCDAge$ versus $\chestBaseAge$ and $\chestFairDAge$ years RMSE) than the prompt-conditioned baselines. Neither difference was statistically distinguishable across the three training runs ($p=\numResultsF{}$ for race and $p=\numResultsG{}$ for age, Welch's tests). These marginal alignment measures complement the intersectional fidelity and evaluations at unseen intersections below.

\paragraph{Distributional coverage and supporting checks.}
CompDiff also reached more of the real chest image distribution. The joint UMAP projection in Figure~\ref{fig:umap} shows greater overlap with the cloud of real images; this is a qualitative view of the features, rather than a direct display of FID. Quantitative metrics in the original RadImageNet feature space supported the same finding (Table~\ref{tab:overall-pr}). CompDiff had the highest Precision, the fraction of synthetic samples within the neighbourhoods of real images, and Coverage, the fraction of neighbourhoods of real images reached by synthetic samples~\cite{kynkaanniemi2019precision,naeem2020reliable}. RoentGen-v2 had the highest Recall, a related measure that also depends on the spread of neighbourhoods of synthetic images.

\begin{table*}[!tp]
\centering
\caption{Distributional fidelity and coverage on the full chest X-ray test set with its natural demographic distribution, computed in RadImageNet ResNet-50 feature space with $k=10$ manifold neighbours. Higher is better for all four metrics. Precision~\cite{kynkaanniemi2019precision} and Density~\cite{naeem2020reliable} measure \emph{fidelity} (how often synthetic samples fall inside the real manifold), with Density a refinement of Precision that is less sensitive to outliers. Recall~\cite{kynkaanniemi2019precision} and Coverage~\cite{naeem2020reliable} measure \emph{diversity} (how much of the real manifold is reached by synthetic samples), with Coverage the replacement for Recall that is robust to mode collapse (uses manifold radii measured on real images rather than synthetic images, so it is not inflated by synthetic outliers).}
\label{tab:overall-pr}
\noindent{\fontsize{8}{9.6}\selectfont
\begin{adjustbox}{max width=\textwidth}\begin{tabular}{lcccc}
\toprule
\textbf{Method} & \textbf{Precision} $\uparrow$ & \textbf{Recall} $\uparrow$ & \textbf{Density} $\uparrow$ & \textbf{Coverage} $\uparrow$ \\
\midrule
RoentGen-v2    & 0.329 & \textbf{0.480} & 0.072 & 0.084 \\
FairDiffusion  & 0.344 & 0.234          & 0.070 & 0.099 \\
\textbf{CompDiff} & \textbf{0.396} & 0.387 & \textbf{0.091} & \textbf{0.139} \\
\bottomrule
\end{tabular}\end{adjustbox}
}
\end{table*}

Paired real--synthetic MS-SSIM fell within the reference range $\numResultsC{}$--$\numResultsD{}$ for every method and did not distinguish them~\cite{moroianu2025improvingperformancerobustnessfairness}. Chest BioViL similarity was nearly identical for CompDiff and FairDiffusion ($\chestCDBioViL$ and $\chestFairDBioViL$), both above RoentGen-v2 ($\chestBaseBioViL$). Authenticity, which compares synthetic--real proximity with distances between neighbouring real images, was at least $\numResultsK{}$ for all three generators~\cite{alaa2022faithful}.

\FloatBarrier
\subsection{Subgroup generation quality}\label{subsec:fairness}

\begin{figure*}[tp]
\centering
\DisplayMark{fig:intersectional/plot-top}
\begin{adjustbox}{max totalsize={\textwidth}{0.67\textheight},center}
\begin{minipage}{\textwidth}
\centering

\includegraphics[width=\textwidth]{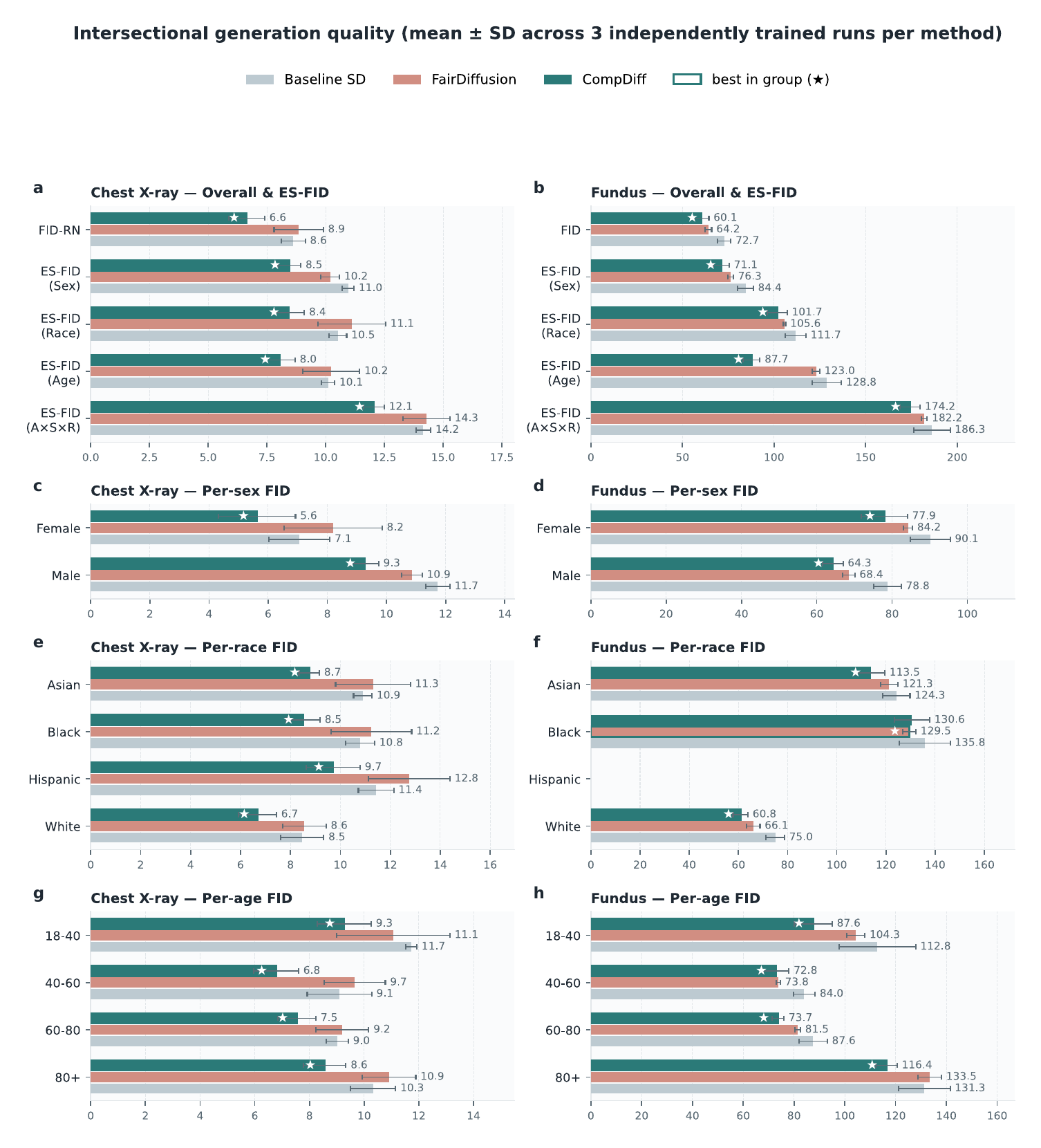}

\end{minipage}
\end{adjustbox}
\par\DisplayMark{fig:intersectional/caption-top}
\caption{Intersectional generation quality (lower is better). Left column (\textbf{a}, \textbf{c}, \textbf{e}, \textbf{g}): chest X-ray; right column (\textbf{b}, \textbf{d}, \textbf{f}, \textbf{h}): fundus. \textbf{a}, \textbf{b}, Overall distance and equity-scaled distance (ES-FID) across sex, race, age, and the full age $\times$ sex $\times$ race intersection (A$\times$S$\times$R), where the intersectional ES-FID uses every demographic cell as the subgroup list. \textbf{c}, \textbf{d}, By sex. \textbf{e}, \textbf{f}, By race. \textbf{g}, \textbf{h}, Per age bin. Each row groups three methods (Baseline SD, FairDiffusion, CompDiff), and a star marks the best method per row. The chest column reports FID-RadImageNet and the fundus column reports Inception FID, following the rationale for the feature backbone of Section~\ref{subsec:eval-synth}. Bars are means over three independently trained runs per method, with error bars giving the standard deviation across those runs. The fundus intersectional ES-FID is dominated by its dispersion term, because cell distances at the small fundus cell sizes are several times the overall distance; it is reported for completeness. The Hispanic row is omitted in panel \textbf{f} because the FairGenMed test set does not include Hispanic patients.}
\label{fig:intersectional}
\par\DisplayMark{fig:intersectional/caption-bottom}
\end{figure*}

CompDiff's fidelity gains extended across sex, race and age groups on both modalities (Figure~\ref{fig:intersectional}), measured with FID-RadImageNet for chest X-rays and Inception FID for fundus images. It achieved the lowest distance in \figbWinsTotal{} of \numResultsL{} overall, equity-scaled and subgroup comparisons. The exception was the Black fundus stratum, where FairDiffusion's mean was lower by $\numResultsQ{}$ FID.

Equity-scaled FID (ES-FID) captures overall fidelity together with variation between subgroups. CompDiff had the lowest ES-FID across sex, race, age and their joint intersection on both modalities. On chest X-rays, intersectional ES-FID was $\chestCDESFIDIntRN$, compared with $\chestBaseESFIDIntRN$ for RoentGen-v2 and $\chestFairDESFIDIntRN$ for FairDiffusion. On fundus, the intersectional measure was dominated by dispersion at small cell sizes, so the measures for each attribute provide the more reliable comparison.

Comparisons within cells were limited by sample size. Among chest X-ray intersections with at least twenty images, CompDiff had the lowest FID-RadImageNet in all \numResultsV{} cells, reducing cell distance by $\numResultsW{}\%$ relative to RoentGen-v2 and $\numResultsX{}\%$ relative to FairDiffusion (paired comparisons, $p < \numResultsY{}$ for both). The methods could not be distinguished in the more sparsely sampled cells. Restricting the aggregate equity analysis to the adequately sampled cells preserved the method ordering.

\subsection{Generation at unseen intersections}\label{subsec:zeroshot}

\begin{figure*}[tp]
\centering
\DisplayMark{fig:zeroshot/plot-top}
\includegraphics[width=\textwidth]{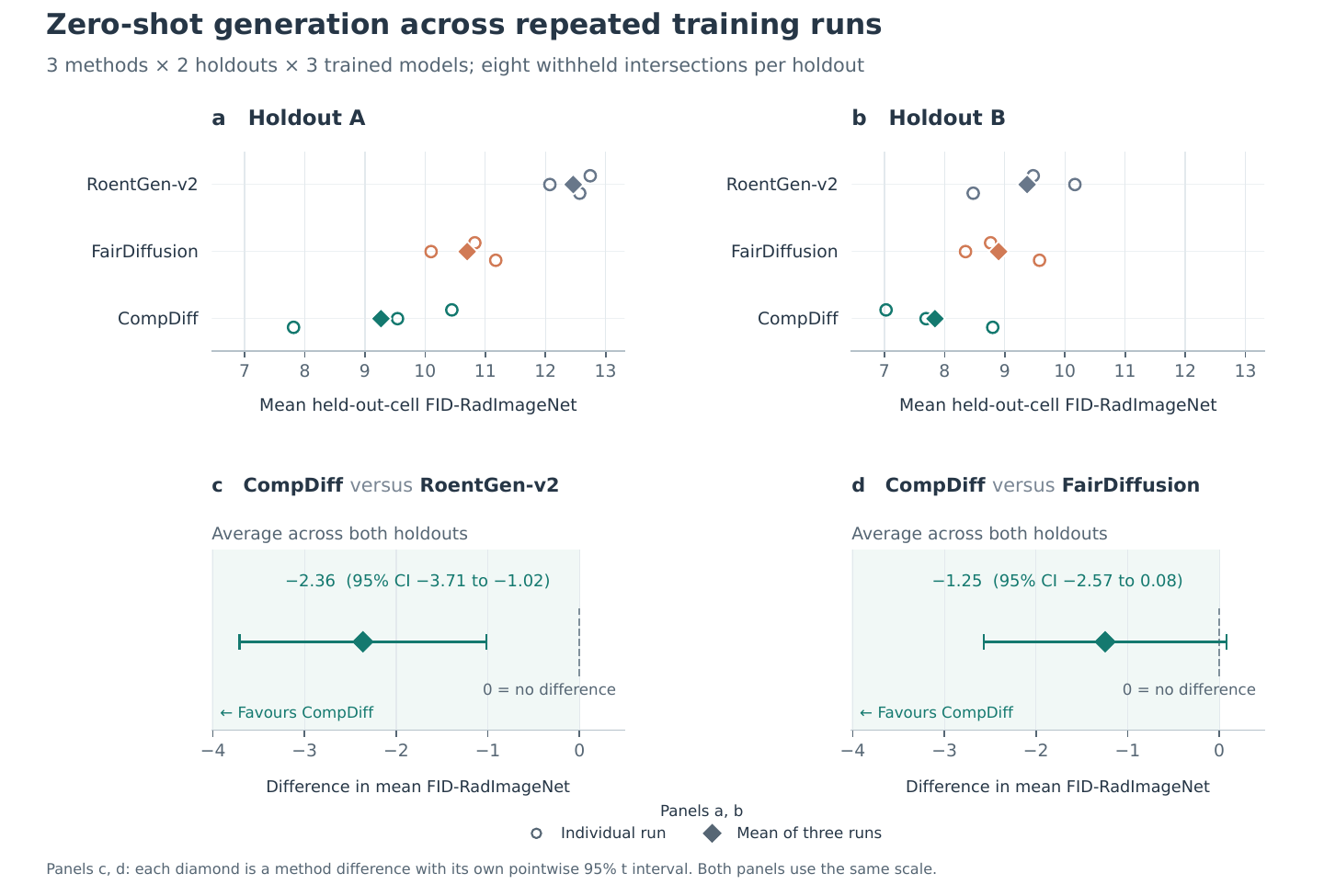}
\par\DisplayMark{fig:zeroshot/caption-top}
\caption{Zero-shot generation across repeated training runs. Each method was trained three times on each of two holdouts, each withholding eight demographic intersections. FID-RadImageNet measures the distance between real and synthetic feature distributions within an intersection; lower values indicate closer agreement. \textbf{a,b}, Mean distance across the eight withheld cells for each trained model (open circles), with the mean across three models shown as a filled diamond. Runs are not paired by seed number. \textbf{c,d}, CompDiff minus comparator differences, averaged equally across cells and holdouts. Each panel shows one difference and its pointwise $\scLevel\%$ Welch--Satterthwaite confidence interval, with a separate zero reference and a common scale. Negative differences favour CompDiff. Each comparison uses twelve trained models; the experiment contains $\scModels$ models in total. Intervals describe variation across training, checkpoint selection and generation on the fixed test cohorts. Cell observations and means appear in Supplementary Figure~\ref{fig:zeroshot-cells}.}
\label{fig:zeroshot}
\par\DisplayMark{fig:zeroshot/caption-bottom}
\end{figure*}

To test whether CompDiff could compose demographic intersections absent from training, we removed eight intersections from each of two MIMIC-CXR training variants. Holdout~A excluded the eight rarest intersections; Holdout~B excluded the eight next rarest. Each method was trained three times per holdout, giving $\scModels$ models. FID-RadImageNet compared generated images with real test images from the same withheld intersection, measuring fidelity within each demographic cell. Figure~\ref{fig:zeroshot} summarises the models' mean distances across their eight withheld cells and the average differences between methods.

CompDiff achieved the lowest mean FID-RadImageNet across three runs in all $\scRNWins$ withheld intersections (Supplementary Figure~\ref{fig:zeroshot-cells}; Supplementary Table~\ref{tab:seed-cells}). Averaged equally across cells and holdouts, the CompDiff minus RoentGen-v2 difference was $\scRNBaseEffect$, with a pointwise $\scLevel\%$ confidence interval of $[\scRNBaseLo,\scRNBaseHi]$. The corresponding difference from FairDiffusion was $\scRNFairEffect$ $[\scRNFairLo,\scRNFairHi]$. These intervals use independently trained models as the replicates. Both average differences remained in favour of CompDiff after omitting any single trained model. With Inception features, CompDiff retained the lowest mean across three runs in $\scFIDWins$ of $\scCells$ cells (Supplementary Figures~\ref{fig:zeroshot-inception} and~\ref{fig:zeroshot-cells-inception}; Supplementary Note~\ref{app:zeroshot-inception}).

Complementary controls tested whether sample size or the choice of fidelity measure changed the comparisons within each cell. Because limited numbers of real and synthetic images can inflate FID (Supplementary Figure~\ref{fig:estimator-size-curve}), we adjusted for this estimated bias. The corrected FID preserved the winning configuration in $\robWinnerInfJ$ of $\robCells$ evaluated cells, including both withheld intersections and those represented in training. Kernel Inception Distance (KID) measures distributional similarity using an unbiased estimator, avoiding the systematic bias from limited sample sizes that affects FID~\cite{binkowski2018mmdgans}. Computed from the same RadImageNet features, KID preserved the winning configuration in $\robWinnerKID$ of $\robCells$ cells. Normalisation to a common sample size supported the same comparisons. Among the $\robHOCells$ eligible withheld cells, corrected FID preserved the complete configuration ordering in all $\robHOOrderInfJ$, and KID in $\robHOOrderKID$. The four smallest Holdout~A cells, with fewer than $\robSelFloor$ images, were excluded (Supplementary Note~\ref{app:estimator-robustness}).

Additional sampling did not close the gap between RoentGen-v2 and the CompDiff variants. Giving RoentGen-v2 a $\robSelBudget$ generation budget and retaining its best quarter under a verifier still left either CompDiff or its unstructured composer variant as the best configuration in each of the $\robSelCondWins$ assessable withheld cells (Supplementary Note~\ref{app:selection-control}). The unstructured variant matched CompDiff's parameter count and retained its separate attribute encoders and supervised output tokens. It preserved fidelity within each cell on both holdouts, showing that generation at unseen intersections did not depend on the factorised composer (Supplementary Figure~\ref{fig:composer-zeroshot}; Section~\ref{subsec:ablations}). Together, these controls support the interface across the tested estimators, sampling budgets and composer choices.

\begin{figure*}[tp]
\centering
\DisplayMark{fig:qualitative_grid/plot-top}
\begin{adjustbox}{max totalsize={\textwidth}{0.67\textheight},center}
\begin{minipage}{\textwidth}
\centering

\includegraphics[width=\textwidth]{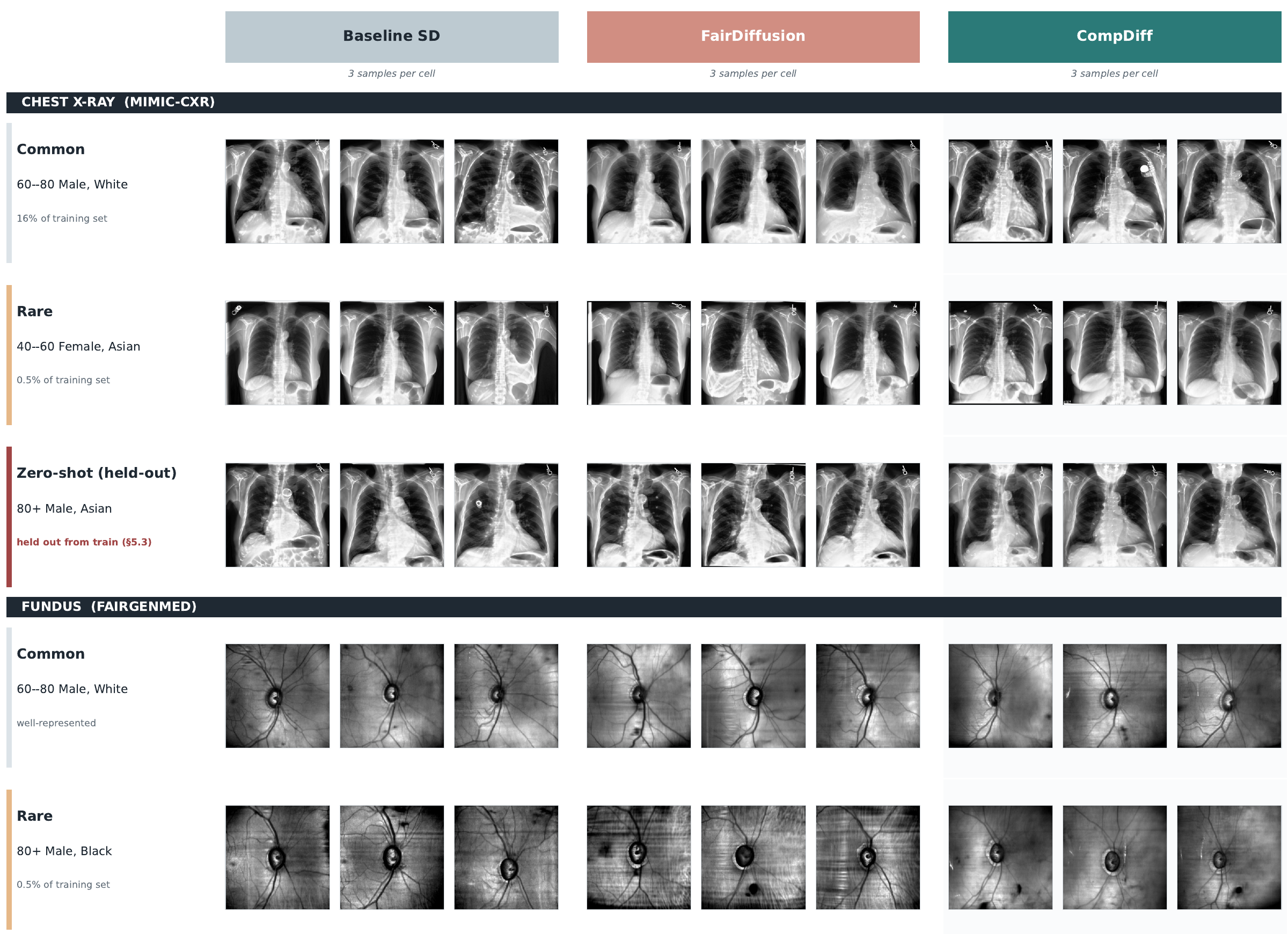}

\end{minipage}
\end{adjustbox}
\par\DisplayMark{fig:qualitative_grid/caption-top}
\caption{Qualitative samples across tiers of subgroup rarity on both modalities. \textbf{Chest X-ray (MIMIC-CXR), top three rows}: \emph{Common} (\numMainFiguresU{}--\numMainFiguresV{} Male, White; $\numMainFiguresW{}\%$ of training), \emph{Rare} (\numMainFiguresX{}--\numMainFiguresY{} Female, Asian; $\numMainFiguresZ{}\%$ of training), and \emph{Zero-shot (withheld)} (\numMainFiguresAA{}+ Male, Asian; one of the eight intersections held out from training in the zero-shot protocol of Section~\ref{subsec:zeroshot-protocol}). \textbf{Fundus (FairGenMed), bottom two rows}: \emph{Common} (\numMainFiguresAB{}--\numMainFiguresAC{} Male, White) and \emph{Rare} (\numMainFiguresAD{}+ Male, Black). The two Common chest/fundus rows, the Rare chest row, and the Rare fundus row are sampled from generators trained on the natural demographic distribution; the Zero-shot chest row is sampled from the generators of Section~\ref{subsec:zeroshot-protocol}, trained with these intersections held out, so that row probes whether each generator can synthesise a demographic intersection it has never observed in training. Within each row, three samples with matched prompts are shown per method (Baseline SD, FairDiffusion, CompDiff); samples in the same column position across the three methods share the same clinical prompt, allowing visual comparison under the same clinical conditioning. The Zero-shot chest row prompts (all \numMainFiguresAE{}+ Male, Asian) are, left to right: a new large mass in the left upper lobe with rightward tracheal deviation concerning for malignancy; moderate cardiomegaly without pulmonary oedema or pleural effusion; and right basal atelectasis without pneumothorax.}
\label{fig:qualitative_grid}
\par\DisplayMark{fig:qualitative_grid/caption-bottom}
\end{figure*}

Figure~\ref{fig:qualitative_grid} shows samples with matched prompts from common and rare groups in both modalities, together with a withheld chest intersection (\numResultsAB{}+ Male Asian). Each column uses the same prompt across generators. The reader study below provides an independent assessment of realism and clinical correspondence on the unseen chest intersections.

\subsection{Radiologist assessment of unseen intersections}\label{subsec:reader-study}

In blinded assessment by two radiologists, CompDiff achieved the highest mean realism and prompt correspondence scores and was selected most often as the most realistic synthetic image from unseen demographic intersections (Table~\ref{tab:reader-study}; protocol in Section~\ref{subsec:reader-protocol}; rating interface in Supplementary Figure~\ref{fig:reader-interface}). The readers completed $\rsNInstances$ assessments of sets of three images spanning the $\numMethodsEK{}$ withheld intersections, covering $\rsNSets$ distinct sets. Mean prompt correspondence with the clinical impression ($-\numMethodsEN{}$ to $+\numMethodsEO{}$ scale) was $+\numResultsAH{}$ for CompDiff, compared with $+\numResultsAI{}$ for FairDiffusion and $-\numResultsAJ{}$ for RoentGen-v2. Prompt correspondence was higher for CompDiff than for both RoentGen-v2 and FairDiffusion in the mixed-effects analysis, with Benjamini--Hochberg-adjusted $p$-values of $\readerCorrBaseBH$ and $\readerCorrFairBH$, respectively.

\begin{table*}[!htp]
\centering
\caption{Blinded radiologist assessment of chest X-ray sources on the zero-shot withheld intersections, by two board-certified radiologists ($\rsNInstances$ rater--set instances over $\rsNSets$ distinct sets, $\rsNRatings$ image ratings, across both withheld splits). Every rated image is of a demographic intersection absent from generator training. Realism is scored $1$--$5$ and prompt correspondence from $-2$ to $+2$ (mean $\pm$ SD; unreadable images excluded from realism, unreadable and N/A from correspondence). ``Most realistic'' is the per-set forced-choice win rate: the fraction of rated sets containing that source in which it was picked as the most realistic image. Real images are a hidden upper anchor present in a subset of sets.}
\label{tab:reader-study}
\noindent{\fontsize{8}{9.6}\selectfont
\begin{tabular}{lccc}
\toprule
\textbf{Source} & \textbf{Realism} $\uparrow$ & \textbf{Prompt corr.} $\uparrow$ & \textbf{Most realistic} $\uparrow$ \\
 & \footnotesize (1--5) & \footnotesize ($-2$ to $+2$) & \footnotesize win rate \\
\midrule
\textbf{CompDiff} & \textbf{4.17} $\pm$ 1.11 & \textbf{+0.47} $\pm$ 1.72 & \textbf{48\%} \\
RoentGen-v2 & 2.36 $\pm$ 1.33 & -0.05 $\pm$ 1.69 & 7\% \\
FairDiffusion & 4.13 $\pm$ 1.05 & +0.01 $\pm$ 1.74 & 37\% \\
Real \footnotesize(hidden anchor) & 4.92 $\pm$ 0.29 & +0.75 $\pm$ 1.86 & 92\% \\
\bottomrule
\end{tabular}

}
\end{table*}

Anatomical realism ranged from $\numResultsAD{}$ (clearly artefactual) to $\numResultsAE{}$ (indistinguishable from a real scan). Mean scores were $\numResultsAF{}$ for CompDiff, $\numResultsAG{}$ for FairDiffusion and $\numResultsAC{}$ for RoentGen-v2 (Section~\ref{app:reader-stats}). Readers selected CompDiff as the most realistic image in $\rsFCCompDiff$ of the sets in which it appeared, compared with $\rsFCFairD$ for FairDiffusion and $\rsFCBase$ for RoentGen-v2. The direct preference comparison with FairDiffusion also favoured CompDiff numerically, with $\numResultsAK{}$ selections to $\numResultsAL{}$ ($p = \numResultsAM{}$; Supplementary Table~\ref{tab:reader-contrasts}). Hidden real images provided an upper reference, with mean realism $\numResultsAN{}$ and $\rsFCRealWins$ selections as most realistic in $\rsFCRealN$ appearances.

\FloatBarrier
\subsection{Downstream classification}\label{subsec:downstream}

\begin{figure*}[tp]
\centering
\DisplayMark{fig:downstream_combined/plot-top}
\begin{adjustbox}{max totalsize={\textwidth}{0.67\textheight},center}
\begin{minipage}{\textwidth}
\centering

\includegraphics[width=\textwidth]{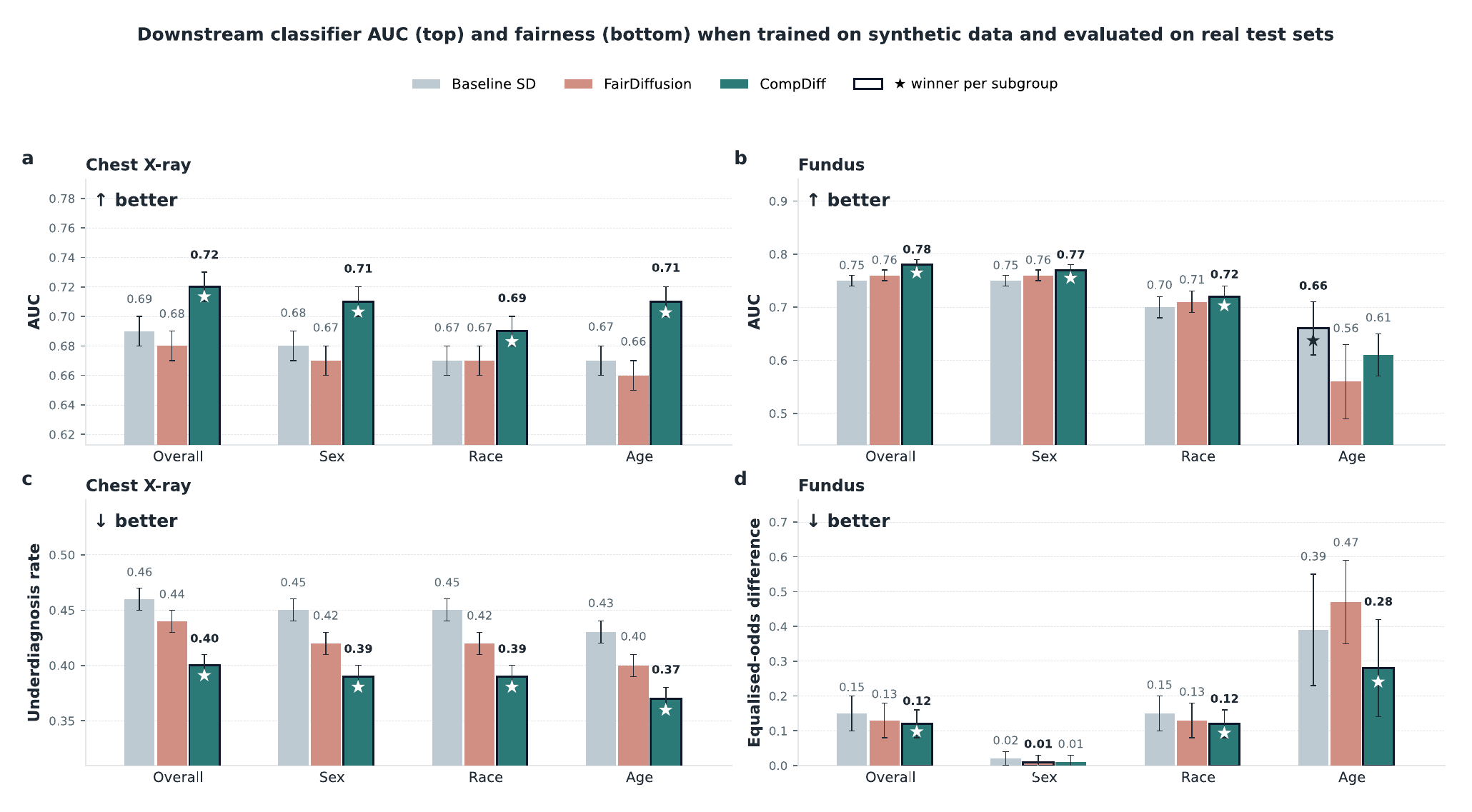}

\end{minipage}
\end{adjustbox}
\par\DisplayMark{fig:downstream_combined/caption-top}
\caption{Downstream classifier performance when trained on synthetic data and evaluated on real test sets. \textbf{a}, \textbf{b}, AUC (higher is better) on chest X-ray (\textbf{a}) and fundus (\textbf{b}). \textbf{c}, \textbf{d}, Subgroup fairness (lower is better): underdiagnosis rate for chest (\textbf{c}) and equalised-odds difference for fundus (\textbf{d}). Each panel reports an overall result and separate summaries for sex, race and age and shows three methods (Baseline SD, FairDiffusion, CompDiff). Error bars are standard deviation across runs. The winning method per subgroup is outlined in dark slate and its label is bold. CompDiff is the winner on every chest X-ray (subgroup, metric) cell and on $\numMainFiguresAF{}/\numMainFiguresAG{}$ fundus cells.}
\label{fig:downstream_combined}
\par\DisplayMark{fig:downstream_combined/caption-bottom}
\end{figure*}

We first compared the utility of images from all three generators for classifier training on chest X-rays and fundus images, evaluating the trained classifiers on real test images. Classifiers trained on CompDiff images achieved higher diagnostic discrimination and lower underdiagnosis rates on real chest X-rays across the overall, sex, race and age evaluations (Figure~\ref{fig:downstream_combined}). Overall AUC was $\numResultsAO{}$, compared with $\numResultsAP{}$ for RoentGen-v2 and $\numResultsAQ{}$ for FairDiffusion; underdiagnosis rates were $\numResultsAU{}$, $\numResultsAV{}$ and $\numResultsAW{}$, respectively. Underdiagnosis measures how often an image with disease is classified as having no finding. The sex, race and age panels summarise performance along each demographic attribute. On fundus, CompDiff achieved the highest AUC in $\numResultsBA{}$ of $\numResultsBB{}$ comparisons. The exception was the age evaluation, where RoentGen-v2 had higher AUC ($\numResultsBE{}$ versus $\numResultsBF{}$), while CompDiff had the lowest equalised-odds difference ($\numResultsBG{}$, compared with $\numResultsBH{}$ for RoentGen-v2 and $\numResultsBI{}$ for FairDiffusion). Equalised-odds difference measures the larger of the gaps between groups in true positive and false positive rates, with lower values indicating smaller disparities.

\begin{figure*}[!htbp]
\centering
\DisplayMark{fig:downstream-d3deep/plot-top}
\begin{adjustbox}{max totalsize={\textwidth}{.67\textheight},center}
\includegraphics[width=\textwidth]{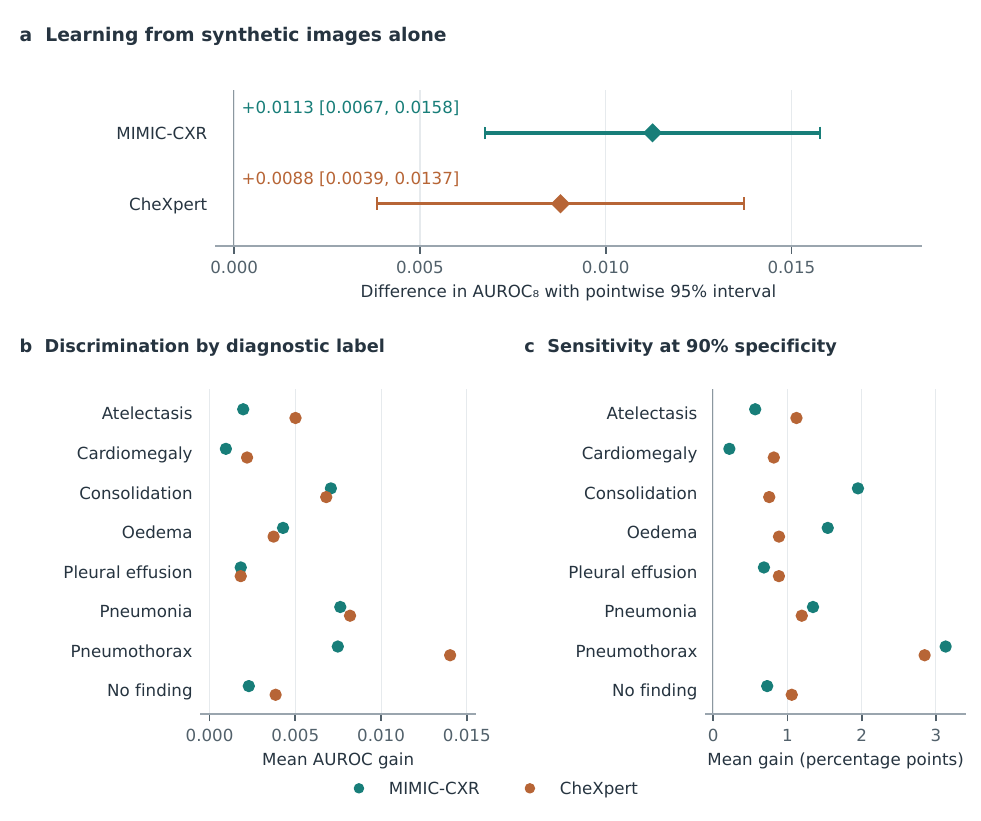}
\end{adjustbox}
\par\DisplayMark{fig:downstream-d3deep/caption-top}
\caption{Diagnostic utility of CompDiff images relative to RoentGen-v2. All panels use the same eight diagnostic labels. \textbf{(a)} AUROC$_8$ difference for classifiers trained on synthetic chest images alone. Diamonds show the contrast averaged equally across four fixed corpus regimes; bars are pointwise 95\% Welch confidence intervals based on four independent classifier runs per generator and regime, conditional on the corpora and real test cohorts. \textbf{(b)} Mean AUROC difference for each label across all 48 matched configurations with pretraining followed by fine-tuning. \textbf{(c)} Mean sensitivity difference at 90\% specificity over the same configurations. Sensitivity is read separately from each test ROC for this exploratory comparison; it does not represent a validated clinical threshold. Panels (b) and (c) are descriptive averages over configurations that share training data and test cohorts. All differences are CompDiff minus RoentGen-v2; positive values favour CompDiff. Colours identify the test dataset. The same trained models are evaluated on both datasets.}
\label{fig:downstream-d3deep}
\par\DisplayMark{fig:downstream-d3deep/caption-bottom}
\end{figure*}

\paragraph{Gains across training configurations and diagnostic labels.}
A separate chest X-ray experiment compared CompDiff with RoentGen-v2 across independent classifier runs. The four synthetic corpora followed the natural demographic distribution at $\numMethodsCP{}\times$, $\numMethodsCQ{}\times$ and $\numMethodsCR{}\times$ the real training set size, or approximately balanced demographic intersections. When synthetic images were the sole source of chest training data, four classifier runs per corpus gave mean AUROC$_8$ values of $\dsSynthMimicComp$ for CompDiff versus $\dsSynthMimicBase$ for RoentGen-v2 on MIMIC-CXR, and $\dsSynthChexComp$ versus $\dsSynthChexBase$ on external CheXpert. AUROC$_8$ averages the same eight diagnostic labels on both datasets. Weighting the four corpora equally, the differences were $+\dsSynthMimicGain$ and $+\dsSynthChexGain$, with pointwise 95\% confidence intervals of $[\dsSynthMimicLo,\dsSynthMimicHi]$ and $[\dsSynthChexLo,\dsSynthChexHi]$, respectively (Figure~\ref{fig:downstream-d3deep}a). CompDiff had the higher mean in every corpus regime on both datasets (Supplementary Figure~\ref{fig:downstream-synthetic-runs}).

\begin{table*}[!hbp]
\centering
\caption{Downstream classifier training recipes. Each recipe is evaluated for both CompDiff and the RoentGen-v2 baseline. The natural synthetic corpora match the demographic distribution of the MIMIC-CXR training split, while the balanced corpus is expanded to approximately uniform coverage of $(\text{age}, \text{sex}, \text{race})$ cells. The pretrain $\rightarrow$ real fine-tune family is additionally swept over six fractions of real data (1\%, 10\%, 30\%, 50\%, 70\%, 100\%), giving 48 matched configurations per evaluation dataset. Balanced-corpus counts are $\dsBalancedRG$ for RoentGen-v2 and $\dsBalancedCD$ for CompDiff.}
\label{tab:recipes}
\noindent{\fontsize{8}{9.6}\selectfont
\begin{adjustbox}{max width=\textwidth}\begin{tabular}{@{} l l r l @{}}
\toprule
\textbf{Family / regime} & \textbf{Pretraining corpus} & \textbf{Images} & \textbf{Initialisation} \\
\midrule
\multicolumn{4}{@{}l}{\textit{Training on synthetic data alone (no real chest images)}} \\
$1\times$ natural   & \textendash{} & 62{,}094  & ImageNet \\
$2\times$ natural   & \textendash{} & 124{,}188 & ImageNet \\
$3\times$ natural   & \textendash{} & 186{,}282 & ImageNet \\
Balanced     & \textendash{} & $\approx415{,}000$ & ImageNet \\
\midrule
\multicolumn{4}{@{}l}{\textit{Pretrain $\rightarrow$ real fine-tune (FT fractions 1\%--100\%)}} \\
$1\times$ natural   & 62{,}094 synthetic     & 62{,}094  & Random; ImageNet \\
$2\times$ natural   & 124{,}188 synthetic    & 124{,}188 & Random; ImageNet \\
$3\times$ natural   & 186{,}282 synthetic    & 186{,}282 & Random; ImageNet \\
Balanced     & $\approx415{,}000$ synthetic    & $\approx415{,}000$ & Random; ImageNet \\
\bottomrule
\end{tabular}\end{adjustbox}
}
\end{table*}

The advantage extended to classifiers pretrained on synthetic images and then fine-tuned on real MIMIC-CXR data. Across the $\dsConfigurations$ matched configurations, CompDiff improved AUROC$_8$ in $\dsWinsMimic$ on MIMIC-CXR and $\dsWinsChex$ on CheXpert. This included $\dsScratchWinsMimic$ of 24 configurations with pretraining from scratch and $\dsImageNetWinsMimic$ of 24 with ImageNet initialisation on each dataset (complete sweep in Supplementary Figures~\ref{fig:downstream-sweep-mimic} and~\ref{fig:downstream-sweep-chexpert}). The comparisons used the matched AdamW recipe at $\dsSweepResolution{}\times\dsSweepResolution{}$ resolution (Table~\ref{tab:recipes}; Section~\ref{subsec:downstream-recipes}).

Mean AUROC gains were positive for all $\dsLabels$ diagnostic labels on both test sets (Figure~\ref{fig:downstream-d3deep}b). Pneumothorax showed the largest external gain, $+\dsPtxAurocChex$ AUROC on CheXpert. An exploratory comparison of the test ROC curves made the gain more interpretable: at 90\% specificity, mean pneumothorax sensitivity increased from $\dsPtxSensitivityMimicBase\%$ to $\dsPtxSensitivityMimicComp\%$ on MIMIC-CXR and from $\dsPtxSensitivityChexBase\%$ to $\dsPtxSensitivityChexComp\%$ on CheXpert (Figure~\ref{fig:downstream-d3deep}c). Mean sensitivity gains were positive for all eight labels. Supplementary Table~\ref{tab:downstream-labels} also reports precision--recall performance and the comparison at 95\% specificity.

\begin{figure*}[!htbp]
\centering
\DisplayMark{fig:pergroup-disparity/plot-top}
\begin{adjustbox}{max totalsize={\textwidth}{.67\textheight},center}
\includegraphics[width=\textwidth]{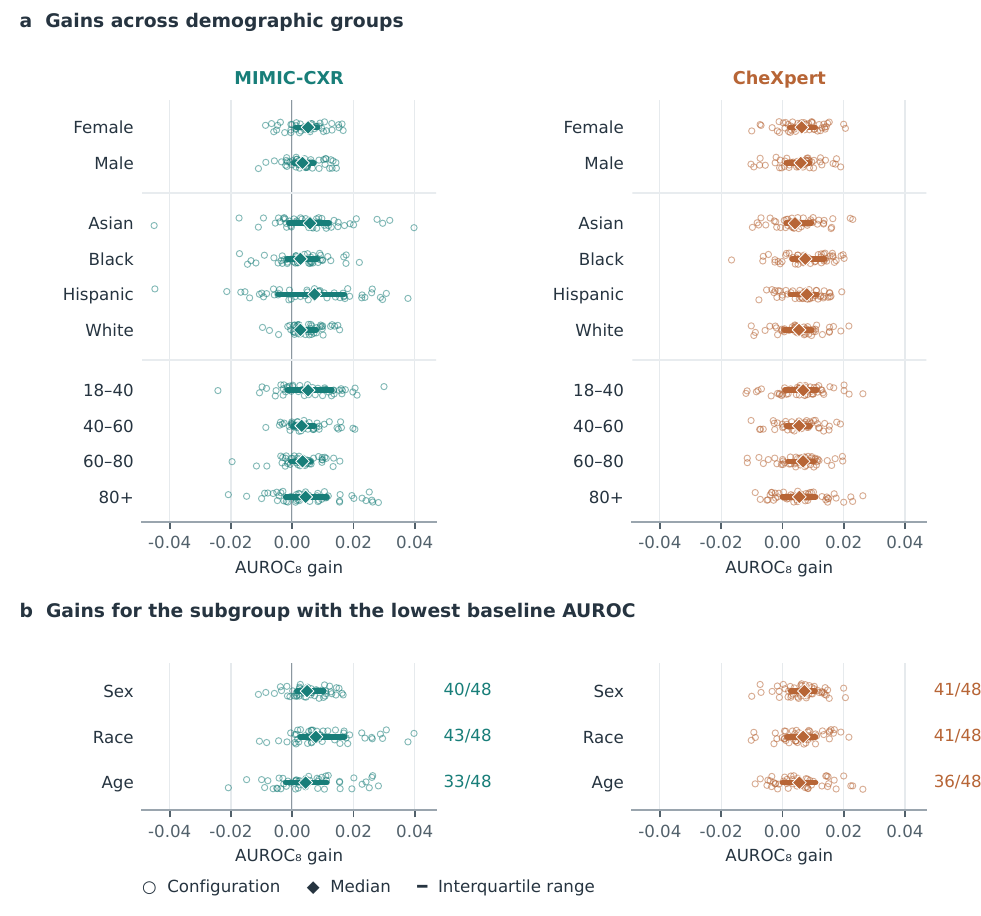}
\end{adjustbox}
\par\DisplayMark{fig:pergroup-disparity/caption-top}
\caption{Subgroup gains across the complete downstream training sweep. \textbf{(a)} CompDiff minus RoentGen-v2 AUROC$_8$ for each sex, race and age group, shown separately for MIMIC-CXR and CheXpert. \textbf{(b)} Gain for the subgroup with the lowest RoentGen-v2 AUROC within each demographic attribute in each configuration, retaining its identity under CompDiff. Hollow circles show the 48 configuration differences; filled diamonds and thick bars show medians and interquartile ranges. These ranges describe variation across configurations, not confidence intervals. Counts in (b) give configurations with a positive gain. Every subgroup score averages the same eight labels, each with at least one positive and one negative image. Configurations share training data and test cohorts. Sample sizes and uncertainty checks for sparse groups are in Supplementary Tables~\ref{tab:downstream-groups} and~\ref{tab:downstream-uncertainty}.}
\label{fig:pergroup-disparity}
\par\DisplayMark{fig:pergroup-disparity/caption-bottom}
\end{figure*}

\paragraph{CompDiff raises the floor on underserved subgroups.}
The overall gains extended across demographic groups. Across the $\dsConfigurations$ training configurations, median AUROC$_8$ gains were positive for all $\dsGroups$ sex, race and age groups separately on MIMIC-CXR and CheXpert (Figure~\ref{fig:pergroup-disparity}a). We then identified the subgroup with the lowest RoentGen-v2 AUROC within each demographic attribute and measured the CompDiff gain in that same subgroup. Median gains were positive for sex, race and age on both datasets (Figure~\ref{fig:pergroup-disparity}b). Across the two datasets, these gains were positive in $\dsWorstPercentSex\%$, $\dsWorstPercentRace\%$ and $\dsWorstPercentAge\%$ of configuration evaluations, respectively. The improvements therefore reached the groups with the lowest baseline performance as well as the overall average. Subgroup sample sizes and patient resampling checks for the sparsely sampled groups are reported in Supplementary Note~\ref{app:downstream-evidence}.

\FloatBarrier
\subsection{Synthetic cohorts for fairness auditing}\label{subsec:eval-instrument}

\begin{figure*}[p]
\centering
\DisplayMark{fig:audit-instrument/plot-top}
\begin{adjustbox}{max totalsize={\textwidth}{0.67\textheight},center}
\begin{minipage}{\textwidth}
\centering

\includegraphics[width=\textwidth]{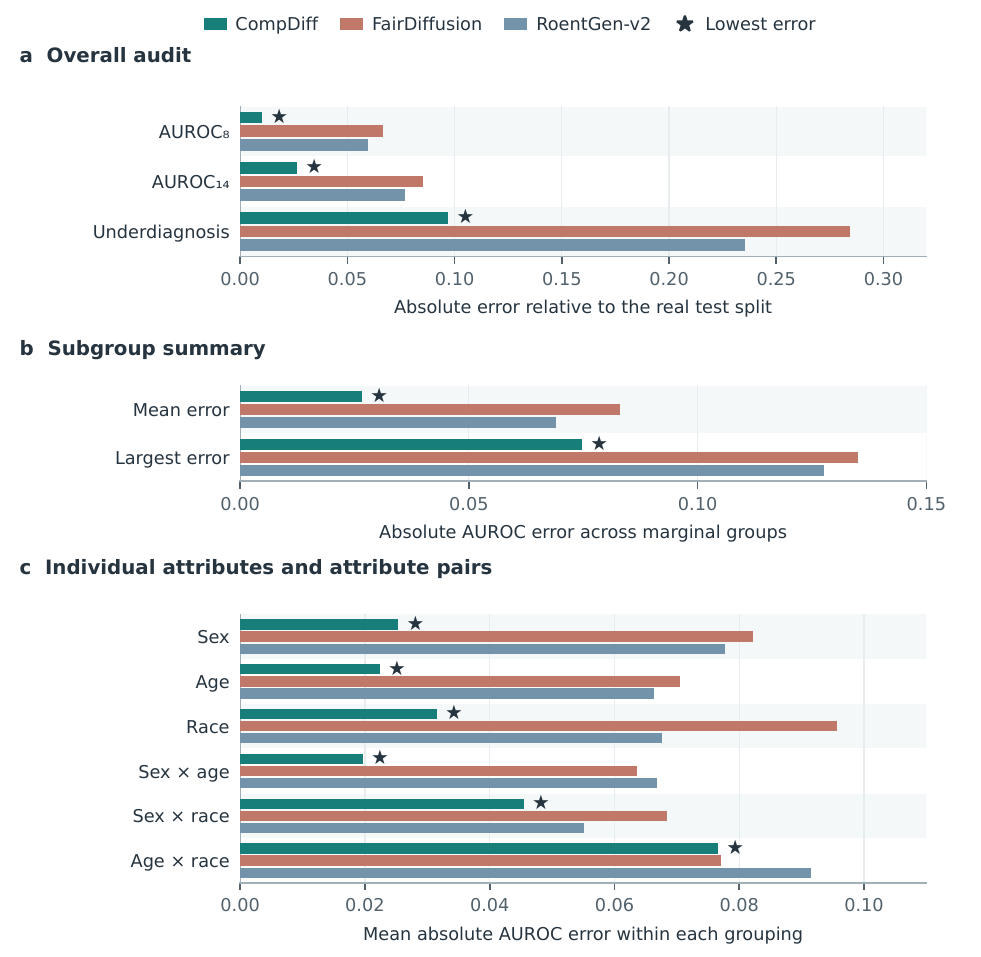}

\end{minipage}
\end{adjustbox}
\par\DisplayMark{fig:audit-instrument/caption-top}
\caption{Agreement between synthetic and real test cohorts when auditing a fixed
DenseNet-\numMainFiguresAP{} trained on real data. Each synthetic cohort matches the
real test split in size and demographic distribution. Bars show absolute differences
from the real test result; lower values indicate closer agreement.
\textbf{a,} Overall errors for AUROC$_8$, AUROC$_{\numMainFiguresAQ{}}$ and underdiagnosis rate.
\textbf{b,} Mean and largest absolute AUROC errors across marginal sex, race and age groups.
\textbf{c,} Mean absolute AUROC error within each attribute and attribute pair.
Colours identify the generator. A star marks the lowest observed error in each comparison,
selected before rounding. All comparisons use a single fixed classifier.}
\label{fig:audit-instrument}
\par\DisplayMark{fig:audit-instrument/caption-bottom}
\end{figure*}

\begin{figure*}[p]
\centering
\DisplayMark{fig:per-cell-calibration/plot-top}
\begin{adjustbox}{max totalsize={\textwidth}{0.76\textheight},center}
\begin{minipage}{\textwidth}
\centering

\includegraphics[width=\textwidth]{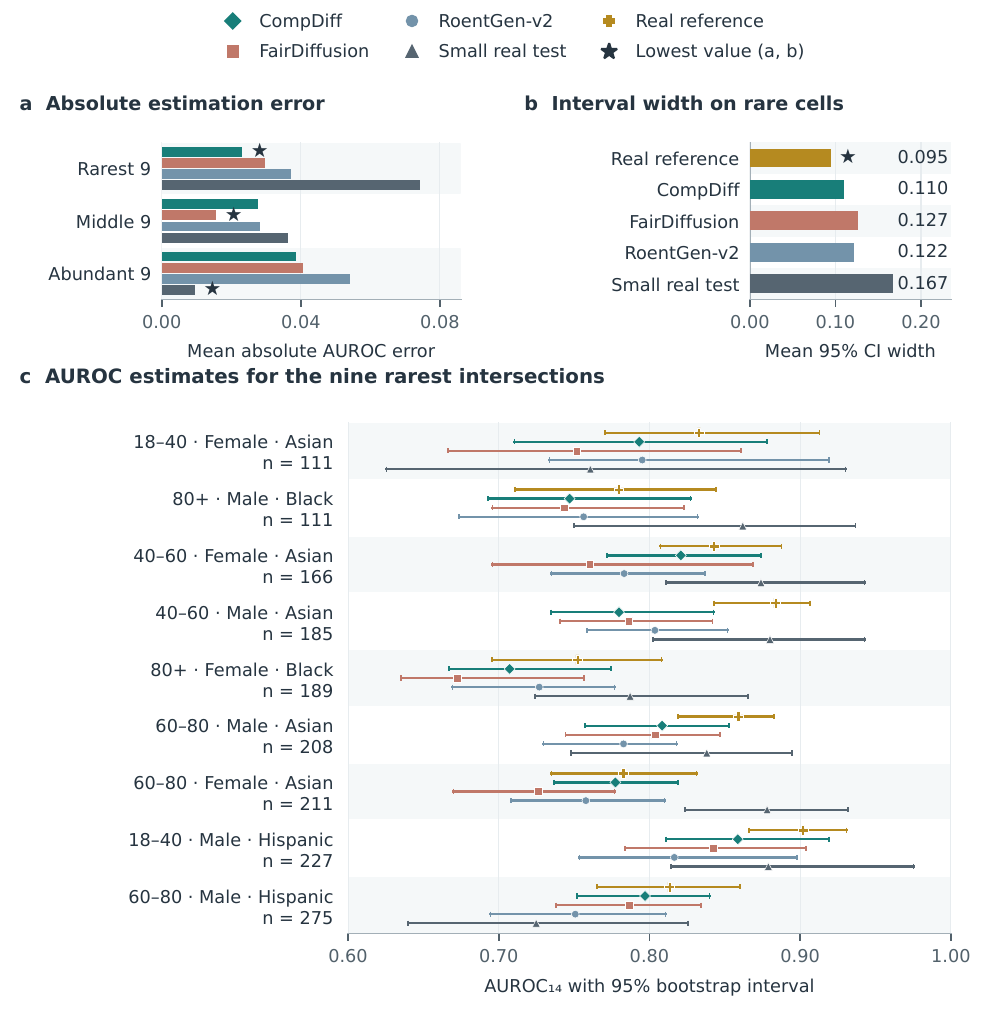}

\end{minipage}
\end{adjustbox}
\par\DisplayMark{fig:per-cell-calibration/caption-top}
\caption{Accuracy and precision of classifier evaluation on rare demographic intersections.
A single DenseNet-\numMainFiguresAR{} trained on real data is evaluated using macro AUROC
over \numMainFiguresAS{} labels. The independent real reference split is excluded from
training and model selection. Each synthetic cohort matches its demographic distribution
and cell sizes; the small real test cohort contains fewer images.
\textbf{a,} Mean absolute difference from the reference AUROC across three rarity tiers
of \numMainFiguresAV{}, \numMainFiguresAW{} and \numMainFiguresAX{} cells, ordered by reference
cohort size. Grouped bars show the four estimators, with a star marking the lowest
observed error in each tier. Lower error indicates greater accuracy relative to the reference.
\textbf{b,} Mean \numMainFiguresAU{}\% confidence-interval width on the
\numMainFiguresBA{} rarest cells. A star identifies the narrowest mean interval among
all five cohorts, including the real reference. Narrower intervals indicate greater precision.
\textbf{c,} AUROC estimates and \numMainFiguresAY{}\% bootstrap intervals for each of those
cells, with the real reference marked by a plus. The displayed $n$ is the reference cell size.
Intervals use \numMainFiguresAT{} bootstrap iterations. The same nine cells contribute to
panels b and c. These are estimates for one classifier; panels a and b do not quantify
variation across classifier training runs.}
\label{fig:per-cell-calibration}
\par\DisplayMark{fig:per-cell-calibration/caption-bottom}
\end{figure*}

We tested two uses of synthetic cohorts for classifier evaluation: reproducing audit results from real data at overall and subgroup levels, and estimating performance on rare intersections where real test data are sparse. Each audit used a single DenseNet-\numResultsCT{} trained on real MIMIC-CXR and held fixed during evaluation. Results are estimates from a single classifier, without replication across training seeds. The two designs are shown in Figures~\ref{fig:audit-instrument} and~\ref{fig:per-cell-calibration}; protocols are given in Section~\ref{subsec:eval-protocol}.

\paragraph{Agreement with audits using real data.}
Synthetic cohorts matched the real test split in size and demographic distribution. CompDiff reproduced the real underdiagnosis rate most closely, with an absolute error of $\numResultsCY{}$, compared with $\numResultsCZ{}$ for RoentGen-v2 and $\numResultsDA{}$ for FairDiffusion. Its overall AUROC$_8$ error was also lower: $\numResultsDD{}$, compared with $\numResultsDE{}$ and $\numResultsDF{}$, respectively. Averaging all $\numResultsDJ{}$ MIMIC-CXR labels, rather than the eight shared with CheXpert, gave AUROC$_{\numResultsDJ{}}$ errors of $\numResultsDG{}$, $\numResultsDH{}$ and $\numResultsDI{}$. Figure~\ref{fig:audit-instrument}a reports these overall comparisons.

Agreement extended to the subgroup audits in Figure~\ref{fig:audit-instrument}b,c. Mean absolute AUROC error across marginal sex, race and age groups was $\numResultsDK{}$ for CompDiff, compared with $\numResultsDL{}$ for RoentGen-v2 and $\numResultsDM{}$ for FairDiffusion. The largest subgroup error was also smallest for CompDiff: $\numResultsDN{}$, compared with $\numResultsDO{}$ and $\numResultsDP{}$, respectively. CompDiff had the lowest error in every comparison of individual attributes and attribute pairs, although the difference from FairDiffusion for Age$\times$Race was within rounding ($\numResultsDQ{}$ versus $\numResultsDR{}$).

\paragraph{Estimation on rare intersections.}
To estimate the same $\numResultsDJ{}$-label macro AUROC within each cell, we used a larger, independent real reference split, excluded from generator training, validation and model selection, and from classifier training. Each synthetic cohort matched this reference in demographic distribution and cell size. Comparisons included the \numResultsDS{} intersections for which all cohorts yielded an AUROC estimate, grouped by reference cohort size.

On the \numResultsDT{} rarest cells, mean absolute estimation error was $\numResultsDU{}$ for CompDiff, $\numResultsDV{}$ for FairDiffusion and $\numResultsDW{}$ for RoentGen-v2, compared with $\numResultsDX{}$ for the small real test cohort (Figure~\ref{fig:per-cell-calibration}a). CompDiff reduced error by $\numResultsDY{}\%$ relative to the real test cohort, which contained only $\numResultsDZ{}$--$\numResultsEA{}$ images per cell on this tier. The advantage depended on cell size: real test data gave the lowest error on the \numResultsEB{} most abundant cells ($\numResultsEC{}$ versus $\numResultsED{}$ for CompDiff), while FairDiffusion performed best on the middle tier ($\numResultsEE{}$ versus $\numResultsEF{}$).

Synthetic cohorts also gave more precise estimates on rare cells. With \numResultsEG{} bootstrap iterations, mean $\numResultsEH{}\%$ confidence interval width on the rarest tier was $\numResultsEJ{}$ for CompDiff, compared with $\numResultsEI{}$ for the small real test cohort, $\numResultsEK{}$ for RoentGen-v2 and $\numResultsEL{}$ for FairDiffusion (Figure~\ref{fig:per-cell-calibration}b,c). The real reference cohort had the narrowest intervals, with a mean width of $\numResultsEN{}$.

The $\numResultsEO{}$--$\numResultsEP{}$ Female Asian cell illustrates the difference. With $\numResultsER{}$ real test images, estimated AUROC was $\numResultsES{}$ with a $\numResultsEH{}\%$ confidence interval of $[\numResultsET{}, \numResultsEU{}]$, excluding the reference estimate of $\numResultsEV{}$ from $\numResultsEQ{}$ images. CompDiff gave an estimate of $\numResultsEW{}$ with a narrower interval, $[\numResultsEX{}, \numResultsEY{}]$, that included the reference estimate.

These findings support synthetic cohorts as a complement to audits using real data for sparsely represented intersections. Real cohorts remain necessary for final validation and performed best on abundant cells.

\subsection{Conditioning interface ablations}\label{subsec:ablations}

Chest X-ray ablations tested which parts of CompDiff supported image fidelity while preserving demographic control (Table~\ref{tab:ablations}). Control was assessed with pretrained demographic predictors: eligible checkpoints required sex accuracy of at least $\ckptFloorSex$, race accuracy of at least $\ckptFloorRace$ and age RMSE no greater than $\ckptFloorAge$ years. We compared fidelity at each variant's selected checkpoint and used matched composer controls to separate the contribution of the interface from its internal wiring.

\begin{table*}[!tp]
\centering
\caption{Ablations, grouped by design question and ordered by effect size. All values are validation metrics at each run's own selected checkpoint, chosen by the rule used throughout this paper (Eq.~\ref{eq:ckpt-rule}: lowest validation FID among checkpoints meeting the controllability floors). Where a run has no eligible checkpoint we report its checkpoint with the lowest age RMSE, since in those runs the age floor is unreachable over the entire trajectory and so disqualifies every checkpoint. Rows marked $\ddagger$ were measured under the configuration in which age is conditioned through the text prompt and only sex and race pass through the conditioner; they are comparable among themselves and to the $\ddagger$ reference row. \emph{Floors} states whether the reported checkpoint meets the floors, and in each ablated row the value that decides its verdict is set in bold. FID-RN abbreviates FID-RadImageNet; AUROC is the XRV disease classifier AUROC on the synthetic images (prompt alignment, Section~\ref{subsec:eval-synth}), not a downstream classifier.}
\label{tab:ablations}
\noindent{\fontsize{8}{9.6}\selectfont
\begin{adjustbox}{max width=\textwidth}\begin{tabular}{@{}lccccccc@{}}
\toprule
\textbf{Variant} & \textbf{FID}$\downarrow$ & \textbf{FID-RN}$\downarrow$ & \textbf{Sex}$\uparrow$ & \textbf{Race}$\uparrow$ & \textbf{Age}$\downarrow$ & \textbf{AUROC}$\uparrow$ & \textbf{Floors} \\
\midrule
\multicolumn{8}{@{}l}{\emph{Where demographics enter}} \\
Demographics in the prompt$^{\ddagger}$ & \textbf{94.5} & 8.90 & 0.99 & 0.99 & 5.79 & 0.75 & \checkmark \\
No demographic conditioning$^{\ddagger}$ & 70.9 & 6.30 & \textbf{0.52} & 0.68 & 17.40 & 0.78 & $\times$ \\
Second text branch$^{\ddagger}$ & \textbf{140.0} & 9.60 & 0.75 & 0.50 & 20.70 & 0.74 & $\times$ \\
Route through timestep modulation & 72.0 & 6.84 & 0.99 & 0.96 & 8.76 & 0.77 & \checkmark \\
\addlinespace
\multicolumn{8}{@{}l}{\emph{Where supervision lands}} \\
No output supervision$^{\ddagger}$ & 80.3 & 7.33 & \textbf{0.51} & 0.69 & 18.10 & 0.70 & $\times$ \\
Token dropout instead of supervision$^{\ddagger}$ & 80.1 & 7.03 & \textbf{0.51} & 0.71 & 16.60 & 0.72 & $\times$ \\
Supervision before projection$^{\ddagger}$ & 79.8 & -- & \textbf{0.54} & 0.68 & 17.90 & 0.71 & $\times$ \\
Single composed token & 123.7 & 17.23 & 0.98 & 0.90 & \textbf{12.41} & 0.63 & $\times$ \\
\addlinespace
\multicolumn{8}{@{}l}{\emph{How attributes are represented}} \\
Fused demographic embedding$^{\ddagger}$ & 80.6 & 5.70 & \textbf{0.50} & 0.70 & 17.20 & 0.70 & $\times$ \\
Binned age embedding$^{\ddagger}$ & 74.8 & 5.50 & 0.96 & 0.87 & \textbf{15.31} & 0.76 & $\times$ \\
\addlinespace
\multicolumn{8}{@{}l}{\emph{How attributes are combined}} \\
Factorised composer (matched pair) & 71.2 & 6.34 & 1.00 & 0.96 & 8.98 & 0.76 & \checkmark \\
Unstructured composer (matched pair) & 80.5 & 5.28 & 0.99 & 0.96 & 8.51 & 0.74 & \checkmark \\
Transformer composer & 113.5 & 14.54 & 0.95 & 0.94 & \textbf{11.33} & 0.67 & $\times$ \\
\addlinespace
\multicolumn{8}{@{}l}{\emph{Loss terms}} \\
No KL regularisation$^{\ddagger}$ & 77.1 & 7.54 & 0.99 & 0.95 & 8.16 & \textbf{0.73} & \checkmark \\
No variational latent$^{\ddagger}$ & 77.6 & -- & 1.00 & 0.96 & \textbf{10.10} & 0.74 & $\times$ \\
No compositional consistency$^{\ddagger}$ & \textbf{88.0} & 9.15 & 0.97 & 0.94 & 8.00 & 0.75 & $\times$ \\
Strong compositional consistency$^{\ddagger}$ & \textbf{97.1} & 11.59 & 0.99 & 0.94 & 10.80 & 0.72 & $\times$ \\
\midrule
\multicolumn{8}{@{}l}{\emph{Reference configurations}} \\
Age in prompt; sex and race in conditioner$^{\ddagger}$ & 75.5 & 7.19 & 0.99 & 0.96 & 8.75 & 0.76 & \checkmark \\
\textbf{CompDiff (final)} & 71.6 & 6.06 & 1.00 & 0.96 & 8.96 & 0.75 & \checkmark \\
\bottomrule
\end{tabular}\end{adjustbox}
}
\end{table*}

\paragraph{Separating demographics from clinical text.}
Encoding demographics in the prompt preserved sex and race accuracy ($\numResultsFA{}$ and $\numResultsFB{}$) but gave poor fidelity (FID $\numResultsFC{}$). Removing demographics improved FID to $\numResultsFD{}$ while reducing sex accuracy to chance ($\numResultsFE{}$). Fewer than $\numResultsFF{}\%$ of training prompts exceeded CLIP's $\numResultsFG{}$-token limit, suggesting that this trade-off involved competition within the text representation rather than widespread truncation (Supplementary Figure~\ref{fig:clip-sensitivity}; \AppendixName~\ref{app:clip-sensitivity}). A second text branch did not resolve it, reaching FID $\numResultsFH{}$ and sex accuracy $\numResultsFI{}$. The dedicated conditioner provided control with substantially better fidelity. Adding a route through timestep modulation alongside its cross-attention tokens also met the control thresholds, at FID $\numResultsFJ{}$ versus $\numResultsFK{}$ for the final model; cross-attention alone was sufficient in this comparison.

\paragraph{Preserving each attribute at the output.}
Removing output supervision reduced sex accuracy to $\numResultsFL{}$; token dropout did not recover it ($\numResultsFM{}$). Applying supervision before projection also failed ($\numResultsFN{}$), consistent with demographic information being lost before it reached the UNet. Separate output tokens mattered: a single composed token failed the age and race thresholds even at its closest checkpoint ($\numResultsFO{}$ years and $\numResultsFP{}$). The attribute encoders also contributed. A fused demographic embedding receiving the same supervision reached only chance sex accuracy ($\numResultsFQ{}$), while discretising age worsened age RMSE to $\numResultsFS{}$ years and race accuracy to $\numResultsFR{}$. Together, these controls support separate encoders for age, sex and race, separate output tokens and supervision after projection.

\paragraph{Composer choice and convergence.}
Replacing pairwise composition with an unstructured MLP while retaining that interface preserved both control and fidelity. The matched composers differed in parameter count by only \ablParamDeltaPct\%. At their selected checkpoints, neither was uniformly better: the unstructured composer had higher FID but lower FID-RadImageNet and age RMSE. At a common training step, differences on the test set remained within the measured variation between independent runs of the final configuration. The unstructured control also preserved fidelity within each cell on both holdout splits, with differences varying in direction (\AppendixName~\ref{app:flat-composer}). Thus, pairwise factorisation was not required for generation at unseen intersections in these experiments.

Factorisation did improve convergence: its runs first met all three control thresholds at steps \ablHierFloorLo{}--\ablHierFloorHi{}, compared with step \ablFlatFloorStep{} for the unstructured control. Composer choice still required care. The transformer variant failed the age threshold ($\numResultsFZ{}$ years); its matched control using timestep modulation met it ($\numResultsGA{}$ years). The evidence therefore supports more than one effective composer, without implying that every architecture is interchangeable.

\paragraph{Regularisation and the variational latent.}
Removing KL regularisation retained age control but reduced disease alignment, whereas removing the variational latent also worsened FID to $\numResultsGE{}$ and age RMSE to $\numResultsGF{}$ years. The stochastic latent therefore contributed beyond its KL penalty. Removing compositional consistency worsened FID to $\numResultsGH{}$, and strengthening it worsened FID further to $\numResultsGJ{}$. The default weight $\lambda_{\mathrm{comp}}=\numResultsGK{}$ gave the better balance among the tested settings.

\FloatBarrier
\section{Discussion}\label{sec:discussion}

CompDiff improved generation for demographic intersections that were poorly represented or absent in training, and made the resulting synthetic cohorts more useful for classifier training and fairness auditing. Across chest radiographs and fundus images, the gains extended beyond average fidelity to subgroup generation quality and clinical alignment. With three training runs per method and holdout, CompDiff had the lowest mean FID-RadImageNet in all $\scCells$ unseen chest intersections. The average differences favoured CompDiff over both RoentGen-v2 and FairDiffusion and remained favourable when any single trained model was omitted. A separate blinded radiologist assessment supported the realism and clinical correspondence of CompDiff images at unseen intersections. These findings address a practical constraint in medical imaging: the groups for which additional data are most needed may also be those a conventional generator represents least well.

For classifier development, CompDiff improved what classifiers could learn from synthetic chest images. Repeated training supported an advantage when synthetic images were the sole chest training source, and fine-tuning on real images extended the gains across diagnostic labels on both MIMIC-CXR and external CheXpert. Higher pneumothorax sensitivity at a common specificity helps interpret the discrimination gain, although clinical operating thresholds would require independent validation. Improvements also reached the demographic groups with the lowest baseline performance, supporting the use of synthetic pretraining to improve performance for underserved groups. Mixing real and synthetic images during training did not show a consistent advantage (Supplementary Table~\ref{tab:downstream-runs}). These findings support the matched training setting studied here and motivate testing which training choices sustain the benefit.

The auditing experiments suggest a second use: constructing evaluation cohorts for intersections too small to support reliable estimates from the available real test split. CompDiff reproduced overall and subgroup audit measurements more closely than the two baselines. On the rarest intersections, its cohorts also gave lower AUROC estimation error and narrower confidence intervals than the small real test cohort. Accuracy and precision are separate findings here: a narrow interval alone would not show that an estimate was close to the real reference. Each synthetic cohort matched a larger, independent real reference in size and demographic distribution, so the experiment evaluates the practical value of generating additional samples. The advantage was concentrated in rare cells; real test data remained the better estimator on abundant cells. Synthetic audits can therefore complement evaluation on real data, with their accuracy checked against an independent real reference where available.

The chest ablations help explain these benefits. Demographic control depended on preserving each attribute through a dedicated encoder, a separate output token and supervision at the point where the UNet received that token. A fused demographic embedding failed despite receiving the same auxiliary supervision. By contrast, an unstructured composer retained fidelity and control, including on unseen intersections, when it preserved the separate encoding and supervision of each attribute. HCN's pairwise factorisation improved convergence but was not necessary for the observed compositional capability. This separates the supported design principle from the particular composer used to implement it. HCN was trained jointly with the unfrozen UNet and text encoder; its use with other backbones or training regimes remains to be evaluated.

This contribution complements existing approaches to synthetic medical data. FairGen and FairDiffusion intervene in how training favours outputs or demographic groups~\cite{li2026fairgen,friedrich2024fairdiffusion}; RoentMod uses counterfactual pathology edits to address shortcuts~\cite{cooke2026roentmod}, and ChexGen supports broader controllable chest radiograph synthesis~\cite{ji2026chexgen}. CompDiff instead tests a demographic conditioning interface on intersections deliberately removed from training. It also differs from methods that compose conditional scores during sampling or regularise their independence~\cite{liu2022composable,gaudi2025coind}: attributes are combined inside one conditioner, which can learn their interactions without an independence assumption. The comparable unstructured composer is consistent with evidence that factorised representations can emerge during diffusion training~\cite{liang2024diffusion}. At the data level, combining examples from parent groups offers another approach to augmentation of minority intersections~\cite{maheshwari2024synthetic}. Matched comparisons with FairGen, RoentMod and ChexGen remain open; the present study does not establish superiority over those systems.

Race requires particular care in interpreting these results. Here it denotes self-reported or administratively recorded categories supplied with the datasets, treated as sociotechnical labels rather than biological ground truth. Race alignment measures the recovery of image features associated with the dataset by a pretrained classifier. Such features may reflect acquisition sites and protocols, disease prevalence, treatment patterns or other correlates that are not fully characterised~\cite{glocker2023algorithmic}. A generator can reproduce those associations along with useful image content. Consequently, demographic alignment does not establish biological fidelity or fairness, and findings from synthetic cohorts require revalidation when transferred to populations with different acquisition and prevalence patterns.

Several limitations define the next steps. Three training runs per method and holdout provide limited precision for estimating variability. Their intervals are conditional on the two selected holdout tasks and fixed real cohorts, and combine training, checkpoint selection and generation variability. Excluding runs with different training and validation schedules preserved the direction of the zero-shot comparisons (Supplementary Note~\ref{app:zeroshot-inception}). The conditioner currently handles age, sex and race; richer clinical and acquisition attributes would test how well the interface scales beyond this setting. The downstream sweep shares synthetic corpora and real test cohorts across configurations; its configuration counts describe consistency within that sweep. Some subgroup labels occur in few patients, limiting precision even when their median gains are positive. Classifier audits used a single fixed model per experiment, so their conclusions require replication across classifier training runs. CheXpert provides evidence of external transfer for downstream training, but broader evaluation across institutions, scanners and labelling conventions is needed for both generation and auditing.

The reader study involved two radiologists with little overlap between their assigned sets. Treating readers as fixed effects supports inference across cases, not across the wider radiologist population; a larger panel with substantial overlap would allow assessment of inter-rater reliability and more precise subgroup judgements. Prompt correspondence from a single image also cannot resolve every statement about interval change. Finally, small reference cohorts affect fidelity estimation as well as classifier auditing. Rare cells may have both fewer and more homogeneous images, and FID estimated from finite samples depends on cohort size. The robustness analyses supported the method ordering within each cell, but do not justify ranking the intrinsic difficulty or fidelity of one demographic intersection against another.

The central implication is that preserving and supervising demographic attributes through a dedicated conditioning interface can make synthetic medical images more useful where real data are scarce. In the settings studied, that interface supported generation at unseen intersections, improved classifier training for underserved groups and enabled more accurate evaluation of rare subgroups.

\FloatBarrier
\section{Methods}\label{sec:methods}\label{sec:experiments}

\subsection{Study design, datasets and ethics}\label{subsec:datasets}\label{subsec:ethics}
We evaluated generation quality and downstream classifier performance on chest radiographs and fundus images. Chest radiographs were also used for the reader study at withheld intersections, the sweep of classifier training configurations and synthetic classifier audits. The generator comparisons used RoentGen-v2, FairDiffusion and CompDiff. Figure~\ref{fig:problem} summarises the experiments.

For chest imaging, we used postero-anterior (PA) MIMIC-CXR views with complete age, sex and race metadata~\cite{johnson2019mimiccxr}. Following the RoentGen-v2 protocol based on patient folders~\cite{moroianu2025improvingperformancerobustnessfairness}, patient folder \texttt{p19} supplied the test split ($\numMethodsBK{}$ images), and \texttt{p10}--\texttt{p18} supplied training and validation ($\numMethodsBL{}$ and $\numMethodsBM{}$ images). Test patients were disjoint from training and validation patients. Counts differ from the published RoentGen-v2 cohort because we required PA views and complete metadata. Supplementary Figure~\ref{fig:zeroshot-design}a shows the intersectional counts. FairGenMed~\cite{friedrich2024fairdiffusion} supplied $\numMethodsBN{}$, $\numMethodsBO{}$ and $\numMethodsBP{}$ scanning laser ophthalmoscopy fundus images for training, validation and testing, respectively. CheXpert provided external evaluation for the chest sweep of classifier training configurations~\cite{irvin2019chexpert}.

The audit of rare intersections used a separate partition of MIMIC-CXR. We formed four demographically stratified splits and retrained each generator on the $\numMethodsBR{}$-image training partition. A larger real reference split of $\numMethodsBQ{}$ images was excluded from generator and classifier training, validation, checkpoint selection, and hyperparameter and model selection. It supplied the reference AUROC estimates for this experiment (Section~\ref{subsec:eval-protocol}).

The study used de-identified datasets under their access conditions: MIMIC-CXR through the PhysioNet data use agreement for credentialed access and FairGenMed under its licence for research use. No new patient data were collected and no patients were contacted. The radiologist readers participated as co-authors, assessed de-identified real and synthetic images, and provided no personal data about themselves.

\subsection{CompDiff architecture and objective}\label{subsec:objective}\label{app:implementation}
CompDiff uses a Hierarchical Conditioner Network (HCN) to encode age, sex and race separately from clinical text (Figure~\ref{fig:architecture}). CLIP encodes the clinical prompt as $E_{\text{text}}\in\mathbb{R}^{B\times77\times d_{\text{ctx}}}$, where $B$ is the batch size and $d_{\text{ctx}}$ is the cross-attention dimension. HCN supplies $T=\numMethodsC{}$ tokens, one per attribute plus a composed token. Concatenating them gives the UNet context $E_{\text{combined}}=[E_{\text{text}},c]\in\mathbb{R}^{B\times81\times d_{\text{ctx}}}$. Demographic attributes enter through HCN and are removed from the text prompt.

\paragraph{Attribute encoding and composition.}
For $V=\{\text{age},\text{sex},\text{race}\}$ and $K=|V|=\numMethodsD{}$, each attribute has an embedding $e_v\in\mathbb{R}^{d_{\text{node}}}$. Sex and race use lookup tables with $\numMethodsE{}$ categories (male, female) and $\numMethodsF{}$ categories (White, Black, Asian, Hispanic/Other), respectively. Missing demographic attributes were excluded from the cohorts. Each encoder also has a learned null embedding for partial conditioning at inference. Age in years, $a$, is encoded continuously:
\begin{equation}
\begin{aligned}
e_{\text{age}}&=\mathrm{MLP}_{\text{age}}\big(\gamma(a)\big),\\
\gamma(a)_{k}&=\Big[\cos\!\big(a\omega_k\big),\ \sin\!\big(a\omega_k\big)\Big],
\end{aligned}
\end{equation}
Here $\omega_k=\eta^{-k/(d_\gamma/\numMethodsI{})}$ for $k=\numMethodsJ{},\dots,d_\gamma/\numMethodsK{}-\numMethodsL{}$, with $\eta=\numMethodsM{}^4$. A dedicated MLP combines each unordered attribute pair,
\begin{equation}
h_{v,w}=f_{v,w}([e_v,e_w])\in\mathbb{R}^{d_{\text{node}}},
\end{equation}
where brackets denote concatenation. For three attributes, the pairs are age--sex, age--race and sex--race. Their representations are fused by a further MLP:
\begin{equation}
\begin{aligned}
h_{\text{demographic}}&=g\big([\,h_{v,w}:\{v,w\}\subseteq V,\; v<w\,]\big)\\
&\in\mathbb{R}^{d_{\text{node}}},
\end{aligned}
\end{equation}
The input dimension of $g$ is $\binom{K}{\numMethodsO{}}d_{\text{node}}$ and its output dimension is $d_{\text{node}}$. The composed state parameterises a diagonal Gaussian through $(\mu,\log\sigma)$; we sample $z$ by reparameterisation during training and use $\mu$ at inference. For each attribute, a dedicated MLP combines its embedding with the composed state,
\begin{equation}
u_v = q_v\big([\,e_v,\;h_{\text{demographic}}\,]\big)\in\mathbb{R}^{d_{\text{node}}},\qquad v\in V,
\end{equation}
and separate projections produce the context tokens:
\begin{equation}
\begin{split}
c=\big[\,&\mathrm{proj}_{\text{age}}(u_{\text{age}}),\;
\mathrm{proj}_{\text{sex}}(u_{\text{sex}}),\\
&\mathrm{proj}_{\text{race}}(u_{\text{race}}),\;
\mathrm{proj}_{\text{cls}}(z)\,\big],
\end{split}
\end{equation}
Each projection is LayerNorm followed by a linear map. The variational latent appears only in the composed token; the attribute tokens are deterministic functions of $u_v$.

We used $d_{\text{node}}=\numMethodsAK{}$ and $d_{\text{ctx}}=\numMethodsAL{}$. The age encoder has $d_\gamma=\numMethodsAM{}$ sinusoidal features over $[\numMethodsAN{},a_{\max}]$, where $a_{\max}=\numMethodsAO{}$ years, followed by $\mathrm{Linear}(d_{\text{node}})\rightarrow\mathrm{SiLU}\rightarrow\mathrm{Linear}(d_{\text{node}})$. The pairwise, composition and contextualisation MLPs use $\mathrm{LayerNorm}\rightarrow\mathrm{Linear}(2d_{\text{node}})\rightarrow\mathrm{SiLU}\rightarrow\mathrm{Dropout}(\numMethodsAP{})\rightarrow\mathrm{Linear}(d_{\text{node}})$. The variational heads use LayerNorm and a linear map, with $\log\sigma$ clamped to $[-\numMethodsAQ{},\numMethodsAR{}]$.

\paragraph{Training objective.}
The joint training objective is
\begin{equation}
\label{eq:loss}
\mathcal{L}=\mathcal{L}_{\text{diff}}+\lambda_{\text{comp}}\mathcal{L}_{\text{comp}}+\lambda_{\text{aux}}\mathcal{L}_{\text{aux}}+ \lambda_{\text{KL}} \mathcal{L}_{\text{KL}}.
\end{equation}
The diffusion term is the standard loss for predicting noise~\cite{ho2020denoising}, $\mathcal{L}_{\text{diff}}=\mathbb{E}_{x_0,\epsilon,t}\|\epsilon-\epsilon_\theta(x_t,t,E_{\text{combined}})\|_2^2$, over clean latent $x_0$, Gaussian noise $\epsilon$ and diffusion timestep $t$. The variational latent is regularised towards a standard normal:
\begin{equation}
\mathcal{L}_{\text{KL}}=\mathbb{E}\big[\,\mathrm{KL}\big(\mathcal{N}(\mu,\sigma^2 I)\,\big\|\,\mathcal{N}(0,I)\big)\,\big],
\end{equation}
Compositional consistency provides an additive reference through $\mathcal{L}_{\text{comp}}=\numMethodsR{}-\cos(h_{\text{demographic}},e_{\text{age}}+e_{\text{sex}}+e_{\text{race}})$. Auxiliary heads supervise the projected tokens consumed by the UNet:
\begin{equation}
\label{eq:aux}
\begin{aligned}
\mathcal{L}_{\text{aux}}=\frac{1}{3+\lambda_{\text{joint}}}\Big[\;
&\mathrm{CE}(\hat y_{\text{sex}},y_{\text{sex}})\\
&+\mathrm{CE}(\hat y_{\text{race}},y_{\text{race}})\\
&+\kappa\,\mathrm{SL}_1(\hat a,\,a/a_{\max})\\
&+\lambda_{\text{joint}}\,\mathrm{CE}(\hat y_{\text{joint}},y_{\text{joint}})\;\Big],
\end{aligned}
\end{equation}
Sex and race use cross-entropy (CE) on their respective tokens. Age uses smooth-$L_1$ regression on $a/a_{\max}\in[\numMethodsS{},\numMethodsT{}]$, rescaled by $\kappa$. The joint head supervises the composed token on the age $\times$ sex $\times$ race label. Adult age bins are $\numMethodsU{}$--$\numMethodsV{}$, $\numMethodsW{}$--$\numMethodsX{}$, $\numMethodsY{}$--$\numMethodsZ{}$ and $\numMethodsAA{}+$, giving $\numMethodsAE{}$ adult intersections. The implementation includes an additional $\numMethodsBA{}$--$\numMethodsBB{}$ bin and $\numMethodsBF{}$ joint outputs; the eight outputs for this extra bin receive no training signal because all included patients are adults.

Auxiliary heads use $\mathrm{LayerNorm}\rightarrow\mathrm{Linear}(\numMethodsAS{})\rightarrow\mathrm{SiLU}\rightarrow\mathrm{Dropout}\rightarrow\mathrm{Linear}(C)$, with $C$ the number of classes or $\numMethodsAT{}$ for age regression. We set $\lambda_{\text{KL}}=\numMethodsAU{}$, annealed linearly over $\numMethodsAV{}$ steps, $\lambda_{\text{comp}}=\numMethodsAW{}$, $\lambda_{\text{aux}}=\numMethodsAX{}$, $\lambda_{\text{joint}}=\numMethodsAY{}$ and $\kappa=\numMethodsAZ{}$. HCN adds $\numMethodsAF{}$M parameters to the $\numMethodsAG{}$B trainable backbone parameters ($\numMethodsAH{}\%$). The auxiliary heads account for $\numMethodsAI{}$M and are discarded at inference, leaving a $\numMethodsAJ{}\%$ parameter increase.

\subsection{Training, sampling and checkpoint selection}\label{subsec:gen-training}\label{subsec:gen-sampling}\label{subsec:checkpoint-schedule}\label{app:checkpoint-selection}
All three generators used Stable Diffusion 2.1~\cite{rombach2022stablediffusion}. RoentGen-v2 encoded demographics in the clinical text prompt~\cite{moroianu2025improvingperformancerobustnessfairness}; FairDiffusion added adaptive subgroup loss weighting through Fair Bayesian Perturbation~\cite{friedrich2024fairdiffusion}; CompDiff used HCN. For the standard chest X-ray and fundus comparisons, each generator was trained independently with three random seeds per modality. The figure label \emph{Baseline SD} denotes RoentGen-v2.

Chest prompts followed \texttt{<AGE> year old <RACE> <SEX>. <IMPRESSION>}. Fundus prompts combined demographics with glaucoma status, vision (normal or mild, moderate or severe loss), cup-disc ratio (normal, borderline or abnormal), and refraction (hyperopia, emmetropia or myopia). CompDiff removed demographics from the text and passed them to HCN; both baselines retained them in text.

The UNet and CLIP text encoder were unfrozen. Training used $\numMethodsBU{}\times512$ images, bf16 mixed precision, gradient checkpointing and an exponential moving average of the weights. AdamW used a learning rate of $\numMethodsBV{}\times\numMethodsBW{}^{-\numMethodsBX{}}$ and a cosine schedule with $\numMethodsBY{}$ warm-up steps. Standard chest X-ray experiments used a batch size per GPU of $\numMethodsBZ{}$ across six GPUs, giving an effective batch of $\numMethodsCA{}$ without gradient accumulation. Their $\numMethodsCB{}$-step budget corresponds to $\numMethodsCC{}$M sample presentations or approximately $\numMethodsCE{}$ epochs on the $\numMethodsCD{}$-image MIMIC-CXR training split.

For these chest experiments and the conditioning ablations, we evaluated $\ckptN$ validation checkpoints from steps $\numMethodsCK{}$ to $\numMethodsCL{}$. Selection minimised validation FID subject to thresholds for demographic control:
\begin{equation}
\begin{aligned}
t^\star &= \operatorname*{arg\,min}_{t \in \mathcal{A}} \ \mathrm{FID}_{\mathrm{val}}(t), \\[2pt]
\mathcal{A} &= \bigl\{\, t \;:\; \mathrm{Acc}_{\mathrm{sex}}(t) \ge \ckptFloorSex, \;\; \mathrm{Acc}_{\mathrm{race}}(t) \ge \ckptFloorRace, \\
&\hphantom{{}={}\bigl\{\,} \mathrm{RMSE}_{\mathrm{age}}(t) \le \ckptFloorAge\ \text{years} \,\bigr\},
\end{aligned}
\label{eq:ckpt-rule}
\end{equation}
The eligibility thresholds require sex and race classification accuracy of at least $\ckptFloorSex$ and $\ckptFloorRace$, respectively, and age RMSE no greater than $\ckptFloorAge$ years. They prevent selecting a checkpoint with low FID that does not preserve the requested demographics. Full trajectories are in Supplementary Tables~\ref{tab:ckpt-baseline}--\ref{tab:ckpt-compdiff}.

The zero-shot comparison across repeated training runs used three runs per method and holdout. All evaluated checkpoints were selected within the first $\scStop$ training steps. Checkpoint selection used mean validation FID-RadImageNet across the withheld cells at scheduled checkpoints of at least $\scSelectFloor$ steps. Supplementary Table~\ref{tab:seed-runs} records the selected checkpoints and selection details.

Sampling used $\numMethodsCM{}$ denoising steps and classifier-free guidance scale $\numMethodsCN{}$ at $\numMethodsCO{}\times512$ resolution. Corpora for classifier training matched the natural demographic distribution at $\numMethodsCP{}\times$, $\numMethodsCQ{}\times$ or $\numMethodsCR{}\times$ the MIMIC-CXR training size ($\numMethodsCS{}$, $\numMethodsCT{}$ and $\numMethodsCU{}$ images), or used a separate corpus designed to balance demographic intersections. The latter expanded clinical impressions across all sex and race combinations, with age reweighting to approximately equalise the four age bins. The generated corpora contained $\dsBalancedRG$ RoentGen-v2 images and $\dsBalancedCD$ CompDiff images. Audit cohorts matched either the standard test split ($\numMethodsCW{}$ images) or the separate real reference split ($\numMethodsCX{}$ images), as described in Section~\ref{subsec:eval-protocol}.

\subsection{Generation quality and unseen-intersection evaluation}\label{subsec:eval-synth}\label{app:manifold-metrics}\label{subsec:zeroshot-protocol}
Table~\ref{tab:metrics-summary} summarises the generation metrics and their directions. Distributional fidelity was assessed with Fr\'echet Inception Distance (FID)~\cite{heusel2017ttur} and its RadImageNet counterpart~\cite{mei2022radimagenet}, using Inception-v3 and RadImageNet ResNet-\numMethodsDE{} features, respectively. Lower distances indicate closer feature distributions. Inception FID was the primary fundus measure because RadImageNet was pretrained on radiological images. We also measured paired real--synthetic structural similarity with MS-SSIM~\cite{wang2003ms-ssim}, interpreted against the $\numMethodsDB{}$--$\numMethodsDC{}$ range used by RoentGen-v2~\cite{moroianu2025improvingperformancerobustnessfairness}, and chest image semantic similarity with BioViL cosine similarity~\cite{boecking2022biovil}.

\begin{table*}[!tp]
\centering
\caption{Summary of all metrics of generation quality used in this paper, grouped by measurement role; procedures are given in Section~\ref{subsec:eval-synth}. An upward arrow in the direction column indicates that higher values are better and a downward arrow that lower values are better. For MS-SSIM neither tail is desirable and we report whether the value falls inside the reference range $0.25$--$0.75$~\cite{moroianu2025improvingperformancerobustnessfairness}. Family (E) (ES-FID) is the extension for subgroup fairness used in Section~\ref{subsec:fairness}.}
\label{tab:metrics-summary}
\noindent{\fontsize{8}{9.6}\selectfont
\begin{tabularx}{\textwidth}{@{}p{2.0cm} l c p{2.45cm} Y@{}}
\toprule
\textbf{Family} & \textbf{Metric} & \textbf{Direction} & \textbf{Property / axis} & \textbf{What it measures} \\
\midrule
\multirow{4}{=}{(A) Similarity between real and synthetic images}
& FID~\cite{heusel2017ttur}                       & $\downarrow$       & Fidelity (population) & Distributional distance, Inception v3 embeddings (backbone trained on natural images) \\
& FID-RadImageNet~\cite{mei2022radimagenet}       & $\downarrow$       & Fidelity (population) & Distributional distance, RadImageNet ResNet-50 embeddings (radiology backbone) \\
& MS-SSIM~\cite{wang2003ms-ssim,moroianu2025improvingperformancerobustnessfairness}                  & in $[0.25, 0.75]$  & Fidelity (per image)  & Multiscale structural similarity between paired real and synthetic images \\
& BioViL cosine sim.~\cite{boecking2022biovil}    & $\uparrow$         & Fidelity (per image)  & Semantic alignment of individual images in a chest X-ray image--language space \\
\midrule
\multirow{5}{=}{(B) Manifold / precision-recall coverage}
& UMAP visualisation                              & qualitative        & Fidelity + Diversity  & 2-D projection of joint real/synth RadImageNet features for visual coverage \\
& Precision~\cite{kynkaanniemi2019precision}      & $\uparrow$         & Fidelity              & Fraction of synth samples inside the real $k$-NN manifold (quality of individual images) \\
& Density~\cite{naeem2020reliable}                & $\uparrow$         & Fidelity              & Refinement less sensitive to outliers of Precision \\
& Recall~\cite{kynkaanniemi2019precision}         & $\uparrow$         & Diversity             & Fraction of real samples inside the synth $k$-NN manifold (unstable; reported only for completeness) \\
& Coverage~\cite{naeem2020reliable}               & $\uparrow$         & Diversity             & Replacement less sensitive to outliers for Recall (uses manifold radius measured on real images) \\
\midrule
(C) Generalisation
& Authenticity~\cite{alaa2022faithful}            & $\uparrow$         & Neighbour similarity      & Fraction of synthetic samples passing the distance check against real neighbours; not proof of non-memorisation \\
\midrule
\multirow{4}{=}{(D) Conditioning alignment}
& Disease AUROC~\cite{cohen2021torchxrayvision}   & $\uparrow$         & Cond.\ fidelity (clinical)   & Agreement with prompted findings: XRV disease AUROC for chest; glaucoma and cup-disc classifiers for fundus \\
& Sex accuracy~\cite{glocker2023algorithmic}      & $\uparrow$         & Cond.\ fidelity (demographic)& Chest X-ray sex classifier accuracy on synth \\
& Race accuracy~\cite{cohen2021torchxrayvision}   & $\uparrow$         & Cond.\ fidelity (demographic)& XRV race classifier accuracy on synth \\
& Age RMSE~\cite{cohen2021torchxrayvision}        & $\downarrow$       & Cond.\ fidelity (demographic)& XRV age regressor RMSE (years) on synth \\
\midrule
(E) Fairness of generation quality
& ES-FID~\cite{luo2024fairvision,tian2024fairseg} & $\downarrow$       & Subgroup equity       & Overall FID plus the mean absolute deviation of subgroup FIDs from that overall value \\
\bottomrule
\end{tabularx}
}
\end{table*}

For chest distributional coverage, we extracted RadImageNet features from every available image in each cohort and fitted one joint UMAP to the pooled features. Real images form the common background in Figure~\ref{fig:umap}. Quantitative Precision and Recall~\cite{kynkaanniemi2019precision} and Density and Coverage~\cite{naeem2020reliable} were computed in the original feature space. Precision measures how many synthetic images lie within the real nearest-neighbour manifold; Recall reverses those roles. Density and Coverage provide alternatives less sensitive to synthetic outliers. Coverage uses neighbourhood radii measured on real images. We used $k=\numMethodsFW{}$ neighbours for cohorts with the natural demographic distribution ($n\approx\numMethodsFX{}$) and $k=\numMethodsFY{}$ for withheld cells ($n=\numMethodsFZ{}$--$\numMethodsGA{}$). Authenticity~\cite{alaa2022faithful} was the fraction of synthetic images whose distance to their nearest real reference image was at least that real image's nearest-neighbour distance within the real reference cohort. We treated this as a nearest-neighbour diagnostic, rather than proof of non-memorisation.

\paragraph{Conditioning alignment and subgroup fidelity.}
Pretrained classifiers assessed whether generated images expressed the conditioning labels. Chest disease alignment used TorchXRayVision's XRV DenseNet-\numMethodsDH{}~\cite{cohen2021torchxrayvision}, with prompted pathology labels as targets for AUROC over $\numMethodsDI{}$ diseases: atelectasis, cardiomegaly, oedema, pneumothorax and pleural effusion. Fundus alignment used pretrained EfficientNet classifiers for glaucoma status and cup-disc ratio categories. Chest sex accuracy used the dedicated classifier of Glocker et al.~\cite{glocker2023algorithmic}; race accuracy and age RMSE used the XRV race classifier and age regressor. Generated $\numMethodsDK{}\times512$ images were downsampled by area interpolation to the $\numMethodsDJ{}\times224$ inputs expected by XRV. Demographic targets were the attributes supplied to each generator, through text for the baselines and HCN for CompDiff. Race accuracy measures agreement with labels associated with the dataset as recovered by the selected classifier; it is not a measure of biological identity.

We computed fidelity separately by sex, race, age and their joint intersection. Equity-scaled FID~\cite{friedrich2024fairdiffusion,luo2024fairvision,tian2024fairseg} combines overall FID with the mean absolute deviation of subgroup distances from that overall value:
\begin{equation}
\begin{aligned}
\text{ES-FID}_{\mathcal{A}^i}
&= \text{FID} \cdot \Bigg(1 + \frac{1}{|\mathcal{A}^i| \cdot \text{FID}}\\
&\qquad\times\sum_{j=1}^{|\mathcal{A}^i|} |\text{FID} - \text{FID}_{\mathcal{A}_j^i}| \Bigg)
\end{aligned}
\end{equation}
Here $\mathcal{A}^i$ is the set of subgroups for attribute $i$, or the full intersectional subgroup list. The measure equals overall FID when every subgroup distance equals that overall value. Chest subgroup comparisons used RadImageNet features and fundus comparisons used Inception features.

\paragraph{Unseen intersections.}
We removed eight demographic intersections from each of two MIMIC-CXR training variants, reallocating the affected patients $\numMethodsDL{}\%$ to test and $\numMethodsDM{}\%$ to validation (Supplementary Figure~\ref{fig:zeroshot-design}). Holdout~A removed the eight rarest intersections, with natural training counts below $\numMethodsDN{}$: $\numMethodsDO{}$--$\numMethodsDP{}$ Asian women and men, and $\numMethodsDQ{}+$ Asian, Black and Hispanic women and men. Holdout~B removed the eight next rarest intersections, with counts $\numMethodsDR{}$--$\numMethodsDS{}$: $\numMethodsDT{}$--$\numMethodsDU{}$ and $\numMethodsDV{}$--$\numMethodsDW{}$ Asian women and men, and $\numMethodsDX{}$--$\numMethodsDY{}$ and $\numMethodsDZ{}$--$\numMethodsEA{}$ Hispanic women and men.

These disjoint sets of target intersections had no training patients after reallocation; other intersections retained their natural counts. Training/validation/test sizes changed from $\numMethodsEB{}/\numMethodsEC{}/\numMethodsED{}$ to $\numMethodsEE{}/\numMethodsEF{}/\numMethodsEG{}$ for Holdout~A and $\numMethodsEH{}/\numMethodsEI{}/\numMethodsEJ{}$ for Holdout~B. Each method was trained independently three times per holdout and used to generate the withheld intersections. FID-RadImageNet within each cell compared each synthetic cohort with real test images from the same intersection. The average across these cell distances gives equal weight to each intersection and differs from FID computed on pooled images.

\FloatBarrier
\subsection{Radiologist reader study}\label{subsec:reader-protocol}
Two board-certified radiologists assessed images from the $\numMethodsEK{}$ withheld chest intersections. Each set contained three candidate radiographs for the same clinical impression and intersection, produced by one selected model per method on the relevant holdout split. The readers completed $\rsNInstances$ set assessments over $\rsNSets$ distinct sets, with $\rsNOverlap$ sets assessed by both readers, giving $\rsNRatings$ image ratings.

Images appeared in randomised order. Readers were blinded to source and to the presence or proportion of real images. A hidden real image from the same intersection replaced one generator in $\rsNReal$ assessments: RoentGen-v2 in $\rsReplacedBase$, CompDiff in $\rsReplacedCompDiff$ and FairDiffusion in $\rsReplacedFairD$. Consequently, CompDiff, FairDiffusion and RoentGen-v2 appeared in $\rsNCompDiff$, $\rsNFairD$ and $\rsNBase$ assessments, respectively.

Readers scored anatomical realism from $\numMethodsEL{}$ (clearly artefactual) to $\numMethodsEM{}$ (indistinguishable from a real scan), and correspondence with the clinical impression from $-\numMethodsEN{}$ (complete mismatch) to $+\numMethodsEO{}$ (perfect correspondence), following Chambon et al.~\cite{chambon2022roentgen}. They also selected the most realistic image in each set. Readers could flag an image as unreadable or mark correspondence as not assessable, including when impressions described interval changes that could not be judged from a single image. One unreadable rating was excluded from both outcomes and $\rsNNotAssessable$ further ratings from correspondence, leaving $\rsRealismN$ realism and $\rsCorrN$ correspondence ratings. Forced-choice rates were calculated among sets containing the source. Supplementary Figure~\ref{fig:reader-interface} shows the viewer.

Statistical models, source comparisons and sensitivity analyses are specified in Section~\ref{subsec:statistics}.

\subsection{Downstream training and classifier auditing}\label{subsec:downstream-recipes}\label{subsec:eval-protocol}
\paragraph{Classifier training on synthetic data.}
The chest training sweep compared CompDiff with RoentGen-v2 under matched classifier recipes (Table~\ref{tab:recipes}). Training on synthetic data alone used no real chest images. In the second family, classifiers were pretrained on synthetic images and fine-tuned on real MIMIC-CXR data. The four synthetic corpora described in Section~\ref{subsec:gen-sampling} were evaluated with random or ImageNet initialisation and fractions of real data of $\numMethodsES{}\%$, $\numMethodsET{}\%$, $\numMethodsEU{}\%$, $\numMethodsEV{}\%$, $\numMethodsEW{}\%$ and $\numMethodsEX{}\%$, giving $\numMethodsEY{}$ pretraining/fine-tuning configurations per evaluation dataset. Recipes using synthetic data alone used ImageNet initialisation.

For each of the four fixed corpora, we also evaluated four independent classifier runs per generator, giving 32 models evaluated on both test sets. Additional experiments repeated pretraining followed by fine-tuning on all real training images and training on mixed real and synthetic images. The repeated-training experiments reported here used $256\times256$ training inputs, batch size 128, an initial learning rate of 0.001 and a 50-epoch limit, with AdamW (Supplementary Table~\ref{tab:downstream-runs}). The four runs shared these settings; their saved test evaluations used the common $224\times224$ preprocessing.

Classifiers were DenseNet-\numMethodsFE{} models~\cite{huang2017densenet}. The original fine-tuning sweep used AdamW with a cosine schedule, $\dsSweepResolution{}\times\dsSweepResolution{}$ classifier inputs, batch size 32 and a maximum of $\numMethodsFH{}$ epochs. The generated images had $512\times512$ resolution and were resized for classifier input. Batch size, centre cropping, random horizontal flips, small rotations, weight decay and early stopping were held constant between generators within each recipe. Evaluation used the real MIMIC-CXR test split and the external CheXpert test set.

AUROC$_8$ is the macro-average over a fixed set of eight diagnostic labels: atelectasis, cardiomegaly, consolidation, oedema, pleural effusion, pneumonia, pneumothorax and no finding. We recomputed all repeated-training and subgroup scores using this same label set and the saved prediction column order. For the exploratory sensitivity analysis, each test ROC supplied the maximum true positive rate at a false positive rate no greater than 0.10 or 0.05. Thresholds were read separately from each classifier's test curve, rather than validated as operating thresholds. AUROC$_8$ was used for the downstream sweep; AUROC$_{\numMethodsFC{}}$ in the MIMIC audits averages all $\numMethodsFD{}$ available pathology labels.

We reported subgroup AUROCs, equity-scaled AUROC (ES-AUC), equalised-odds difference and underdiagnosis. ES-AUC discounts overall AUROC by variation between groups: the implemented score divides overall AUROC by one plus the mean absolute deviation, weighted by group size, of subgroup AUROCs from the overall value. Equalised-odds difference is the larger of the gaps in true positive and false positive rates between groups, computed with \texttt{fairlearn}~\cite{hardt2016equalityopportunity}. Underdiagnosis is the false positive rate for \emph{No Finding}: a diseased image classified as having no finding~\cite{seyyed2021underdiagnosis}. Subgroup AUROC$_8$ required both positive and negative images for every label; all ten sex, race and age groups met this rule on each test set. We retained the same eight labels in every group rather than changing the average when a label was sparse. Supplementary Table~\ref{tab:downstream-groups} reports the smallest positive count and the sensitivity check requiring at least ten positive images for every label. For the analysis of the subgroup with the lowest baseline performance, each matched configuration identifies the subgroup with the lowest RoentGen-v2 AUROC within sex, race or age; its identity is held fixed when computing the CompDiff gain.

\paragraph{Agreement with audits using real data.}
Each audit held a single DenseNet-\numMethodsFJ{} trained on real MIMIC-CXR fixed during evaluation. Substitution of entire cohorts compared the real test split with a synthetic cohort from each generator, matched in size and demographic distribution. For each overall or subgroup metric, the audit error was its absolute difference from the corresponding value on the real test set. Outcomes included AUROC, equity-scaled AUROC, equalised-odds difference and underdiagnosis.

\paragraph{Estimation on rare intersections.}
For rare cells, the larger independent real reference split supplied AUROC estimates for each cell. The small real test split and a synthetic cohort from each generator were evaluated as estimators of these values, with each synthetic cohort matched to the reference split's size and demographic distribution. The small test split contained $\numMethodsFO{}$--$\numMethodsFP{}$ images in its most sparsely populated intersections. Absolute cell estimation errors were averaged within three tiers defined by reference cohort size: the rarest $\numMethodsFQ{}$, middle $\numMethodsFR{}$ and most abundant $\numMethodsFS{}$ cells. The reference split was excluded from training and model selection as described in Section~\ref{subsec:datasets}.

\subsection{Conditioning and composer ablations}\label{subsec:composer-control}
Conditioning and loss controls are specified in Table~\ref{tab:ablations}, including the comparable reference configuration for each group of variants. Validation metrics were reported at the checkpoint selected by the shared rule based on thresholds for demographic control. If no checkpoint met the thresholds, the table reports the checkpoint with the lowest age RMSE.

The unstructured composer retained separate attribute encoders, separate output tokens, supervision of the output tokens and the same losses, replacing pairwise composition with MLPs matched in depth and parameter count that receive the concatenated attribute embeddings. Composer sizes were $\ablComposerParamsHier$ and $\ablComposerParamsFlat$, differing by $\ablParamDeltaPct\%$. The unstructured output is split into attribute tokens whose association with each attribute is established by supervision. Both composers were also retrained on each holdout split. The factorised checkpoints were the CompDiff seed-$\scSeedFirst$ models in Supplementary Table~\ref{tab:seed-runs}; the unstructured control used the checkpoint with the lowest validation FID on Holdout~A, also matching the factorised arm in training step, and the checkpoint with the lowest validation FID-RadImageNet on Holdout~B. Further comparisons used a common step with three generation seeds or each arm's best validation checkpoint under either criterion. Steps to the thresholds for demographic control were recorded separately. Supporting composer comparisons are reported in Supplementary Note~\ref{app:flat-composer}.

\paragraph{Supplementary analysis of demographic text.}\label{subsec:clip-protocol}
For the supporting CLIP analysis, we tokenised all $\numMethodsGE{}$ training prompts ($\numMethodsGF{}$ unique clinical impressions), with and without demographics, to measure truncation at CLIP's $\numMethodsGG{}$-token limit. A separate probe of the frozen encoder used $\numMethodsGH{}$ clinical impressions and $\numMethodsGI{}$ demographic prefixes: young, middle-aged or elderly, combined with the two sex and four race labels in the form \texttt{<AGE> <RACE> <SEX> patient.}. For each impression, cosine perturbation relative to the embedding without demographic text was averaged over prefixes. These combinations using descriptive age terms differ from the $\numMethodsGK{}$ age-bin intersections. Linear probes measured recoverability of demographic labels from embeddings without demographic text. Clinical density was the fraction of words recognised as clinical entities. The visualisation used LOWESS with span $\numMethodsGL{}$, a power-law fit in log--log space, and means within quartiles of word count with $\numMethodsGM{}\%$ bootstrap intervals from $\numMethodsGN{}$ resamples. Results are reported in Supplementary Note~\ref{app:clip-sensitivity}.

\subsection{Fidelity robustness and sampling controls}\label{subsec:robustness-protocol}
Controls for the fidelity estimator and sampling budget used one selected model per configuration and holdout, including the factorised and unstructured composers. These controls were separate from the comparison across repeated training runs. Fidelity robustness analyses rescored their images using KID-RadImageNet~\cite{binkowski2018mmdgans} and corrections for sample size. Holding generator, intersection and image pool fixed, we subsampled equal numbers of synthetic and real images from large cells and measured change relative to $s=\robSizeAnchor$. Following Chong and Forsyth~\cite{chong2020unbiasedfid}, FID was fitted against inverse candidate size $1/k$ and jointly against candidate and reference sizes $(1/k,1/m)$. We compared local slopes with those fitted over a wide range of sample sizes, validated extrapolation from $\robValN$ images against withheld targets from large cells, and normalised cell scores to a common sample size using the fitted curves describing the dependence on sample size. Ranking and agreement on the winning configuration were assessed relative to the uncorrected FID-RadImageNet ordering of those configurations, with Spearman correlations within each cell. Corrected values were sensitivity analyses; primary values remained uncorrected. Withheld and well represented cells were compared within split and generator.

For post hoc selection, a verifier combined XRV demographic agreement with a nearest-neighbour realism term. Its real reference half was disjoint from the half used to evaluate selected images. Selecting the best quarter gave a $\robSelBudget$ sampling budget. We compared selected and random samples using bootstrap intervals, with Mann--Whitney tests between withheld and well represented groups. Comparisons included $\numMethodsGR{}\times$ and $\robSelBudget$ budgets and an oracle selector with access to evaluation references. Selection varied prompt and sampling noise jointly. Cells below $\robSelFloor$ images were excluded, including $\robSelDropped$ Holdout~A cells. Supporting results are in Supplementary Notes~\ref{app:estimator-robustness} and~\ref{app:selection-control}.

\subsection{Statistical analysis and reporting}\label{subsec:statistics}
Difference tests were two-sided with a nominal significance level of $\alpha=0.05$; the equivalence analysis used two one-sided tests at the same level. Holm adjustment controlled family-wise error across the two primary zero-shot FID-RadImageNet comparisons. The reader study controlled the false discovery rate at $q=0.05$ across its eight source contrasts using Benjamini--Hochberg adjustment. Other comparisons were exploratory, with adjustment for tests within demographic cells as specified below. Confidence intervals were pointwise 95\% intervals without multiplicity adjustment. We report $p$-values with their comparisons and identify adjusted values explicitly. The reader and downstream reanalyses and zero-shot sensitivity analyses were specified after inspection of the existing results.

\paragraph{Replication and comparisons across trained models.}
Generation summaries are means and standard deviations across three independently trained models per method and modality, with fundus generation repeats first averaged within each model. Zero-shot comparisons used three models per method and holdout; the eight cell scores from a model were repeated outcomes. Classifier comparisons used four independent runs per generator and fixed corpus, giving 32 classifiers evaluated on both datasets. The datasets did not provide additional independent training replicates. Equal seed or run numbers were not treated as matched pairs.

For both experiments, we compared CompDiff with comparator $b$ within each stratum $g$ and weighted strata equally: $G=2$ holdouts for zero-shot generation and $G=4$ corpora for classifier training. Writing $\bar Y_{ga}$, $s_{ga}^2$ and $n_{ga}$ for the mean, variance and number of independent runs for method $a$ in stratum $g$,
\begin{equation}
\begin{aligned}
\widehat\Delta_b&=\frac{1}{G}\sum_{g=1}^{G}(\bar Y_{g,\mathrm{CD}}-\bar Y_{g,b}),\\
V_b&=\sum_{g,a}q_{ga},\qquad q_{ga}=\frac{s_{ga}^{2}}{G^2n_{ga}},\\
\nu_b&=\frac{V_b^2}{\sum_{g,a}q_{ga}^{2}/(n_{ga}-1)},\qquad a\in\{\mathrm{CD},b\}.
\end{aligned}
\label{eq:seed-contrast}
\end{equation}
Welch--Satterthwaite intervals were $\widehat\Delta_b\pm t_{0.975,\nu_b}\sqrt{V_b}$; zero-shot tests used $\widehat\Delta_b/\sqrt{V_b}$ with $\nu_b$ degrees of freedom. This approach allows different variances between methods and strata. Each zero-shot contrast used twelve trained models; classifier contrasts used all 32 classifiers. Inference conditions on the fixed holdouts, generated corpora and real test cohorts. It assumes independent, approximately normal run means; three or four runs per group do not permit a reliable assessment of normality.

Zero-shot sensitivity checks used a model with common variance; fixed effects for each combination of cell and method with CR1 covariance clustered by trained model and a $t$ distribution with degrees of freedom equal to the number of clusters minus one; and a percentile bootstrap resampling complete runs within method and holdout, enumerating $\scBootstrapDraws$ resamples. Permuting method labels for complete runs within holdouts enumerated $\scPermutations$ assignments per comparison. This requires exchangeability, a stronger assumption than equality of means with unequal variances. Additional checks omitted each model in turn, restricted CompDiff or all methods to seeds $\scSeedSecond$ and $\scSeedThird$, and repeated scoring with Inception features. Classifier sensitivity checks omitted each run in turn and included all available models descriptively. These checks examine sensitivity to modelling choices; they do not establish normality or replace the primary analysis.

\paragraph{Reader ratings and preferences.}\label{app:reader-stats}
Ratings were analysed separately for realism and correspondence with the mixed-effects model
\begin{equation}
y_{isr} = \mu + \beta_{s} + \gamma_{r} + u_{i} + \varepsilon_{isr},
\label{eq:reader-lmm}
\end{equation}
where $y_{isr}$ is the score from reader $r$ for source $s$ in set $i$, $\beta_s$ and $\gamma_r$ are fixed source and reader effects, $u_i\sim\mathcal{N}(\numMethodsGU{},\sigma^{\numMethodsGV{}}_{\mathrm{set}})$ is a random set intercept, and $\varepsilon_{isr}\sim\mathcal{N}(\numMethodsGW{},\sigma^{\numMethodsGX{}})$ is the residual. Models were fitted by restricted maximum likelihood. Source contrasts were estimated within sets. The two readers were treated as fixed because the panel was too small to estimate a variance across the reader population reliably.

The realism model included $\rsRealismN$ ratings from $\rsNSets$ image sets; the correspondence model included $\rsCorrN$ ratings from $\readerCorrSets$ sets. The model accounts for shared image sets and reader calibration. Its Gaussian random effects and residuals are working assumptions for the bounded rating scales, rather than empirically established normal distributions. We assessed sensitivity using sign-flip permutation tests on differences within reader and set ($\rsNPerm$ draws), Wilcoxon signed-rank tests, ordinary least squares with standard errors clustered by image set, and a model treating reader as a variance component. The paired checks assume symmetry of the differences under the null; their sample sizes are reported with the contrasts in Supplementary Table~\ref{tab:reader-robustness}.

The contrast family comprised all three generator pairs for each rating outcome and CompDiff against each baseline for forced choice. Rating comparisons used Wald tests and intervals. Forced-choice comparisons used exact binomial tests against equal preference among assessments containing both sources, conditional on one of them being selected: $n=50$ for CompDiff versus RoentGen-v2 and $n=79$ for CompDiff versus FairDiffusion. Preferences for a third source did not enter these tests. Realism equivalence between CompDiff and FairDiffusion used two one-sided $t$-tests on $\rsTOSTN$ paired assessments with a $\pm\rsTOSTMargin$ margin, chosen as a conventional half-point difference rather than a clinically established threshold. All eight source contrasts and the sensitivity results are reported in Supplementary Note~\ref{app:reader-results}.

\paragraph{Configuration summaries and patient resampling.}
The 48 fine-tuning configurations shared training data, pretraining weights and test cohorts. Label means, subgroup medians, interquartile ranges and counts of positive differences describe this fixed sweep. For the three MIMIC-CXR groups with fewer than ten positive images for at least one label, 1,000 bootstrap draws sampled patients with replacement within each group, retaining all images and model predictions for each sampled patient. Each draw recomputed the median of the 48 matched configuration differences. Draws lacking either class for any of the eight labels were undefined. Supplementary Table~\ref{tab:downstream-uncertainty} reports patient counts, the number of valid draws and their 2.5th--97.5th percentile range. These ranges describe sensitivity to patient sampling conditional on valid label coverage and fixed models; they do not establish nominal confidence coverage in these sparse groups.

\paragraph{Other comparisons.}
Welch's $t$-tests compared race accuracy and age error across three training runs per method. Tests within individual demographic cells required at least two available runs per method and used Welch's test with Benjamini--Hochberg adjustment at $q=0.05$ across cells, separately for each comparator. Standard-distribution intersectional fidelity comparisons paired method means within the sixteen cells with at least twenty images, using paired $t$-tests and Wilcoxon signed-rank checks. Pairing controls for cell identity; these cells share trained generators and are not independent training replicates. Approximate normality concerns the paired differences for the $t$-test, and symmetry concerns those differences for the signed-rank check. Composer comparisons used one training run per composer and regime and were interpreted against the final configuration's variation across training runs. Sampling-control comparisons used bootstrap intervals and Mann--Whitney tests as described in Section~\ref{subsec:robustness-protocol}. Audit summaries describe a single classifier; rare-cell AUROC intervals used $\numMethodsGT{}$ bootstrap iterations at $\numMethodsGS{}\%$ confidence, with estimation errors and interval widths summarised by rarity tier. Tier means do not quantify variation across training seeds.

AI tools assisted literature checking, manuscript restructuring, language revision and statistical reanalysis of existing experimental outputs. No additional training or image generation was performed for the reanalysis. The authors retain responsibility for the scientific content and its verification.

\section*{Data availability}
The source datasets are distributed by their respective custodians under their access and use conditions. MIMIC-CXR~\cite{johnson2019mimiccxr} is available to credentialed users through PhysioNet (\url{https://physionet.org/content/mimic-cxr/2.1.0/}) after completion of the required training and signature of its data use agreement. FairGenMed~\cite{friedrich2024fairdiffusion} is distributed by its authors for non-commercial research use under a CC~BY-NC-ND~4.0 licence (\url{https://huggingface.co/datasets/harvardairobotics/FairGenMed}). CheXpert~\cite{irvin2019chexpert}, used for external classifier evaluation, is described by Stanford at \url{https://stanfordmlgroup.github.io/competitions/chexpert/}, which provides the dataset access route. The synthetic training corpora used in the classifier experiments, a natural corpus of $\numMethodsCS{}$ chest X-ray images and a demographically balanced corpus of $\dsBalancedCD$ images, each with its generation prompt, demographic conditioning and finding labels, are deposited on the Hugging Face Hub at \HFMark\url{https://huggingface.co/datasets/mahmoudibra98/compdiff-chest-xray-synthetic} under a CC~BY-NC~4.0 licence with gated research-use access; other synthetic cohorts can be regenerated from the released weights. The de-identified per image reader study ratings are available from the corresponding author on reasonable request.

\section*{Code availability}
The CompDiff implementation is available at \GitHubMark\url{https://github.com/mahmoudibrahim98/CompDiff}. Pretrained model weights are released on the Hugging Face Hub: the chest X-ray model at \HFMark\url{https://huggingface.co/mahmoudibra98/compdiff-chest-xray} and the fundus model at \HFMark\url{https://huggingface.co/mahmoudibra98/compdiff-fundus}.

\section*{Acknowledgements}
This work was funded by the European Union under the Horizon Europe grant 101095435. Views and opinions expressed are however those of the author(s) only and do not necessarily reflect those of the European Union. Neither the European Union nor the granting authority can be held responsible for them.

\section*{Author contributions}
Mahmoud K. Ibrahim designed the study, developed the model, ran all experiments, produced the figures, and wrote the manuscript. C.S., M.D., B.E. and G.E. provided supervision and reviewed the manuscript. Maryam K. Ibrahim helped design and manage the radiologist reader study, which A.J. and M.M.S. conducted.

\section*{Competing interests}
The authors declare no competing interests.


\end{document}


\maketitle
\noindent This document contains supporting results and illustrations. The main Methods describes the experimental design and the statistical procedures; Supplementary Notes~\ref{app:zeroshot-inception}, \ref{app:reader-results} and~\ref{app:downstream-evidence} report the supporting zero-shot, reader-study and downstream results, respectively, and Note~\ref{app:clip-sensitivity} the prompt tokenisation and text-encoder probes. Figure and table numbers in this document form separate supplementary sequences; references to sections of the main paper and displays open the accompanying \texttt{main.pdf}.
\begin{figure}[tp]
\centering
\DisplayMark{fig:zeroshot-design/plot-top}
\begin{adjustbox}{max totalsize={\textwidth}{0.67\textheight},center}
\begin{minipage}{\textwidth}
\centering

\includegraphics[width=\textwidth]{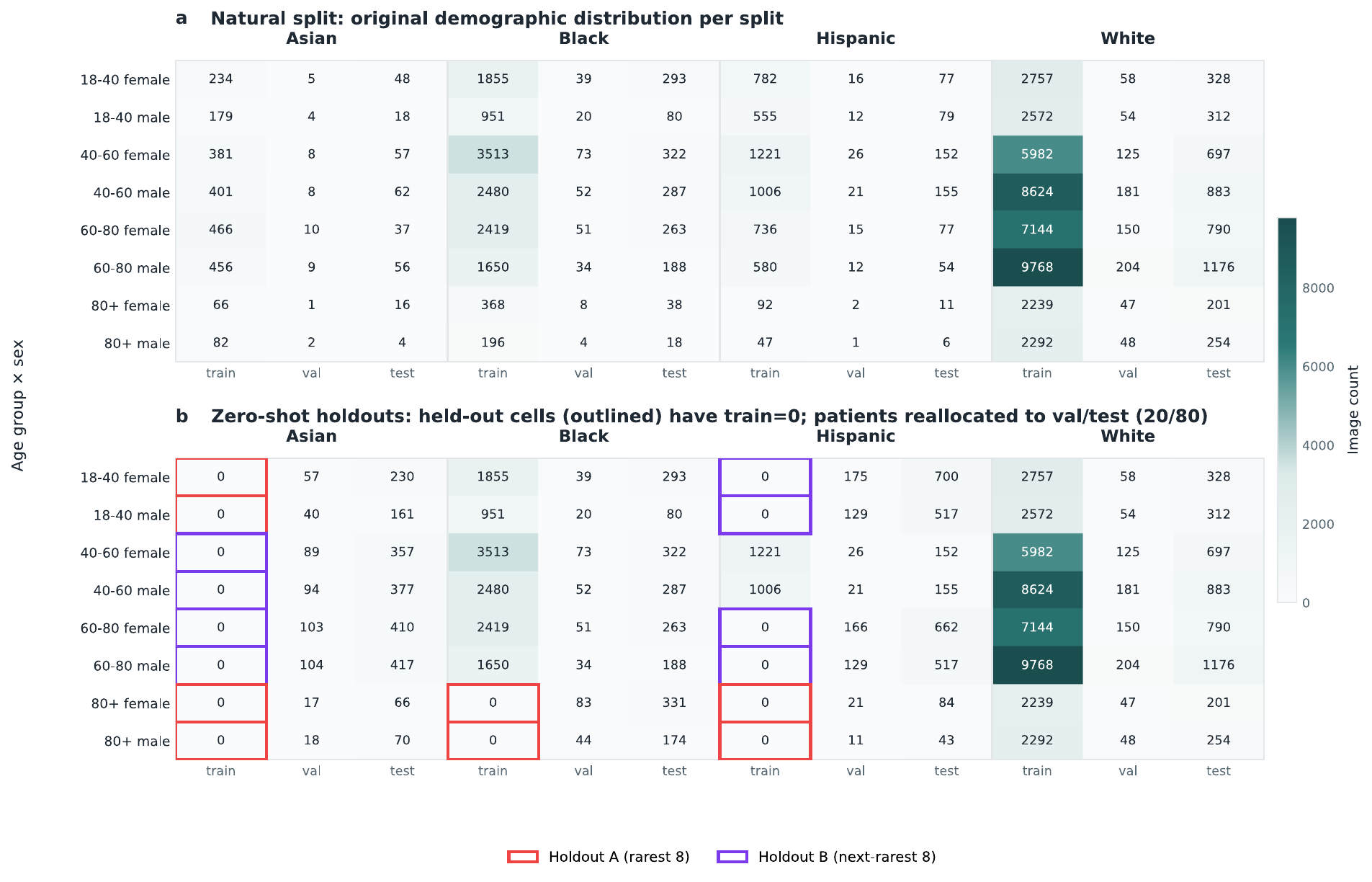}

\end{minipage}
\end{adjustbox}
\par\DisplayMark{fig:zeroshot-design/caption-top}
\caption{Two zero-shot withheld splits of the MIMIC-CXR distribution. Rows: age-bin $\times$ sex. Columns: race (outer) $\times$ split (inner). \textbf{a}, Natural distribution: image counts per age $\times$ sex $\times$ race $\times$ split intersection before any manipulation. \textbf{b}, Distribution after reallocation under the two zero-shot splits: each split removes eight disjoint three-way intersectional cells from train (count shown as $\numMainFiguresA{}$) and reallocates the affected patients, $\numMainFiguresB{}\%$ to test and $\numMainFiguresC{}\%$ to validation, inflating the val/test counts for the withheld cells. Red outlines mark the eight cells of Holdout~A (the rarest \numMainFiguresD{}, natural training count below $\numMainFiguresE{}$); violet outlines mark the eight cells of Holdout~B (the next rarest \numMainFiguresF{}, natural training count $\numMainFiguresG{}$--$\numMainFiguresH{}$). The two splits are disjoint, so each cell falls into exactly one category.}
\label{fig:zeroshot-design}
\par\DisplayMark{fig:zeroshot-design/caption-bottom}
\end{figure}

\clearpage
\FloatBarrier
\section{Zero-shot results across repeated training runs and sensitivity analyses}\label{app:zeroshot-inception}\label{app:seed-campaign}

\begin{figure}[htbp]
\centering
\DisplayMark{fig:zeroshot-cells/plot-top}
\includegraphics[width=\textwidth]{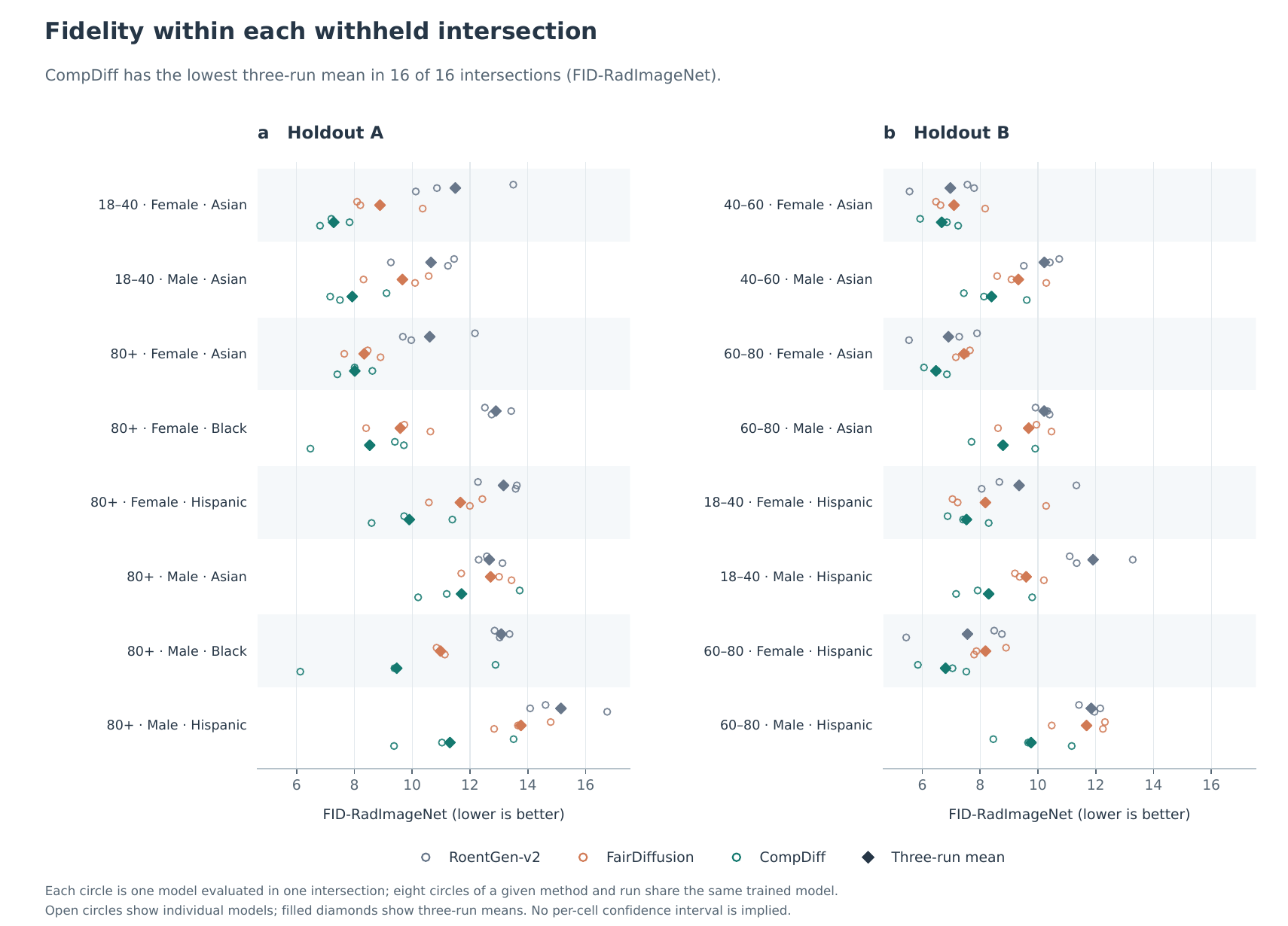}
\par\DisplayMark{fig:zeroshot-cells/caption-top}
\caption{FID-RadImageNet within each withheld intersection across repeated training runs. Panels a and b show the eight target intersections in Holdouts A and B, respectively. Colours identify the generators. Open circles show individual trained models and filled diamonds show the mean across three models for that method and intersection. The eight cell observations contributed by one model share that fitted generator and are not independent training replicates. Lower distances indicate closer real--synthetic feature distributions. CompDiff has the lowest mean across three runs in all $\scRNWins$ cells; these descriptive cell means do not imply that every individual run wins. No cell confidence interval is shown. Numerical means and standard deviations are in Supplementary Table~\ref{tab:seed-cells}.}
\label{fig:zeroshot-cells}
\par\DisplayMark{fig:zeroshot-cells/caption-bottom}
\end{figure}

The comparison across repeated training runs comprised $\scModels$ selected models: three methods, two holdouts and three runs per method and holdout. CompDiff had the lowest mean FID-RadImageNet across three runs in all $\scRNWins$ withheld intersections (Supplementary Figure~\ref{fig:zeroshot-cells}). Supplementary Table~\ref{tab:seed-cells} gives the cell means and standard deviations in both feature spaces.

\FloatBarrier
\begin{table}[htbp]
\centering
\caption{Fidelity at each withheld intersection across independently trained models.}
\label{tab:seed-cells}
\begingroup\small
\setlength{\tabcolsep}{4pt}
\begin{adjustbox}{max width=\textwidth}
\begin{tabular}{llrrr}
\toprule
Holdout & Intersection & CompDiff & FairDiffusion & RoentGen-v2 \\
\midrule
\multicolumn{5}{l}{\textbf{FID-RadImageNet}} \\
A & 18--40 Female Asian & $\mathbf{\scCellRNACDMean{}\pm\scCellRNACDSD{}}$ & $\scCellRNAFDMean{}\pm\scCellRNAFDSD{}$ & $\scCellRNARGMean{}\pm\scCellRNARGSD{}$ \\
A & 18--40 Male Asian & $\mathbf{\scCellRNBCDMean{}\pm\scCellRNBCDSD{}}$ & $\scCellRNBFDMean{}\pm\scCellRNBFDSD{}$ & $\scCellRNBRGMean{}\pm\scCellRNBRGSD{}$ \\
A & 80+ Female Asian & $\mathbf{\scCellRNCCDMean{}\pm\scCellRNCCDSD{}}$ & $\scCellRNCFDMean{}\pm\scCellRNCFDSD{}$ & $\scCellRNCRGMean{}\pm\scCellRNCRGSD{}$ \\
A & 80+ Female Black & $\mathbf{\scCellRNDCDMean{}\pm\scCellRNDCDSD{}}$ & $\scCellRNDFDMean{}\pm\scCellRNDFDSD{}$ & $\scCellRNDRGMean{}\pm\scCellRNDRGSD{}$ \\
A & 80+ Female Hispanic & $\mathbf{\scCellRNECDMean{}\pm\scCellRNECDSD{}}$ & $\scCellRNEFDMean{}\pm\scCellRNEFDSD{}$ & $\scCellRNERGMean{}\pm\scCellRNERGSD{}$ \\
A & 80+ Male Asian & $\mathbf{\scCellRNFCDMean{}\pm\scCellRNFCDSD{}}$ & $\scCellRNFFDMean{}\pm\scCellRNFFDSD{}$ & $\scCellRNFRGMean{}\pm\scCellRNFRGSD{}$ \\
A & 80+ Male Black & $\mathbf{\scCellRNGCDMean{}\pm\scCellRNGCDSD{}}$ & $\scCellRNGFDMean{}\pm\scCellRNGFDSD{}$ & $\scCellRNGRGMean{}\pm\scCellRNGRGSD{}$ \\
A & 80+ Male Hispanic & $\mathbf{\scCellRNHCDMean{}\pm\scCellRNHCDSD{}}$ & $\scCellRNHFDMean{}\pm\scCellRNHFDSD{}$ & $\scCellRNHRGMean{}\pm\scCellRNHRGSD{}$ \\
B & 18--40 Female Hispanic & $\mathbf{\scCellRNICDMean{}\pm\scCellRNICDSD{}}$ & $\scCellRNIFDMean{}\pm\scCellRNIFDSD{}$ & $\scCellRNIRGMean{}\pm\scCellRNIRGSD{}$ \\
B & 18--40 Male Hispanic & $\mathbf{\scCellRNJCDMean{}\pm\scCellRNJCDSD{}}$ & $\scCellRNJFDMean{}\pm\scCellRNJFDSD{}$ & $\scCellRNJRGMean{}\pm\scCellRNJRGSD{}$ \\
B & 40--60 Female Asian & $\mathbf{\scCellRNKCDMean{}\pm\scCellRNKCDSD{}}$ & $\scCellRNKFDMean{}\pm\scCellRNKFDSD{}$ & $\scCellRNKRGMean{}\pm\scCellRNKRGSD{}$ \\
B & 40--60 Male Asian & $\mathbf{\scCellRNLCDMean{}\pm\scCellRNLCDSD{}}$ & $\scCellRNLFDMean{}\pm\scCellRNLFDSD{}$ & $\scCellRNLRGMean{}\pm\scCellRNLRGSD{}$ \\
B & 60--80 Female Asian & $\mathbf{\scCellRNMCDMean{}\pm\scCellRNMCDSD{}}$ & $\scCellRNMFDMean{}\pm\scCellRNMFDSD{}$ & $\scCellRNMRGMean{}\pm\scCellRNMRGSD{}$ \\
B & 60--80 Female Hispanic & $\mathbf{\scCellRNNCDMean{}\pm\scCellRNNCDSD{}}$ & $\scCellRNNFDMean{}\pm\scCellRNNFDSD{}$ & $\scCellRNNRGMean{}\pm\scCellRNNRGSD{}$ \\
B & 60--80 Male Asian & $\mathbf{\scCellRNOCDMean{}\pm\scCellRNOCDSD{}}$ & $\scCellRNOFDMean{}\pm\scCellRNOFDSD{}$ & $\scCellRNORGMean{}\pm\scCellRNORGSD{}$ \\
B & 60--80 Male Hispanic & $\mathbf{\scCellRNPCDMean{}\pm\scCellRNPCDSD{}}$ & $\scCellRNPFDMean{}\pm\scCellRNPFDSD{}$ & $\scCellRNPRGMean{}\pm\scCellRNPRGSD{}$ \\
\midrule
\multicolumn{5}{l}{\textbf{Inception FID}} \\
A & 18--40 Female Asian & $\mathbf{\scCellFIDACDMean{}\pm\scCellFIDACDSD{}}$ & $\scCellFIDAFDMean{}\pm\scCellFIDAFDSD{}$ & $\scCellFIDARGMean{}\pm\scCellFIDARGSD{}$ \\
A & 18--40 Male Asian & $\mathbf{\scCellFIDBCDMean{}\pm\scCellFIDBCDSD{}}$ & $\scCellFIDBFDMean{}\pm\scCellFIDBFDSD{}$ & $\scCellFIDBRGMean{}\pm\scCellFIDBRGSD{}$ \\
A & 80+ Female Asian & $\mathbf{\scCellFIDCCDMean{}\pm\scCellFIDCCDSD{}}$ & $\scCellFIDCFDMean{}\pm\scCellFIDCFDSD{}$ & $\scCellFIDCRGMean{}\pm\scCellFIDCRGSD{}$ \\
A & 80+ Female Black & $\scCellFIDDCDMean{}\pm\scCellFIDDCDSD{}$ & $\mathbf{\scCellFIDDFDMean{}\pm\scCellFIDDFDSD{}}$ & $\scCellFIDDRGMean{}\pm\scCellFIDDRGSD{}$ \\
A & 80+ Female Hispanic & $\scCellFIDECDMean{}\pm\scCellFIDECDSD{}$ & $\mathbf{\scCellFIDEFDMean{}\pm\scCellFIDEFDSD{}}$ & $\scCellFIDERGMean{}\pm\scCellFIDERGSD{}$ \\
A & 80+ Male Asian & $\mathbf{\scCellFIDFCDMean{}\pm\scCellFIDFCDSD{}}$ & $\scCellFIDFFDMean{}\pm\scCellFIDFFDSD{}$ & $\scCellFIDFRGMean{}\pm\scCellFIDFRGSD{}$ \\
A & 80+ Male Black & $\scCellFIDGCDMean{}\pm\scCellFIDGCDSD{}$ & $\mathbf{\scCellFIDGFDMean{}\pm\scCellFIDGFDSD{}}$ & $\scCellFIDGRGMean{}\pm\scCellFIDGRGSD{}$ \\
A & 80+ Male Hispanic & $\mathbf{\scCellFIDHCDMean{}\pm\scCellFIDHCDSD{}}$ & $\scCellFIDHFDMean{}\pm\scCellFIDHFDSD{}$ & $\scCellFIDHRGMean{}\pm\scCellFIDHRGSD{}$ \\
B & 18--40 Female Hispanic & $\mathbf{\scCellFIDICDMean{}\pm\scCellFIDICDSD{}}$ & $\scCellFIDIFDMean{}\pm\scCellFIDIFDSD{}$ & $\scCellFIDIRGMean{}\pm\scCellFIDIRGSD{}$ \\
B & 18--40 Male Hispanic & $\mathbf{\scCellFIDJCDMean{}\pm\scCellFIDJCDSD{}}$ & $\scCellFIDJFDMean{}\pm\scCellFIDJFDSD{}$ & $\scCellFIDJRGMean{}\pm\scCellFIDJRGSD{}$ \\
B & 40--60 Female Asian & $\mathbf{\scCellFIDKCDMean{}\pm\scCellFIDKCDSD{}}$ & $\scCellFIDKFDMean{}\pm\scCellFIDKFDSD{}$ & $\scCellFIDKRGMean{}\pm\scCellFIDKRGSD{}$ \\
B & 40--60 Male Asian & $\mathbf{\scCellFIDLCDMean{}\pm\scCellFIDLCDSD{}}$ & $\scCellFIDLFDMean{}\pm\scCellFIDLFDSD{}$ & $\scCellFIDLRGMean{}\pm\scCellFIDLRGSD{}$ \\
B & 60--80 Female Asian & $\mathbf{\scCellFIDMCDMean{}\pm\scCellFIDMCDSD{}}$ & $\scCellFIDMFDMean{}\pm\scCellFIDMFDSD{}$ & $\scCellFIDMRGMean{}\pm\scCellFIDMRGSD{}$ \\
B & 60--80 Female Hispanic & $\mathbf{\scCellFIDNCDMean{}\pm\scCellFIDNCDSD{}}$ & $\scCellFIDNFDMean{}\pm\scCellFIDNFDSD{}$ & $\scCellFIDNRGMean{}\pm\scCellFIDNRGSD{}$ \\
B & 60--80 Male Asian & $\mathbf{\scCellFIDOCDMean{}\pm\scCellFIDOCDSD{}}$ & $\scCellFIDOFDMean{}\pm\scCellFIDOFDSD{}$ & $\scCellFIDORGMean{}\pm\scCellFIDORGSD{}$ \\
B & 60--80 Male Hispanic & $\scCellFIDPCDMean{}\pm\scCellFIDPCDSD{}$ & $\scCellFIDPFDMean{}\pm\scCellFIDPFDSD{}$ & $\mathbf{\scCellFIDPRGMean{}\pm\scCellFIDPRGSD{}}$ \\
\bottomrule
\end{tabular}
\end{adjustbox}
\par\smallskip
\begin{minipage}{\textwidth}\footnotesize Values are mean $\pm$ sample standard deviation across $\scRuns$ trained models per method and holdout. Bold indicates the lowest mean within an intersection and feature space. Distances compare synthetic and real images within the same cell; lower values indicate closer feature distributions.\end{minipage}
\endgroup
\end{table}

\FloatBarrier

\begin{figure}[htbp]
\centering
\DisplayMark{fig:zeroshot-inception/plot-top}
\includegraphics[width=\textwidth]{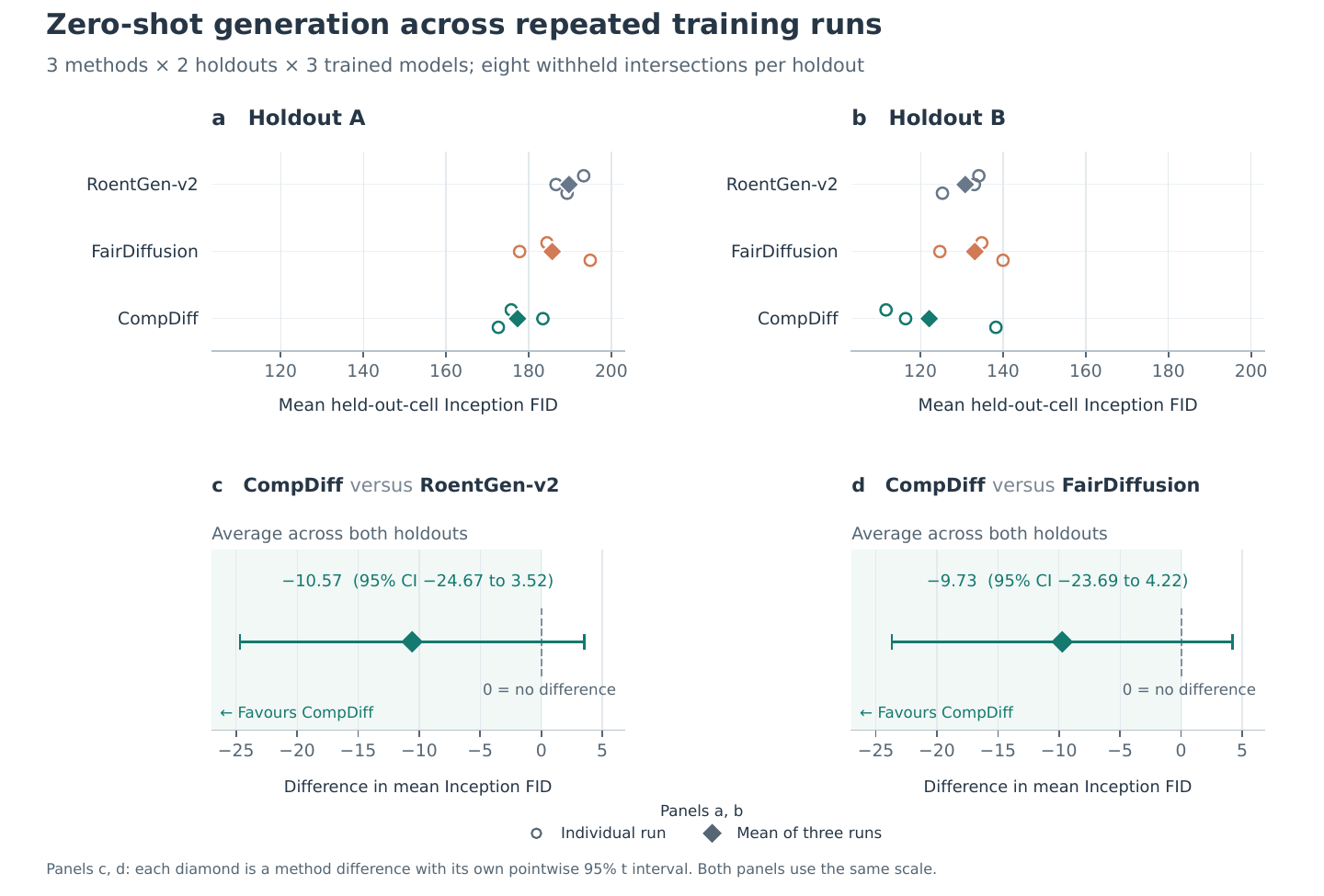}
\par\DisplayMark{fig:zeroshot-inception/caption-top}
\caption{Zero-shot comparison across repeated training runs using Inception-v3 features. This uses the same models and generated cohorts as Figure~\ref{fig:zeroshot}, changing only the feature backbone. \textbf{a,b}, Each open circle is one model's mean FID across its eight withheld intersections; filled diamonds show means across three independently trained models. \textbf{c,d}, CompDiff minus comparator differences averaged equally across cells and holdouts, with pointwise $\scLevel\%$ Welch--Satterthwaite confidence intervals. Negative differences favour CompDiff. Each difference has its own zero reference, and both panels use the same scale. Both intervals for the average differences include zero. Results for individual cells appear in Supplementary Figure~\ref{fig:zeroshot-cells-inception}.}
\label{fig:zeroshot-inception}
\par\DisplayMark{fig:zeroshot-inception/caption-bottom}
\end{figure}

With Inception features, CompDiff had the lowest mean across three runs in $\scFIDWins$ of $\scCells$ cells. Its average differences were $\scFIDBaseEffect$ against RoentGen-v2, with a pointwise $\scLevel\%$ interval of $[\scFIDBaseLo,\scFIDBaseHi]$, and $\scFIDFairEffect$ against FairDiffusion, with an interval of $[\scFIDFairLo,\scFIDFairHi]$. Both intervals included zero (Supplementary Figure~\ref{fig:zeroshot-inception}); the respective unadjusted two-sided $p$-values were $\scFIDBaseP$ and $\scFIDFairP$. The cell results in Supplementary Figure~\ref{fig:zeroshot-cells-inception} show where the feature backbones differ in their rankings; changing the backbone preserved the direction of the average effects, but not every cell's ordering.

\FloatBarrier
\begin{figure}[htbp]
\centering
\DisplayMark{fig:zeroshot-cells-inception/plot-top}
\includegraphics[width=\textwidth]{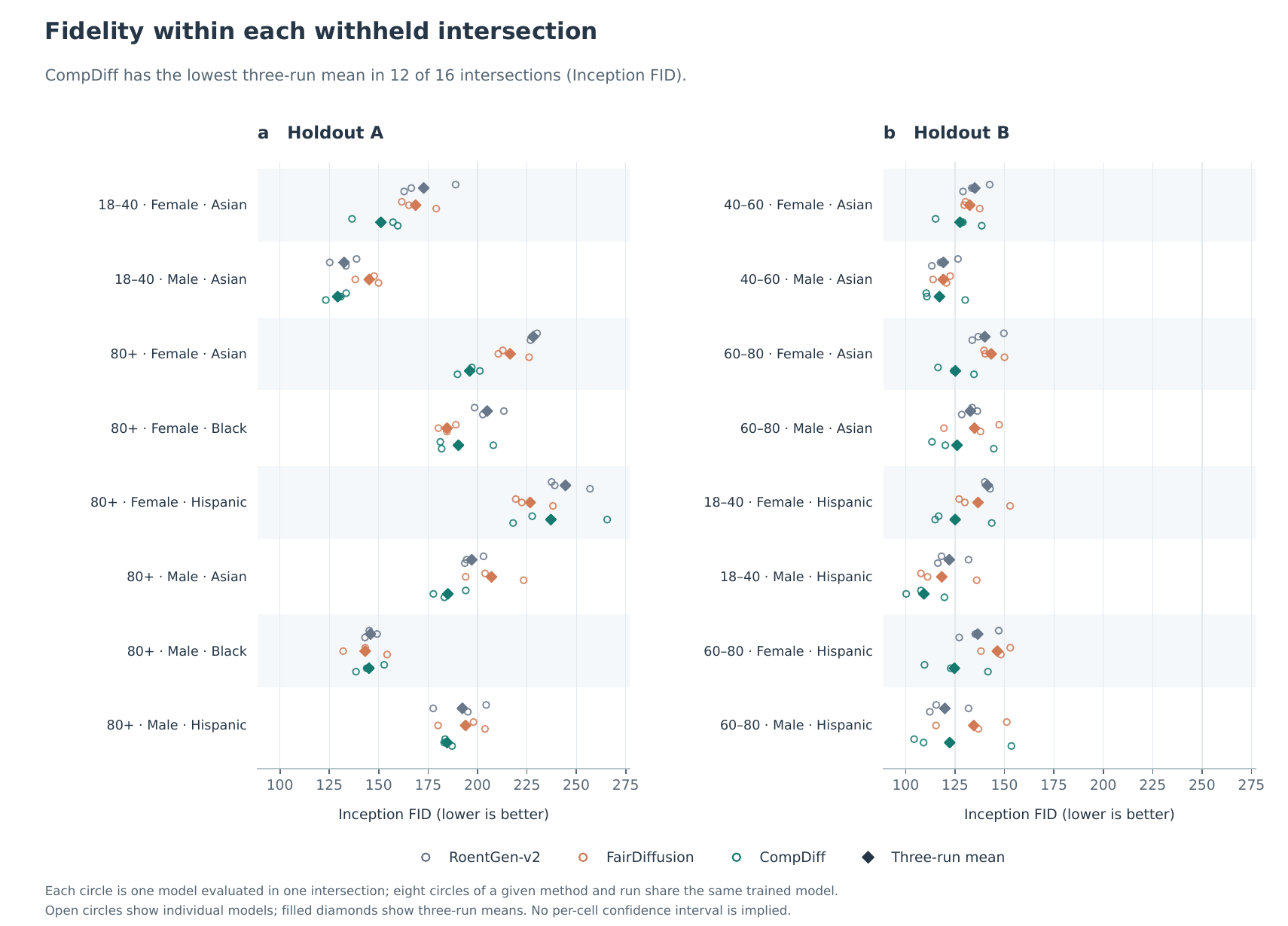}
\par\DisplayMark{fig:zeroshot-cells-inception/caption-top}
\caption{Inception FID within each withheld intersection across repeated training runs. This figure uses the same generated cohorts and conventions as Supplementary Figure~\ref{fig:zeroshot-cells}. Open circles show individual model scores and filled diamonds their means across three runs, separated by method within each cell. CompDiff has the lowest mean in $\scFIDWins$ of $\scCells$ cells. The corresponding average effects and intervals across trained models are shown in Supplementary Figure~\ref{fig:zeroshot-inception}; the cell observations are repeated outcomes from those same models, not additional independent replicates.}
\label{fig:zeroshot-cells-inception}
\par\DisplayMark{fig:zeroshot-cells-inception/caption-bottom}
\end{figure}

\FloatBarrier

For the average FID-RadImageNet differences, two-sided $p$-values were $\scRNBaseP$ against RoentGen-v2 and $\scRNFairP$ against FairDiffusion; after Holm adjustment across these two comparisons, they were $\scRNBaseHolm$ and $\scRNFairHolm$, respectively.

Across alternative variance models and subsets of runs, both average FID-RadImageNet differences favoured CompDiff. Intervals for RoentGen-v2 excluded zero throughout. For FairDiffusion, the primary interval and both subset intervals included zero, whereas intervals based on a common variance, covariance clustered by training run or the percentile bootstrap excluded it. These checks do not replace the primary analysis. Permutations of method labels for entire runs gave two-sided $p=\scRNBasePermutationP$ against RoentGen-v2 and $p=\scRNFairPermutationP$ against FairDiffusion under the stronger exchangeability assumption. Omitting any single trained model preserved the direction of both average differences, with estimates ranging from $\scRNBaseLooLo$ to $\scRNBaseLooHi$ against RoentGen-v2 and $\scRNFairLooLo$ to $\scRNFairLooHi$ against FairDiffusion.

Supplementary Table~\ref{tab:seed-runs} records every selected checkpoint. All selected test manifests passed checks for full split coverage and finite required metrics. Excluding the two CompDiff seed-$\scSeedFirst$ models, or restricting every method to the other two runs, preserved the direction of both FID-RadImageNet contrasts. The analysis targets the fixed cells and real test cohorts; it does not estimate uncertainty from resampling patients or choosing different holdout tasks.

\begin{table}[htbp]
\centering
\caption{Selected checkpoints for the zero-shot comparison across repeated training runs.}
\label{tab:seed-runs}
\begingroup\small
\setlength{\tabcolsep}{4pt}
\begin{adjustbox}{max width=\textwidth}
\begin{tabular}{llrr}
\toprule
Holdout & Generator & Training seed & Selected step \\
\midrule
A & CompDiff$\dagger$ & \scRunASeed{} & \scRunAStep{} \\
A & CompDiff & \scRunBSeed{} & \scRunBStep{} \\
A & CompDiff & \scRunCSeed{} & \scRunCStep{} \\
A & FairDiffusion & \scRunDSeed{} & \scRunDStep{} \\
A & FairDiffusion & \scRunESeed{} & \scRunEStep{} \\
A & FairDiffusion & \scRunFSeed{} & \scRunFStep{} \\
A & RoentGen-v2 & \scRunGSeed{} & \scRunGStep{} \\
A & RoentGen-v2 & \scRunHSeed{} & \scRunHStep{} \\
A & RoentGen-v2 & \scRunISeed{} & \scRunIStep{} \\
B & CompDiff$\dagger$ & \scRunJSeed{} & \scRunJStep{} \\
B & CompDiff & \scRunKSeed{} & \scRunKStep{} \\
B & CompDiff & \scRunLSeed{} & \scRunLStep{} \\
B & FairDiffusion & \scRunMSeed{} & \scRunMStep{} \\
B & FairDiffusion & \scRunNSeed{} & \scRunNStep{} \\
B & FairDiffusion & \scRunOSeed{} & \scRunOStep{} \\
B & RoentGen-v2 & \scRunPSeed{} & \scRunPStep{} \\
B & RoentGen-v2 & \scRunQSeed{} & \scRunQStep{} \\
B & RoentGen-v2 & \scRunRSeed{} & \scRunRStep{} \\
\bottomrule
\end{tabular}
\end{adjustbox}
\par\smallskip
\begin{minipage}{\textwidth}\footnotesize All runs used effective batch $\scBatch$ and a horizon for the cosine learning-rate schedule of $\scHorizon$ steps. Test evaluation had full split coverage for every selected model. $\dagger$ The two CompDiff seed-$\scSeedFirst$ runs used a different policy for validation coverage. For Holdout B, step $\scLegacyStep$ was retained although the validation minimum was at step $\scLegacyMinimum$, for which no test generation was available. Exclusion analyses are described in Supplementary Note~\ref{app:seed-campaign}. Run paths, manifest hashes and cell measurements are retained with the analysis source.\end{minipage}
\endgroup
\end{table}

\FloatBarrier

The comparison and sensitivity procedures are specified in Section~\ref{subsec:statistics}.

\FloatBarrier
\section{Robustness of the zero-shot ranking to the choice of fidelity estimator}
\label{app:estimator-robustness}

Cell evaluation places FID in the regime where it is least well behaved. The
estimator is biased upward at small sample sizes, and an intersectional cell is by
construction the smallest sample in the study: the rarest cells of Holdout~A contain
fewer than one hundred images. Because rarity and sample size are the same quantity
here, any table of cell fidelity risks confounding a property of the generator with
a property of the estimator. The following controls assess this effect in one selected model per configuration and holdout, including both composer variants; they do not repeat the full training-run comparison in Supplementary Note~\ref{app:seed-campaign}.

\subsection*{The estimator is size-dependent at fixed model and fixed data}

\begin{figure}[htbp]
\centering
\DisplayMark{fig:estimator-size-curve/plot-top}
\begin{adjustbox}{max totalsize={\textwidth}{0.67\textheight},center}
\begin{minipage}{\textwidth}
\centering

\includegraphics[width=0.55\textwidth]{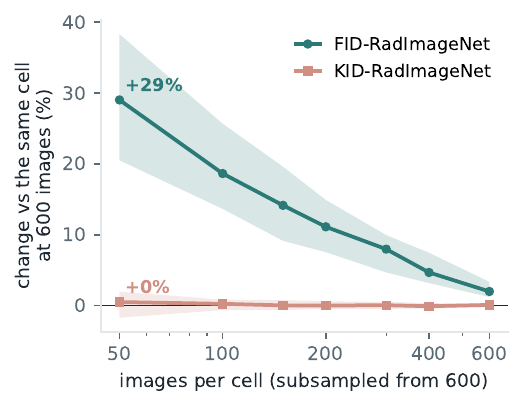}

\end{minipage}
\end{adjustbox}
\par\DisplayMark{fig:estimator-size-curve/caption-top}
\caption{The fidelity estimator is size-dependent; the kernel estimator is not. Both
metrics are read on the \emph{same} subsamples: each cell's candidate and reference
sets are reduced to a common size and the metric is expressed as a percentage change
against that cell's own reading at $\robSizeAnchor$ images. The generator, the cell and
the image pool are fixed, so any vertical movement is attributable to the estimator
alone. Lines are medians over the $\robSizeCombos$ cell $\times$ arm curves with the
interquartile range shaded; both series share one axis, so the difference in magnitude
can be read directly. FID-RadImageNet inflates as the cell shrinks, by a median
$\numSupplementFiguresE{}\%$ at $\numSupplementFiguresF{}$ images, while KID-RadImageNet~\cite{binkowski2018mmdgans}, which is
unbiased, stays at zero on the identical images. The ranking consequences of this bias
are given in Supplementary Table~\ref{tab:estimator-robustness}.}
\label{fig:estimator-size-curve}
\par\DisplayMark{fig:estimator-size-curve/caption-bottom}
\end{figure}

Relative to an anchor of $s=\robSizeAnchor$,
FID-RadImageNet inflates by \robInflOneFifty{} at $s=\numAppendixB{}$ and by \robInflSixty{} at
$s=\numAppendixC{}$ (absolute: $+\robInflOneFiftyAbs$ and $+\robInflSixtyAbs$ FID-RN). At $s=\numAppendixD{}$
the inflation is positive in all \robSizeCombos{} cell$\times$arm combinations
examined, ranging from \robInflMin{} to \robInflMax. The kernel counterpart KID,
which is an unbiased estimator~\cite{binkowski2018mmdgans}, is flat on the identical
subsamples (mean drift \robKIDDriftA{} and \robKIDDriftB{} on Holdouts~A and~B,
positive in \robKIDPosFrac{} combinations, i.e.\ scatter without direction). The
contrast between the two metrics on the same images is direct evidence that the
inflation is an artefact of the estimator rather than a degradation of the samples,
and Supplementary Figure~\ref{fig:estimator-size-curve} shows it directly.

\subsection*{Extrapolation does not recover values for small cells}

A natural remedy is to extrapolate FID to infinite sample size~\cite{chong2020unbiasedfid}. In this setting the remedy is only
partial. Fitting FID-RN against $\numAppendixE{}/k$ on large cells is well conditioned in the
aggregate ($R^2=\robRsqK$ median, slope $\approx\robSlopeK$, coefficient of variation
across cells \robSlopeCV, so the bias is close to universal rather than
specific to each cell), and the reference side carries $\robSlopeRatio\times$ the slope of the
candidate side, confirming that both arms of the comparison contribute. The fit is
nonetheless misspecified near the sample sizes that matter: the local slope at
$k\!\approx\!n$ is $\robCurv\times$ the wide-range fitted slope
($\robSlopeLocal$ versus $\robSlopeK$), because every cell here has $k$ far below the
feature dimension $d=\robFeatDim$. Validated against a withheld target on
\robValCombos{} large cell$\times$arm combinations, extrapolation from a
$\robValN$-image subsample recovers only \robRecovered{} of the gap to the cell's
own reading from the full sample and retains a systematic $+\robValBias$ FID-RN overestimate
(relative MAE \robValRelMAE, versus \robValRelMAERaw{} for the uncorrected
subsample; SD across runs $\robValSD$). We therefore report corrected values as a
robustness check only, and do not use them in place of the primary measurements.

\subsection*{Rankings within each cell are largely preserved across estimators}

Supplementary Table~\ref{tab:estimator-robustness} compares the arm rankings across estimators.

\begin{table}[!htbp]
\centering
\caption{Robustness of the zero-shot arm ranking to the fidelity estimator. Each row
rescores every cell with a different estimator and compares the resulting cell
arm ordering against the ordering under the primary metric. Agreement is reported over
all cells and over the withheld intersectional cells alone. Estimators: FID$_\infty$
extrapolates to infinite sample size in the candidate dimension ($1/k$) or jointly in
candidate and reference ($1/k,1/m$); FID at common $n$ rescores every cell at a shared
sample size; KID is an unbiased kernel estimator. Ordering agreement counts require the
complete arm ordering to match.}
\label{tab:estimator-robustness}
\noindent{\fontsize{8}{9.6}\selectfont
\begin{adjustbox}{max width=\textwidth}\begin{tabular}{lcccc}
\toprule
\textbf{Estimator} & \textbf{Same ordering} & \textbf{Same winner} & \textbf{Withheld: same ordering} & \textbf{Median $\rho$} \\
\midrule
FID$_\infty$ ($1/k$)        & \robOrderInfK/\robCells & \robCells/\robCells      & \robHOCells/\robHOCells   & \robCellRhoInfK \\
FID$_\infty$ ($1/k$, $1/m$) & \robOrderInfJ/\robCells & \robWinnerInfJ/\robCells & \robHOOrderInfJ/\robHOCells & \robRhoInfJ \\
KID                          & \robOrderKID/\robCells  & \robWinnerKID/\robCells  & \robHOOrderKID/\robHOCells  & \robCellRhoKID \\
\bottomrule
\end{tabular}\end{adjustbox}
}
\end{table}

The conclusions drawn in Section~\ref{subsec:zeroshot-protocol} are comparisons
\emph{between arms within a cell}, evaluated at a common sample size. A size-dependent
bias that is shared by all arms in a cell cancels in such a comparison, and the data
confirm this. Across the \robCells{} cells for which all arms are available, the
arm ordering is identical under $\numAppendixF{}/k$-corrected FID in \robOrderInfK{} cells and
under the joint $(\numAppendixG{}/k,\numAppendixH{}/m)$ correction in \robOrderInfJ{} cells; the winning arm is
unchanged in all \robCells{} cells under the former and in \robWinnerInfJ{} under the latter. Restricted to the withheld intersectional cells large
enough to subsample (the \robHOCells{} cells with at least \robSelFloor{} images; the four
smallest Holdout~A cells fall below that floor and are excluded here exactly as in
\AppendixName~\ref{app:selection-control}), both the ordering and the winner are preserved in
\robHOOrderInfJ{} of \robHOCells{} cells, with median Spearman correlation within each cell $\rho=\robRhoInfJ$. Under KID,
an estimator with a different bias structure entirely, the winner is unchanged in
\robWinnerKID{} of \robCells{} cells and the full ordering in \robOrderKID.
Restricting attention to the three generators of Figure~\ref{fig:zeroshot} can only
increase these agreement rates, since agreement on a total order implies agreement on
each of its suborders.

\subsection*{Fidelity comparisons between cells are not supported by these data}

The same analysis places a limit on what cell tables can be asked to show. Sample
size explains only a median $R^2=\robRsqRaw$ of the variation in FID-RN \emph{across}
cells, and rescoring every cell at a common sample size removes just
\robVarRemoved{} of that variation: the spread between cells is therefore not
predominantly a size artefact either. But the direction of the remaining spread does
not support a rare-cell deficit. Comparing withheld against well represented cells
within each split$\times$arm panel, the withheld cells read \emph{better} by a median
of \robGapRaw{} under the primary metric, \robGapInfJ{} under joint extrapolation,
\robGapDBC{} at common sample size and \robGapKID{} under KID; the withheld group is
the worse of the two in at most \robPanelsWorseRaw{} of \robPanels{} panels under any
estimator, and in \robPanelsWorseKID{} under KID.

We read this as a limitation of the comparison rather than as evidence that rare cells
are generated better. A rare intersection may have a narrower and more homogeneous real
reference distribution, which can be an easier target to match irrespective of generator
behaviour, and the reference side of the estimator is precisely where the residual
size sensitivity is largest. The defensible conclusion is negative and methodological:
fidelity within each cell should be compared \emph{between arms within a cell at matched sample
size}, as in Section~\ref{subsec:zeroshot-protocol} and
Figure~\ref{fig:zeroshot}, and differences between cells should not be interpreted as
differences in generation quality.

\FloatBarrier
\section{Post hoc sample selection does not substitute for demographic conditioning}
\label{app:selection-control}

A conditioning architecture earns its complexity only if the same benefit cannot be
obtained by sampling more and discarding the worse samples. We test this directly.
The selection control is specified in Section~\ref{subsec:robustness-protocol}.

Selection works, and it works for every arm. Over \robSelSigCells{} withheld
cell$\times$arm combinations the improvement is positive with a bootstrap confidence
interval excluding zero in all of them, with a median relative gain of \robSelHO. It
is, however, neither specific to rare cells nor large: well represented cells improve
by a statistically indistinguishable \robSelCommon{} (Mann--Whitney $p=\robSelP$),
two thirds of the gain is already present at a $\numAppendixI{}\times$ budget (\robSelTwoX), and an
oracle selector that is allowed to see the evaluation references reaches only
\robSelCeil, of which the deployable verifier already captures \robSelCapture.

The direct comparison tests whether selection can replace the conditioner. Giving
RoentGen-v2 a $\robSelBudget$ selection budget and giving CompDiff none, the selected
baseline attains a lower FID-RN in \robSelBaseWins{} of \robSelHOCells{}
withheld cells. Ranking all arm$\times$mode configurations jointly, a
configuration with a dedicated demographic conditioner is best in \robSelCondWins{} of
\robSelHOCells{} cells, and no prompt-conditioned arm, with or without selection,
is best in any cell. Sampling compute and demographic conditioning are therefore not
interchangeable: a fourfold increase in sampling budget does not recover what routing
demographics through a dedicated demographic conditioner provides, and the two compose rather than
compete.

The best configuration varies between the factorised composer and the unstructured composer with a matched parameter count. Both preserve the separate attribute encoders and supervision of the output tokens. This agrees with the composer control: the benefit does not require pairwise factorisation, provided the encoders and supervised output tokens are retained.

Two limitations bound this experiment. Selection operates over the pool of samples
actually generated, so it varies prompt and sampling noise jointly and does not isolate
the contribution of the noise draw. And cells below \robSelFloor{} images do not admit
a stable contrast between selected and random samples, which excludes \robSelDropped{} of the eight
Holdout~A withheld cells, the rarest in the study, from this analysis. The conclusions therefore apply to the intersections with sufficient samples for this
selection analysis.

\FloatBarrier
\section{Zero-shot generalisation with an unstructured composer of matched size}\label{app:flat-composer}

The composer control of Section~\ref{subsec:composer-control} supports the mechanism analysis in Section~\ref{subsec:ablations}. It compares one selected model per composer and holdout.

\begin{figure}[htbp]
\centering
\DisplayMark{fig:composer-zeroshot/plot-top}
\begin{adjustbox}{max totalsize={\textwidth}{0.67\textheight},center}
\begin{minipage}{\textwidth}
\centering

\includegraphics[width=\textwidth]{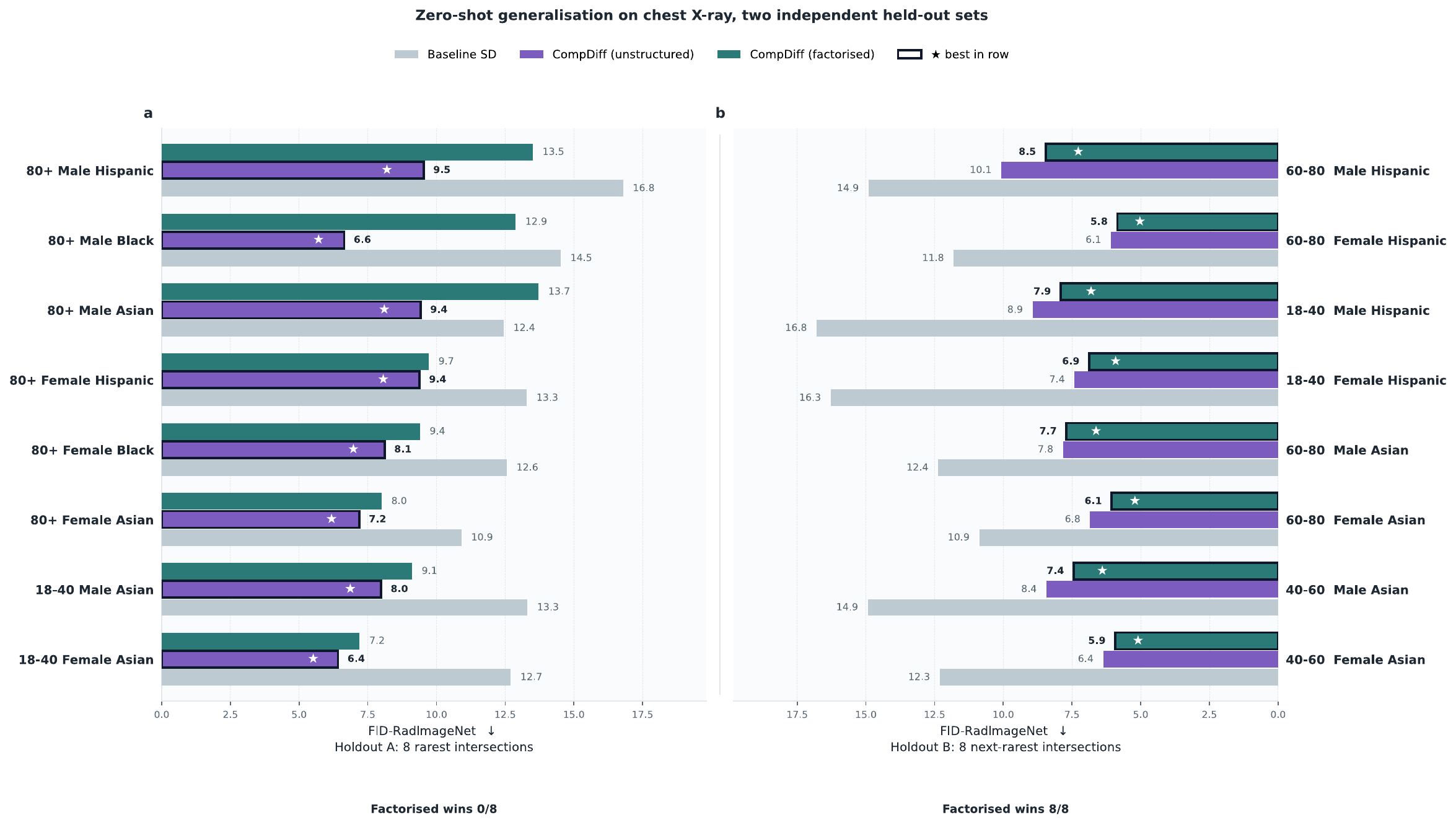}

\end{minipage}
\end{adjustbox}
\par\DisplayMark{fig:composer-zeroshot/caption-top}
\caption{Fidelity of the two composers at unseen intersections. \textbf{a}, Holdout~A, the eight rarest intersections. \textbf{b}, Holdout~B, the eight next-rarest intersections. Both panels show FID-RadImageNet within each cell (lower is better) for the factorised composer (the seed-$\scSeedFirst$ CompDiff models in Supplementary Table~\ref{tab:seed-runs}), the unstructured composer with a matched parameter count, and the RoentGen-v2 baseline as a scale reference; FairDiffusion is omitted because this figure answers a composer question. Every arm is scored at full coverage of its split ($\numSupplementFiguresG{}$ and $\numSupplementFiguresH{}$ images). Checkpoint selection is specified in Section~\ref{subsec:composer-control}. The footer counts wins for the \emph{factorised} arm against the other two arms shown, so $\numSupplementFiguresI{}/\numSupplementFiguresJ{}$ on Holdout~A means the unstructured control attains the lowest value in all eight of those cells, and $\numSupplementFiguresK{}/\numSupplementFiguresL{}$ on Holdout~B the reverse; the sign of the composer difference does not reproduce across the two splits.}
\label{fig:composer-zeroshot}
\par\DisplayMark{fig:composer-zeroshot/caption-bottom}
\end{figure}

Supplementary Figure~\ref{fig:composer-zeroshot} plots FID-RadImageNet within each cell for the two composers on all \numAppendixJ{} withheld cells, with the prompt-conditioned baseline retained to give the scale on which both composers sit. The pattern is the dissociation summarised in the main text: the unstructured composer attains the lowest value on every Holdout~A cell and the factorised composer on every Holdout~B cell, while both stay well below the baseline throughout. The gap between composers is comparable to the variability across independent training runs of the final configuration (FID-RadImageNet spread \ablSpreadRN{}; Supplementary Table~\ref{tab:composer-paired-test}), so no effect of composer topology reproduces across the two splits in either direction. The same holds at the validation layer, where both arms are scored under an identical protocol: across the \ablValPairings{} pairings that Holdout~B admits (matched by training step, and each arm's checkpoint with the best validation score under either selection criterion), the largest gap between composers in withheld mean FID-RadImageNet is \ablValMaxGap. Supplementary Table~\ref{tab:steps-to-floor} reports the one consistent difference, the convergence speed of demographic control. This is a comparison of composer variants; Figure~\ref{fig:zeroshot} compares three independently trained models per method and holdout.

\begin{table}[!htbp]
\centering
\caption{Paired composer control on the full data corpus, evaluated on the withheld \emph{test} split at a matched step of $10{,}000$ (all $7{,}039$ test images, mean $\pm$ SD over three generation seeds). The two arms were trained together under an identical recipe and differ only in the composer. The last row is the spread across three independent training runs of the final configuration, the scale against which these differences should be read.}
\label{tab:composer-paired-test}
\noindent{\fontsize{8}{9.6}\selectfont
\begin{adjustbox}{max width=\textwidth}\begin{tabular}{@{}lcccccc@{}}
\toprule
\textbf{Composer} & \textbf{FID}$\downarrow$ & \textbf{FID-RN}$\downarrow$ & \textbf{Sex}$\uparrow$ & \textbf{Race}$\uparrow$ & \textbf{Age}$\downarrow$ & \textbf{AUROC}$\uparrow$ \\
\midrule
Factorised & 62.56 $\pm$ 0.16 & 6.08 & 0.998 & 0.962 & 8.84 $\pm$ 0.06 & 0.78 \\
Unstructured (param.-matched) & 61.99 $\pm$ 0.15 & 7.05 & 0.995 & 0.955 & 10.85 $\pm$ 0.10 & 0.79 \\
\midrule
\emph{Spread across training runs} & 3.18 & 2.09 & -- & -- & 4.23 & -- \\
\bottomrule
\end{tabular}\end{adjustbox}
}
\end{table}

\begin{table}[!htbp]
\centering
\caption{First checkpoint at which each full data run satisfies all three controllability floors of Eq.~\ref{eq:ckpt-rule}. The three factorised runs are independent training runs of the final configuration.}
\label{tab:steps-to-floor}
\noindent{\fontsize{8}{9.6}\selectfont
\begin{adjustbox}{max width=\textwidth}\begin{tabular}{@{}lc@{}}
\toprule
\textbf{Run} & \textbf{All floors first met} \\
\midrule
Factorised, run 1 & 10{,}000 \\
Factorised, run 2 & 7{,}500 \\
Factorised, run 3 (matched pair) & 8{,}500 \\
Unstructured (parameter-matched) & 11{,}000 \\
\bottomrule
\end{tabular}\end{adjustbox}
}
\end{table}

\FloatBarrier

\FloatBarrier
\section{Reader study contrasts and sensitivity analyses}\label{app:reader-results}
The reader protocol is described in Section~\ref{subsec:reader-protocol}, and the statistical analyses in Section~\ref{app:reader-stats}. Supplementary Table~\ref{tab:reader-contrasts} reports the complete contrast family; Supplementary Table~\ref{tab:reader-robustness} reports the sensitivity analyses.

\begin{table}[!htbp]
\centering
\caption{Mixed-effects contrasts on the blinded reader study. Rating estimates are differences in mean score from Equation~\eqref{eq:reader-lmm}, with two-sided Wald tests and 95\% intervals. Realism uses $\rsRealismN$ ratings from $\rsNSets$ sets; correspondence uses $\rsCorrN$ ratings from $\readerCorrSets$ sets. Both models include the two readers. $p_{\mathrm{BH}}$ is Benjamini--Hochberg-adjusted across the $\rsFamilySize$ comparisons. Forced-choice rows give selections for each source among assessments containing both. Two-sided exact binomial tests condition on selection of either compared source, giving 50 and 79 trials, respectively; preferences for a third source are excluded. Numerical $p$-values are shown to three significant digits.}
\label{tab:reader-contrasts}
\noindent{\fontsize{8}{9.6}\selectfont
\begin{tabular}{llcccc}
\toprule
\textbf{Outcome} & \textbf{Contrast} & $\Delta$ & \textbf{95\% CI} & $p$ & $p_{\mathrm{BH}}$ \\
\midrule
Realism & CompDiff vs.\ RoentGen-v2 & $+1.81$ & $[+1.48,\,+2.15]$ & $4.20\times10^{-26}$ & $3.36\times10^{-25}$ \\
 & CompDiff vs.\ FairDiffusion & $+0.04$ & $[-0.29,\,+0.37]$ & $0.812$ & $0.812$ \\
 & FairDiffusion vs.\ RoentGen-v2 & $+1.77$ & $[+1.44,\,+2.10]$ & $1.14\times10^{-25}$ & $4.56\times10^{-25}$ \\
\addlinespace
Prompt corr. & CompDiff vs.\ RoentGen-v2 & $+0.55$ & $[+0.20,\,+0.91]$ & $0.00236$ & $0.00472$ \\
 & CompDiff vs.\ FairDiffusion & $+0.42$ & $[+0.07,\,+0.77]$ & $0.0191$ & $0.0306$ \\
 & FairDiffusion vs.\ RoentGen-v2 & $+0.13$ & $[-0.21,\,+0.48]$ & $0.45$ & $0.514$ \\
\addlinespace
Most realistic & CompDiff vs.\ RoentGen-v2 & 44--6 & \footnotesize (preferences) & $3.24\times10^{-8}$ & $8.65\times10^{-8}$ \\
 & CompDiff vs.\ FairDiffusion & 44--35 & \footnotesize (preferences) & $0.368$ & $0.491$ \\
\bottomrule
\end{tabular}

}
\end{table}

\begin{table}[!htbp]
\centering
\caption{Robustness of the reader study contrasts. $\Delta$ and the Wald column come from the mixed-effects model of Equation~\eqref{eq:reader-lmm}. The paired sample size counts reader--set assessments with both source ratings and applies to the permutation and Wilcoxon tests. The permutation column uses $\rsNPerm$ sign-flip draws; the Wilcoxon column uses signed ranks. The Cluster column gives Wald tests from ordinary least squares with standard errors clustered by image set. The mixed and ordinary least-squares models use 287 realism ratings from 92 sets or 277 correspondence ratings from 89 sets. All tests are two-sided; $p$-values are unadjusted and shown to three significant digits.}
\label{tab:reader-robustness}
\noindent{\fontsize{8}{9.6}\selectfont
\setlength{\tabcolsep}{3.5pt}
\begin{tabular}{llcccccc}
\toprule
\textbf{Outcome} & \textbf{Contrast} & $\Delta$ & $n_{\mathrm{paired}}$ & \textbf{Wald} & \textbf{Permutation} & \textbf{Wilcoxon} & \textbf{Cluster} \\
 & & & & $p$ & $p$ & $p$ & $p$ \\
\midrule
Realism & CompDiff vs.\ RoentGen-v2 & $+1.81$ & 84 & $4.20\times10^{-26}$ & $1.00\times10^{-4}$ & $2.34\times10^{-10}$ & $7.34\times10^{-21}$ \\
 & CompDiff vs.\ FairDiffusion & $+0.04$ & 89 & $0.812$ & $0.673$ & $0.608$ & $0.807$ \\
 & FairDiffusion vs.\ RoentGen-v2 & $+1.77$ & 89 & $1.14\times10^{-25}$ & $1.00\times10^{-4}$ & $2.97\times10^{-10}$ & $4.38\times10^{-19}$ \\
\addlinespace
Prompt corr. & CompDiff vs.\ RoentGen-v2 & $+0.55$ & 80 & $0.00236$ & $0.0132$ & $0.012$ & $0.0121$ \\
 & CompDiff vs.\ FairDiffusion & $+0.42$ & 85 & $0.0191$ & $0.0498$ & $0.0458$ & $0.011$ \\
 & FairDiffusion vs.\ RoentGen-v2 & $+0.13$ & 86 & $0.45$ & $0.52$ & $0.48$ & $0.699$ \\
\bottomrule
\end{tabular}

}
\end{table}

The source comparisons are consistent across the sensitivity analyses. The CompDiff--FairDiffusion correspondence contrast moves from $p = \numAppendixAH{}$ under the mixed-effects model to $p = \readerCorrFairPerm$ under the permutation test and is the weaker of the two correspondence results. Treating reader as a variance component reproduces every contrast to two decimal places.

The realism difference between CompDiff and FairDiffusion is $\rsRealCDFDEst$ points (\numAppendixAJ{}\% confidence interval $[\rsRealCDFDLo, \rsRealCDFDHi]$; $p = \rsRealCDFDP$). The equivalence test gives $p = \rsTOSTP$ on $\rsTOSTN$ paired sets, supporting equivalence within the analytical margin. For correspondence, the variance between sets is $\rsVarSetCorr$ against a residual of $\rsVarResidCorr$ (intraclass correlation $\rsICCCorr$). For realism, the set variance is estimated at zero against a residual of $\rsVarResidRealism$. These differences explain why the pooled correspondence standard deviations overstate the uncertainty of within-set comparisons and why the permutation and cluster-robust checks support the realism inference.

The $\rsNReal$ hidden real anchors occur in different sets. Their correspondence mean adjusted by the model is unstable; Table~\ref{tab:reader-study} therefore reports the raw mean. The realism and forced-choice anchor comparisons are unaffected.

\FloatBarrier
\section{Reader study rating interface}\label{app:reader-interface}

\begin{figure}[htbp]
\centering
\DisplayMark{fig:reader-interface/plot-top}
\begin{adjustbox}{max totalsize={\textwidth}{0.67\textheight},center}
\begin{minipage}{\textwidth}
\centering

\includegraphics[width=0.97\textwidth]{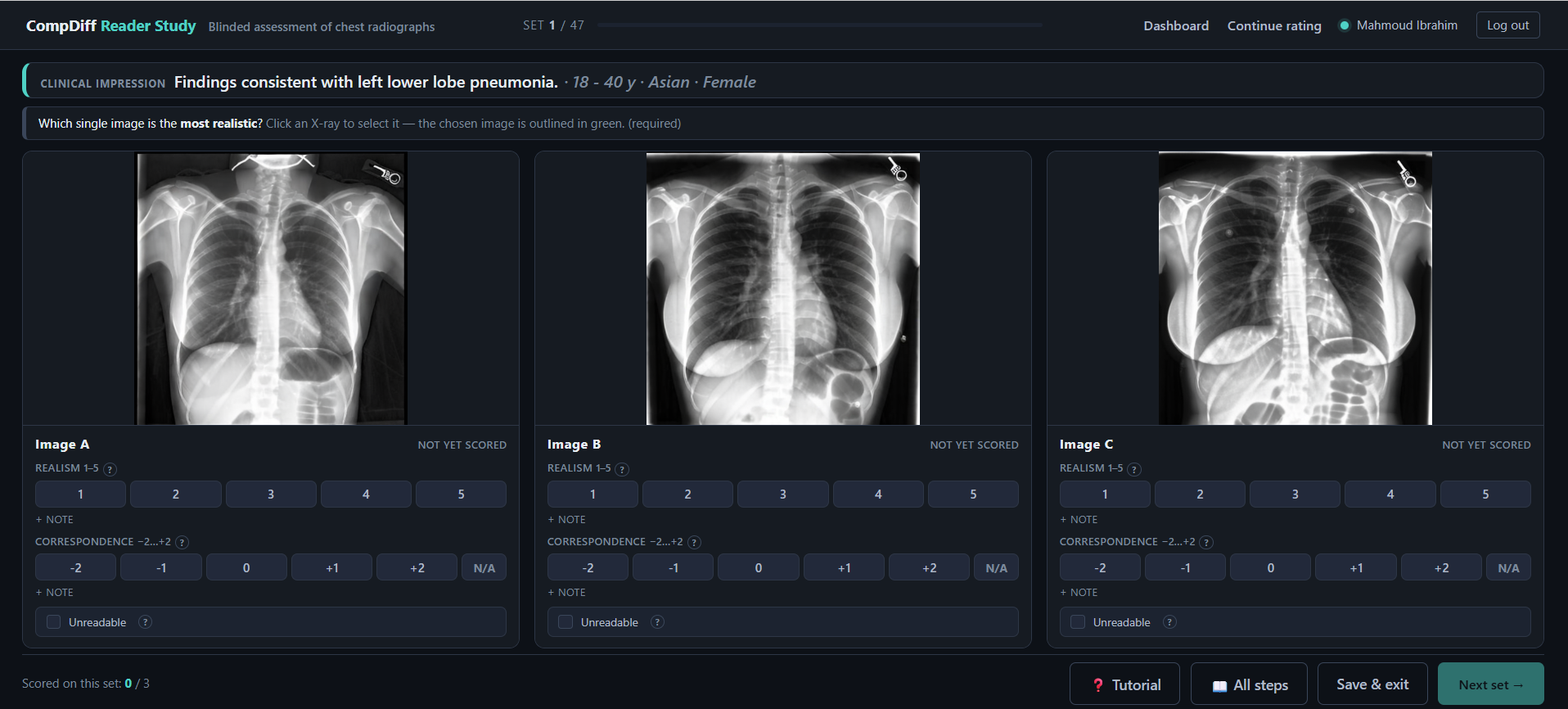}

\end{minipage}
\end{adjustbox}
\par\DisplayMark{fig:reader-interface/caption-top}
\caption{The blinded reader study rating interface (Section~\ref{subsec:reader-study}). The clinical impression is displayed at the top, followed by three chest X-rays displayed side by side in randomised order; each image carries an anatomical realism control ($\numSupplementFiguresV{}$--$\numSupplementFiguresW{}$) and a prompt correspondence control ($-\numSupplementFiguresX{}$ to $+\numSupplementFiguresY{}$), with a forced ``most realistic'' choice below the row and unreadable and N/A options for each image. Raters are blinded to which images are synthetic or real and in what proportion.}
\label{fig:reader-interface}
\par\DisplayMark{fig:reader-interface/caption-bottom}
\end{figure}

Supplementary Figure~\ref{fig:reader-interface} reproduces the viewer used in the blinded reader protocol (Section~\ref{subsec:reader-protocol}).

\FloatBarrier
\section{Validation trajectories}\label{app:checkpoint-results}
Supplementary Tables~\ref{tab:ckpt-baseline}--\ref{tab:ckpt-compdiff} report the complete validation trajectories, with the checkpoints selected by Equation~\eqref{eq:ckpt-rule} marked $\star$.

\paragraph{Overfitting within the shared budget.} All three generators degrade well inside the shared budget. Each reaches its minimum validation FID within the first $\numAppendixAK{}$ steps (CompDiff at step $\ckptCDargmin$, Baseline at $\ckptBasesel$, FairDiffusion at $\ckptFairDsel$); by step $\numAppendixAL{}$ every generator exceeds an FID of $\numAppendixAM{}$, and by step $\numAppendixAN{}$ each sits at $\numAppendixAO{}$--$\numAppendixAP{}\times$ its own minimum, with MS-SSIM roughly halved (from $\numAppendixAQ{}$--$\numAppendixAR{}$ at step $\numAppendixAS{}$ to $\numAppendixAT{}$--$\numAppendixAU{}$) and disease AUROC below its peak during training for all three (Supplementary Tables~\ref{tab:ckpt-baseline}--\ref{tab:ckpt-compdiff}). Because every trajectory is logged to the end of the budget, each selected checkpoint is a minimum over the complete trajectory rather than the best point of a truncated one.

\begin{table}[!htbp]
\centering
\caption{Validation trajectory for the \textbf{Baseline SD} generator over the shared $30{,}000$-step budget (training run with seed~11). This run is one of the three independently trained Baseline SD models whose selected checkpoints are averaged into the Baseline SD bars of the main text; the values tabulated here are those of this single run, not the means across three runs, and its selected checkpoint is marked $\star$. All three generators are scored at the identical 24 checkpoints. The selected checkpoint (step 8,500) is the checkpoint with the lowest FID among those meeting the demographic control floors of Section~\ref{app:checkpoint-selection}; rows marked $\dagger$ fail at least one floor and are therefore not eligible, regardless of their FID. Metrics follow Section~\ref{subsec:eval-synth} (FID-RIN = FID-RadImageNet; MS-SSIM as defined there, read against the $0.25$--$0.75$ band; Sex/Race = prompt alignment accuracy; Age RMSE in years) and are computed on the validation split.}
\label{tab:ckpt-baseline}
\noindent{\fontsize{8}{9.6}\selectfont
\begin{adjustbox}{max width=\textwidth}\begin{tabular}{rccccccc}
\toprule
\textbf{Step} & \textbf{FID}\,$\downarrow$ & \textbf{FID-RIN}\,$\downarrow$ & \textbf{MS-SSIM} & \textbf{AUROC}\,$\uparrow$ & \textbf{Sex}\,$\uparrow$ & \textbf{Race}\,$\uparrow$ & \textbf{Age RMSE}\,$\downarrow$ \\
\midrule
1,000 & 96.71 & 8.08 & 0.34 & 0.63 & 1.000 & 0.968 & 4.73 \\
2,500 & 96.18 & 9.85 & 0.32 & 0.75 & 0.999 & 0.972 & 4.34 \\
3,500 & 98.21 & 9.93 & 0.30 & 0.70 & 0.999 & 0.977 & 5.71 \\
5,000 & 91.33 & 9.55 & 0.31 & 0.69 & 0.997 & 0.981 & 5.13 \\
6,000 & 92.43 & 8.84 & 0.32 & 0.69 & 0.998 & 0.986 & 5.29 \\
7,500 & 90.67 & 10.32 & 0.29 & 0.73 & 0.994 & 0.972 & 5.63 \\
\textbf{$\star$~8,500} & \textbf{89.99} & \textbf{8.63} & \textbf{0.31} & \textbf{0.72} & \textbf{0.989} & \textbf{0.972} & \textbf{7.70} \\
10,000 & 92.88 & 9.47 & 0.30 & 0.76 & 0.995 & 0.969 & 6.67 \\
11,000 & 96.67 & 10.84 & 0.27 & 0.78 & 0.991 & 0.979 & 8.53 \\
12,500 & 99.95 & 11.77 & 0.28 & 0.70 & 0.990 & 0.970 & 7.17 \\
13,500 & 105.81 & 12.35 & 0.26 & 0.70 & 0.985 & 0.961 & 9.26 \\
15,000 & 107.51 & 12.54 & 0.25 & 0.65 & 0.988 & 0.969 & 8.27 \\
16,000 & 108.82 & 12.81 & 0.26 & 0.63 & 0.991 & 0.979 & 9.03 \\
17,500$^{\dagger}$ & 109.97 & 12.71 & 0.24 & 0.63 & 0.976 & 0.946 & 12.20 \\
18,500$^{\dagger}$ & 109.82 & 14.47 & 0.24 & 0.58 & 0.975 & 0.953 & 9.42 \\
20,000$^{\dagger}$ & 140.94 & 17.59 & 0.18 & 0.62 & 0.984 & 0.962 & 10.18 \\
21,000 & 143.33 & 18.68 & 0.16 & 0.56 & 0.995 & 0.981 & 8.03 \\
22,500$^{\dagger}$ & 152.85 & 18.41 & 0.16 & 0.53 & 0.975 & 0.937 & 13.54 \\
23,500$^{\dagger}$ & 149.92 & 18.70 & 0.15 & 0.61 & 0.978 & 0.924 & 14.63 \\
25,000$^{\dagger}$ & 148.03 & 18.76 & 0.16 & 0.57 & 0.979 & 0.934 & 12.56 \\
26,000$^{\dagger}$ & 150.76 & 18.77 & 0.16 & 0.56 & 0.983 & 0.930 & 11.65 \\
27,500 & 125.83 & 14.92 & 0.21 & 0.65 & 0.985 & 0.969 & 8.90 \\
28,500$^{\dagger}$ & 150.81 & 19.65 & 0.15 & 0.56 & 0.985 & 0.959 & 10.30 \\
30,000$^{\dagger}$ & 150.03 & 18.23 & 0.16 & 0.64 & 0.978 & 0.923 & 10.50 \\
\bottomrule
\end{tabular}\end{adjustbox}
}
\end{table}

\begin{table}[!htbp]
\centering
\caption{Validation trajectory for the \textbf{FairDiffusion} generator over the shared $30{,}000$-step budget (training run with seed~42). This run is one of the three independently trained FairDiffusion models whose selected checkpoints are averaged into the FairDiffusion bars of the main text; the values tabulated here are those of this single run, not the means across three runs, and its selected checkpoint is marked $\star$. All three generators are scored at the identical 24 checkpoints. The selected checkpoint (step 11,000) is the checkpoint with the lowest FID among those meeting the demographic control floors of Section~\ref{app:checkpoint-selection}; rows marked $\dagger$ fail at least one floor and are therefore not eligible, regardless of their FID. Metrics follow Section~\ref{subsec:eval-synth} (FID-RIN = FID-RadImageNet; MS-SSIM as defined there, read against the $0.25$--$0.75$ band; Sex/Race = prompt alignment accuracy; Age RMSE in years) and are computed on the validation split.}
\label{tab:ckpt-fairdiffusion}
\noindent{\fontsize{8}{9.6}\selectfont
\begin{adjustbox}{max width=\textwidth}\begin{tabular}{rccccccc}
\toprule
\textbf{Step} & \textbf{FID}\,$\downarrow$ & \textbf{FID-RIN}\,$\downarrow$ & \textbf{MS-SSIM} & \textbf{AUROC}\,$\uparrow$ & \textbf{Sex}\,$\uparrow$ & \textbf{Race}\,$\uparrow$ & \textbf{Age RMSE}\,$\downarrow$ \\
\midrule
1,000$^{\dagger}$ & 110.99 & 6.74 & 0.36 & 0.61 & 0.998 & 0.902 & 7.83 \\
2,500 & 94.71 & 8.38 & 0.35 & 0.64 & 1.000 & 0.967 & 5.83 \\
3,500 & 100.23 & 8.48 & 0.32 & 0.70 & 0.999 & 0.972 & 9.12 \\
5,000 & 95.55 & 7.74 & 0.35 & 0.65 & 0.999 & 0.962 & 7.96 \\
6,000 & 94.57 & 8.72 & 0.35 & 0.68 & 0.998 & 0.979 & 6.22 \\
7,500 & 90.74 & 8.40 & 0.34 & 0.67 & 0.997 & 0.980 & 7.29 \\
8,500 & 94.64 & 8.89 & 0.34 & 0.68 & 0.998 & 0.970 & 7.58 \\
10,000 & 98.90 & 10.05 & 0.32 & 0.75 & 0.998 & 0.976 & 7.81 \\
\textbf{$\star$~11,000} & \textbf{86.93} & \textbf{9.56} & \textbf{0.33} & \textbf{0.73} & \textbf{0.999} & \textbf{0.981} & \textbf{5.56} \\
12,500 & 98.14 & 9.85 & 0.32 & 0.71 & 0.998 & 0.977 & 7.79 \\
13,500 & 100.64 & 11.24 & 0.31 & 0.74 & 0.999 & 0.976 & 5.55 \\
15,000 & 99.00 & 11.74 & 0.29 & 0.72 & 0.999 & 0.982 & 6.07 \\
16,000 & 104.32 & 13.95 & 0.27 & 0.72 & 0.996 & 0.977 & 7.52 \\
17,500 & 115.92 & 15.21 & 0.25 & 0.57 & 1.000 & 0.987 & 6.32 \\
18,500 & 97.70 & 12.22 & 0.28 & 0.73 & 0.996 & 0.982 & 8.12 \\
20,000 & 112.69 & 14.30 & 0.25 & 0.64 & 0.994 & 0.983 & 7.20 \\
21,000 & 123.85 & 15.63 & 0.22 & 0.62 & 0.985 & 0.963 & 9.40 \\
22,500 & 117.35 & 14.81 & 0.24 & 0.61 & 0.988 & 0.979 & 9.05 \\
23,500 & 119.03 & 15.18 & 0.23 & 0.65 & 0.991 & 0.971 & 9.30 \\
25,000 & 119.25 & 15.54 & 0.23 & 0.65 & 0.996 & 0.973 & 7.09 \\
26,000 & 132.83 & 18.18 & 0.20 & 0.59 & 0.991 & 0.960 & 8.22 \\
27,500 & 115.29 & 14.25 & 0.24 & 0.65 & 0.993 & 0.981 & 7.67 \\
28,500 & 150.26 & 20.06 & 0.17 & 0.53 & 0.992 & 0.965 & 7.76 \\
30,000$^{\dagger}$ & 123.66 & 16.88 & 0.19 & 0.65 & 0.980 & 0.970 & 10.26 \\
\bottomrule
\end{tabular}\end{adjustbox}
}
\end{table}

\begin{table}[!htbp]
\centering
\caption{Validation trajectory for the \textbf{CompDiff} generator over the shared $30{,}000$-step budget (training run with seed~42). This run is one of the three independently trained CompDiff models whose selected checkpoints are averaged into the CompDiff bars of the main text; the values tabulated here are those of this single run, not the means across three runs, and its selected checkpoint is marked $\star$. All three generators are scored at the identical 24 checkpoints. The selected checkpoint (step 10,000) is the checkpoint with the lowest FID among those meeting the demographic control floors of Section~\ref{app:checkpoint-selection}; rows marked $\dagger$ fail at least one floor and are therefore not eligible, regardless of their FID. Metrics follow Section~\ref{subsec:eval-synth} (FID-RIN = FID-RadImageNet; MS-SSIM as defined there, read against the $0.25$--$0.75$ band; Sex/Race = prompt alignment accuracy; Age RMSE in years) and are computed on the validation split.}
\label{tab:ckpt-compdiff}
\noindent{\fontsize{8}{9.6}\selectfont
\begin{adjustbox}{max width=\textwidth}\begin{tabular}{rccccccc}
\toprule
\textbf{Step} & \textbf{FID}\,$\downarrow$ & \textbf{FID-RIN}\,$\downarrow$ & \textbf{MS-SSIM} & \textbf{AUROC}\,$\uparrow$ & \textbf{Sex}\,$\uparrow$ & \textbf{Race}\,$\uparrow$ & \textbf{Age RMSE}\,$\downarrow$ \\
\midrule
1,000$^{\dagger}$ & 78.39 & 5.89 & 0.35 & 0.64 & 0.552 & 0.663 & 17.75 \\
2,500$^{\dagger}$ & 65.76 & 4.69 & 0.37 & 0.70 & 0.872 & 0.868 & 14.58 \\
3,500$^{\dagger}$ & 63.53 & 4.79 & 0.37 & 0.73 & 0.983 & 0.911 & 14.45 \\
5,000$^{\dagger}$ & 66.74 & 5.68 & 0.36 & 0.71 & 0.993 & 0.923 & 13.58 \\
6,000$^{\dagger}$ & 69.01 & 6.01 & 0.36 & 0.77 & 0.992 & 0.921 & 13.23 \\
7,500$^{\dagger}$ & 71.00 & 7.02 & 0.37 & 0.76 & 0.997 & 0.960 & 13.35 \\
8,500$^{\dagger}$ & 75.17 & 7.33 & 0.35 & 0.71 & 0.998 & 0.961 & 10.39 \\
\textbf{$\star$~10,000} & \textbf{71.63} & \textbf{6.06} & \textbf{0.38} & \textbf{0.75} & \textbf{0.999} & \textbf{0.956} & \textbf{8.96} \\
11,000 & 82.56 & 8.07 & 0.33 & 0.80 & 0.999 & 0.966 & 8.98 \\
12,500 & 87.10 & 9.33 & 0.31 & 0.74 & 1.000 & 0.977 & 8.54 \\
13,500 & 102.08 & 12.70 & 0.25 & 0.77 & 0.985 & 0.967 & 8.97 \\
15,000 & 106.69 & 13.19 & 0.25 & 0.73 & 0.993 & 0.962 & 8.36 \\
16,000 & 103.41 & 12.57 & 0.26 & 0.70 & 0.998 & 0.966 & 8.01 \\
17,500 & 99.30 & 11.15 & 0.26 & 0.70 & 0.992 & 0.965 & 9.09 \\
18,500 & 121.15 & 14.87 & 0.22 & 0.64 & 1.000 & 0.977 & 7.66 \\
20,000$^{\dagger}$ & 115.68 & 13.82 & 0.23 & 0.62 & 0.983 & 0.975 & 10.28 \\
21,000 & 110.41 & 14.20 & 0.23 & 0.62 & 0.996 & 0.974 & 9.31 \\
22,500$^{\dagger}$ & 115.63 & 16.08 & 0.19 & 0.63 & 0.982 & 0.969 & 10.55 \\
23,500 & 137.83 & 18.65 & 0.17 & 0.63 & 0.991 & 0.980 & 9.97 \\
25,000 & 127.50 & 17.50 & 0.17 & 0.58 & 0.990 & 0.972 & 9.44 \\
26,000$^{\dagger}$ & 135.04 & 17.93 & 0.16 & 0.61 & 0.980 & 0.975 & 10.02 \\
27,500$^{\dagger}$ & 120.47 & 16.03 & 0.19 & 0.56 & 0.972 & 0.972 & 11.26 \\
28,500$^{\dagger}$ & 116.95 & 16.42 & 0.18 & 0.67 & 0.988 & 0.987 & 10.95 \\
30,000$^{\dagger}$ & 137.40 & 17.16 & 0.18 & 0.59 & 0.981 & 0.970 & 11.64 \\
\bottomrule
\end{tabular}\end{adjustbox}
}
\end{table}

\FloatBarrier

\section{Downstream diagnostic and subgroup evidence}\label{app:downstream-evidence}

The analyses in Sections~\ref{subsec:downstream-recipes} and~\ref{subsec:statistics} use a consistent set of eight diagnostic labels throughout: atelectasis, cardiomegaly, consolidation, oedema, pleural effusion, pneumonia, pneumothorax and no finding. Repeated classifier runs measure variability in classifier training on fixed generated corpora; they do not represent independently trained generators.

Inference and sensitivity procedures are specified in Section~\ref{subsec:statistics}.

\begin{figure*}[!htbp]
\centering
\DisplayMark{fig:downstream-synthetic-runs/plot-top}
\begin{adjustbox}{max totalsize={\textwidth}{.67\textheight},center}
\includegraphics[width=\textwidth]{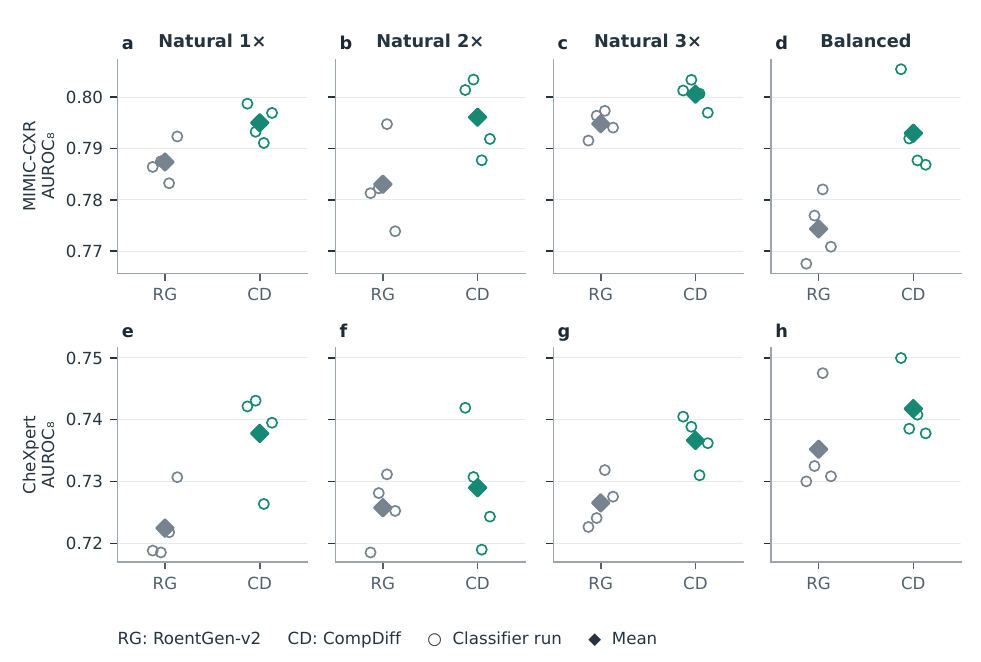}
\end{adjustbox}
\par\DisplayMark{fig:downstream-synthetic-runs/caption-top}
\caption{Independent classifier runs using synthetic chest images alone. \textbf{a}--\textbf{d}, Evaluation on MIMIC-CXR for the natural 1$\times$, 2$\times$ and 3$\times$ corpora and the balanced corpus. \textbf{e}--\textbf{h}, The same classifiers evaluated on CheXpert, in the same order. Each panel shows four classifier runs per generator for one fixed corpus regime under the AdamW recipe. Hollow circles are individual runs and filled diamonds are their means. RG denotes RoentGen-v2 and CD denotes CompDiff. AUROC$_8$ uses the same eight diagnostic labels as Figure~\ref{fig:downstream-d3deep}. The same 32 trained classifiers are evaluated on MIMIC-CXR and CheXpert. Natural corpora contain one, two or three times the real training set size; the balanced corpus samples demographic intersections uniformly.}
\label{fig:downstream-synthetic-runs}
\par\DisplayMark{fig:downstream-synthetic-runs/caption-bottom}
\end{figure*}

\paragraph{Repeated training and the complete configuration sweep.}
For classifiers trained on synthetic images alone under the verified common AdamW recipe, CompDiff had the higher mean AUROC$_8$ in every corpus regime on both test datasets (Supplementary Figure~\ref{fig:downstream-synthetic-runs}). Omitting any single classifier run retained a positive average contrast on each dataset. Supplementary Table~\ref{tab:downstream-runs} compares the three training strategies using AdamW. Mixing real and synthetic images in one training stage did not yield a consistent advantage.

\begin{table}[!htbp]
\centering
\caption{Repeated classifier training using AdamW and a consistent set of eight diagnostic labels.}
\label{tab:downstream-runs}
\begingroup\small
\setlength{\tabcolsep}{4pt}
\begin{adjustbox}{max width=\textwidth}
\begin{tabular}{llrrrr}
\toprule
Training & Test set & RoentGen-v2 & CompDiff & Difference & 95\% interval \\
\midrule
Synthetic alone & MIMIC-CXR & 0.7849 & 0.7962 & $+0.0113$ & $[0.0067, 0.0158]$ \\
Synthetic alone & CheXpert & 0.7275 & 0.7363 & $+0.0088$ & $[0.0039, 0.0137]$ \\
Pretrain, fine-tune & MIMIC-CXR & 0.8577 & 0.8591 & $+0.0013$ & $[-0.0026, 0.0053]$ \\
Pretrain, fine-tune & CheXpert & 0.7810 & 0.7839 & $+0.0029$ & $[-0.0007, 0.0066]$ \\
Mixed training & MIMIC-CXR & 0.8457 & 0.8467 & $+0.0010$ & $[-0.0006, 0.0026]$ \\
Mixed training & CheXpert & 0.7588 & 0.7528 & $-0.0061$ & $[-0.0096, -0.0025]$ \\
\bottomrule
\end{tabular}
\end{adjustbox}
\par\smallskip
\begin{minipage}{\textwidth}\footnotesize Each row averages four synthetic corpus regimes equally, with four independent classifier runs per generator and regime. The same classifiers are evaluated on both test sets. Intervals are pointwise Welch intervals across classifier runs, conditional on the corpora and test cohorts; no pairing by run index is assumed. Training uses $256\times256$ inputs, batch size 128, an initial learning rate of 0.001 and up to 50 epochs; test inputs are $224\times224$. Pretraining followed by fine-tuning uses ImageNet initialisation and all real training images; mixed training combines real and synthetic images in one stage. Differences are CompDiff minus RoentGen-v2. These additional comparisons are exploratory and intervals are not adjusted for multiplicity.\end{minipage}
\endgroup
\end{table}

Including all five available models from the AdamW family, with the additional model that could inherit a different recipe, also retained positive average differences and positive corpus means on both datasets (Supplementary Table~\ref{tab:downstream-all-models}).

\begin{table}[!htbp]
\centering
\caption{Sensitivity to including all available classifiers trained on synthetic images alone within the AdamW family.}
\label{tab:downstream-all-models}
\begingroup\small
\setlength{\tabcolsep}{4pt}
\begin{adjustbox}{max width=\textwidth}
\begin{tabular}{lrrrr}
\toprule
Test set & RoentGen-v2 & CompDiff & Difference & Positive corpora / 4 \\
\midrule
MIMIC-CXR & 0.7864 & 0.7949 & $+0.0086$ & 4 \\
CheXpert & 0.7282 & 0.7365 & $+0.0083$ & 4 \\
\bottomrule
\end{tabular}
\end{adjustbox}
\par\smallskip
\begin{minipage}{\textwidth}\footnotesize Descriptive averages include all five available models per generator and corpus, including the additional model that may come from an earlier training recipe. The four corpora receive equal weight. These models are not treated as five exchangeable repeats of the same training protocol; the main intervals use the four runs with a verified common recipe. Including the additional models retains positive average differences and positive corpus means for CompDiff under the AdamW family on both test sets.\end{minipage}
\endgroup
\end{table}

\FloatBarrier

\begin{figure*}[!htbp]
\centering
\DisplayMark{fig:downstream-sweep-mimic/plot-top}
\begin{adjustbox}{max totalsize={\textwidth}{.67\textheight},center}
\includegraphics[width=\textwidth]{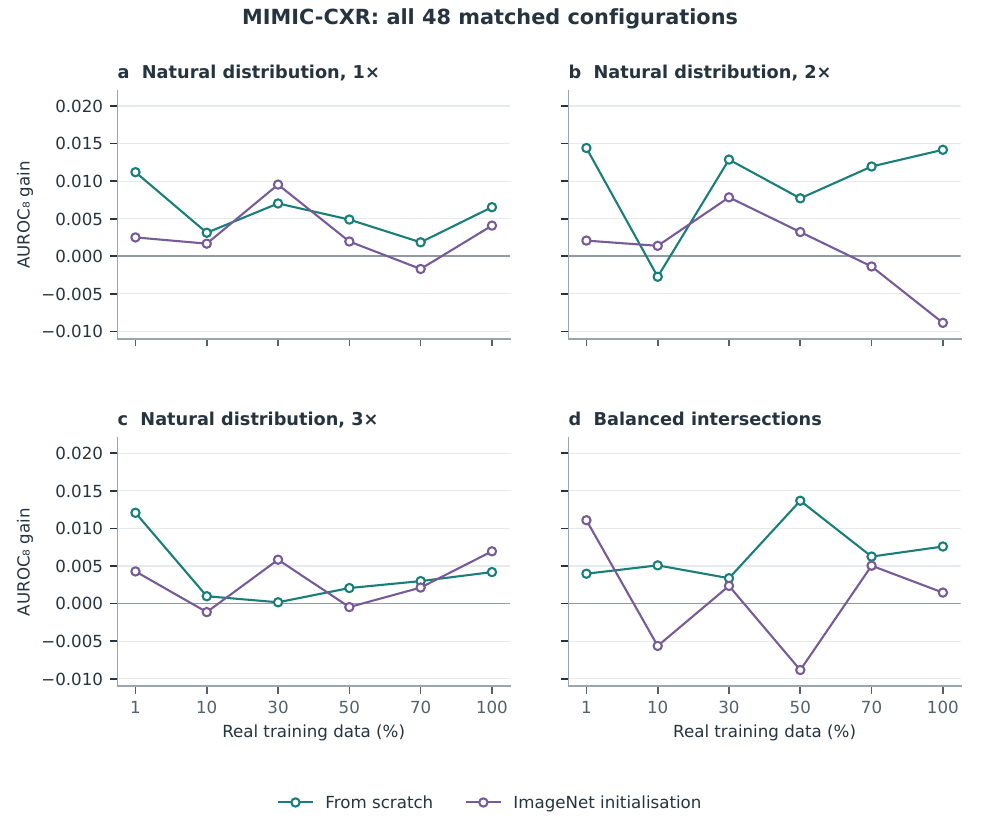}
\end{adjustbox}
\par\DisplayMark{fig:downstream-sweep-mimic/caption-top}
\caption{Complete sweep of downstream training configurations on MIMIC-CXR. Panels show the four synthetic corpus regimes; lines distinguish initialisation from scratch and from ImageNet. Every point is a CompDiff minus RoentGen-v2 AUROC$_8$ difference at one fraction of real data used for fine-tuning, giving 48 matched configurations. Positive values favour CompDiff. All scores average the same eight diagnostic labels. Axes are shared with the corresponding figure for the other test dataset. Lines connect settings in the sweep and do not imply independent replications or interpolation between evaluated fractions.}
\label{fig:downstream-sweep-mimic}
\par\DisplayMark{fig:downstream-sweep-mimic/caption-bottom}
\end{figure*}

\begin{figure*}[!htbp]
\centering
\DisplayMark{fig:downstream-sweep-chexpert/plot-top}
\begin{adjustbox}{max totalsize={\textwidth}{.67\textheight},center}
\includegraphics[width=\textwidth]{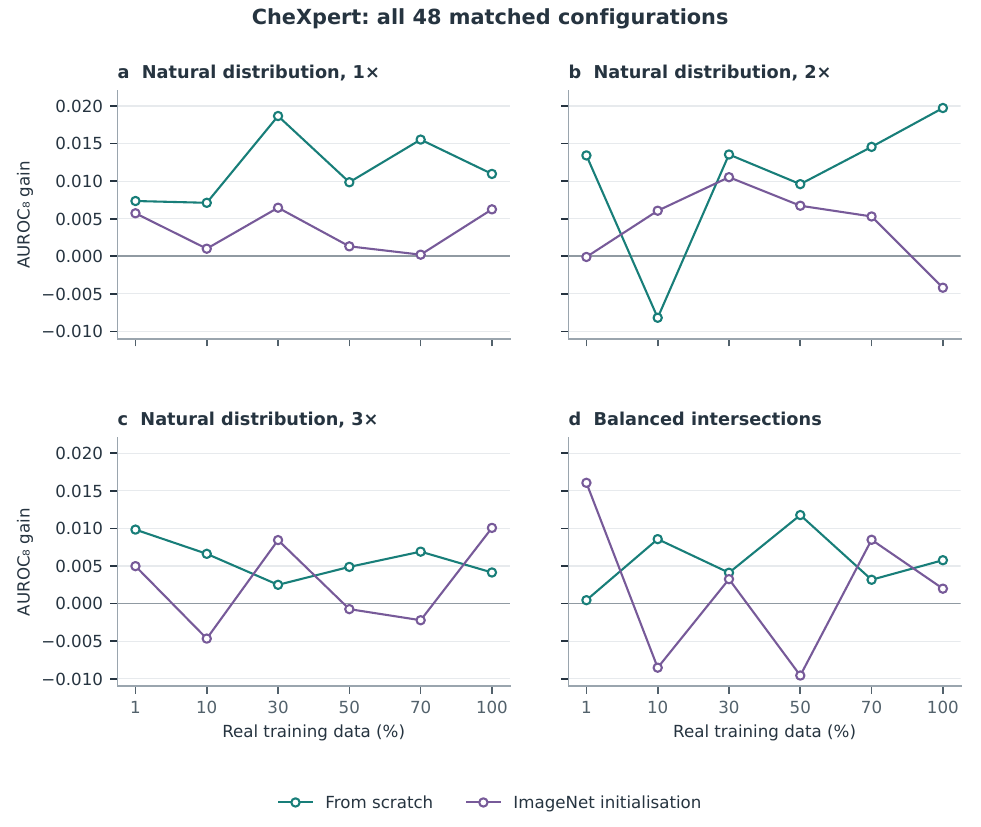}
\end{adjustbox}
\par\DisplayMark{fig:downstream-sweep-chexpert/caption-top}
\caption{Complete sweep of downstream training configurations on CheXpert. Panels show the four synthetic corpus regimes; lines distinguish initialisation from scratch and from ImageNet. Every point is a CompDiff minus RoentGen-v2 AUROC$_8$ difference at one fraction of real data used for fine-tuning, giving 48 matched configurations. Positive values favour CompDiff. All scores average the same eight diagnostic labels. Axes are shared with the corresponding figure for the other test dataset. Lines connect settings in the sweep and do not imply independent replications or interpolation between evaluated fractions.}
\label{fig:downstream-sweep-chexpert}
\par\DisplayMark{fig:downstream-sweep-chexpert/caption-bottom}
\end{figure*}

Supplementary Figures~\ref{fig:downstream-sweep-mimic} and~\ref{fig:downstream-sweep-chexpert} show every configuration in the fine-tuning sweep. The mean CompDiff gain over the six real data fractions was positive for each combination of initialisation and synthetic corpus on both datasets. Mean AUROC and area under the precision--recall curve (AUPRC) gains were positive for all eight labels on both datasets (Supplementary Table~\ref{tab:downstream-labels}). The exploratory sensitivity comparison at 95\% specificity also retained positive mean gains for all eight labels, supporting the direction of the comparison at 90\% specificity.

\FloatBarrier
\begin{table}[!htbp]
\centering
\caption{Diagnostic gains across the complete fine-tuning sweep.}
\label{tab:downstream-labels}
\begingroup\small
\setlength{\tabcolsep}{4pt}
\begin{adjustbox}{max width=\textwidth}
\begin{tabular}{llrrrr}
\toprule
Test set & Label & AUROC gain & AUPRC gain & Sens. at 90\% & Sens. at 95\% \\
\midrule
MIMIC-CXR & Atelectasis & $+0.0020$ & $+0.0042$ & $+0.57$ & $+0.11$ \\
MIMIC-CXR & Cardiomegaly & $+0.0010$ & $+0.0091$ & $+0.22$ & $+0.32$ \\
MIMIC-CXR & Consolidation & $+0.0071$ & $+0.0037$ & $+1.95$ & $+2.12$ \\
MIMIC-CXR & Oedema & $+0.0043$ & $+0.0029$ & $+1.54$ & $+1.26$ \\
MIMIC-CXR & Pleural Effusion & $+0.0018$ & $+0.0074$ & $+0.68$ & $+1.02$ \\
MIMIC-CXR & Pneumonia & $+0.0076$ & $+0.0128$ & $+1.34$ & $+0.80$ \\
MIMIC-CXR & Pneumothorax & $+0.0075$ & $+0.0196$ & $+3.13$ & $+3.03$ \\
MIMIC-CXR & No Finding & $+0.0023$ & $+0.0024$ & $+0.73$ & $+0.75$ \\
CheXpert & Atelectasis & $+0.0050$ & $+0.0054$ & $+1.12$ & $+0.63$ \\
CheXpert & Cardiomegaly & $+0.0022$ & $+0.0079$ & $+0.82$ & $+0.96$ \\
CheXpert & Consolidation & $+0.0068$ & $+0.0039$ & $+0.75$ & $+0.35$ \\
CheXpert & Oedema & $+0.0037$ & $+0.0034$ & $+0.89$ & $+0.54$ \\
CheXpert & Pleural Effusion & $+0.0018$ & $+0.0063$ & $+0.88$ & $+1.23$ \\
CheXpert & Pneumonia & $+0.0082$ & $+0.0052$ & $+1.19$ & $+1.18$ \\
CheXpert & Pneumothorax & $+0.0140$ & $+0.0220$ & $+2.85$ & $+2.68$ \\
CheXpert & No Finding & $+0.0039$ & $+0.0085$ & $+1.06$ & $+1.13$ \\
\bottomrule
\end{tabular}
\end{adjustbox}
\par\smallskip
\begin{minipage}{\textwidth}\footnotesize Values are mean CompDiff minus RoentGen-v2 differences across all 48 matched configurations. Sensitivity differences are in percentage points, at the stated specificity. Sensitivity is the maximum true positive rate at a false positive rate no greater than 0.10 or 0.05 on each test ROC. The thresholds are selected separately on the test curves for this exploratory comparison; they are not validated clinical operating thresholds. Configurations share training data and test cohorts, so these descriptive averages do not represent 48 independent replications.\end{minipage}
\endgroup
\end{table}

\FloatBarrier

\paragraph{Subgroup coverage and sparse labels.}
All ten sex, race and age groups had positive median AUROC$_8$ gains separately on each dataset (Supplementary Table~\ref{tab:downstream-groups}). Within sex, race and age, the subgroup with the lowest baseline AUROC improved in 40, 43 and 33 of 48 MIMIC-CXR configurations, respectively, and in 41, 41 and 36 of 48 CheXpert configurations. These counts describe consistency across the fixed sweep. They neither imply improvement in every configuration nor establish that disparities between groups narrowed.

Three MIMIC-CXR groups included a diagnostic label with fewer than ten positive images: Asian, Hispanic and age 80+. Requiring at least ten positives for every label excludes those groups; the median gain remains positive for all seven remaining MIMIC-CXR groups and all ten CheXpert groups. Patient resampling shows the uncertainty in the three sparse groups (Supplementary Table~\ref{tab:downstream-uncertainty}). In the Hispanic group, the least represented label occurred in only two patients; 137 of 1,000 resamples lacked the coverage needed to compute all eight AUROCs. The reported percentile ranges are therefore conditional on valid label coverage and fixed trained models. These results support the descriptive subgroup pattern while identifying where additional patients are needed for precise estimation.

\begin{table}[!htbp]
\centering
\caption{Subgroup sample sizes and gains across training configurations.}
\label{tab:downstream-groups}
\begingroup\small
\setlength{\tabcolsep}{4pt}
\begin{adjustbox}{max width=\textwidth}
\begin{tabular}{llrrrrr}
\toprule
Test set & Group & Images & Min. positive & Median gain & Interquartile range & Positive / 48 \\
\midrule
MIMIC-CXR & Female & 3407 & 44 & $+0.0052$ & $[+0.0012, +0.0082]$ & 37 \\
MIMIC-CXR & Male & 3632 & 65 & $+0.0033$ & $[+0.0006, +0.0071]$ & 37 \\
MIMIC-CXR & Asian & 298 & 6 & $+0.0058$ & $[-0.0011, +0.0120]$ & 33 \\
MIMIC-CXR & Black & 1489 & 10 & $+0.0027$ & $[-0.0016, +0.0081]$ & 33 \\
MIMIC-CXR & Hispanic & 611 & 4 & $+0.0073$ & $[-0.0046, +0.0170]$ & 31 \\
MIMIC-CXR & White & 4641 & 71 & $+0.0027$ & $[+0.0013, +0.0078]$ & 42 \\
MIMIC-CXR & 18--40 & 1159 & 14 & $+0.0052$ & $[-0.0016, +0.0130]$ & 32 \\
MIMIC-CXR & 40--60 & 2460 & 37 & $+0.0031$ & $[+0.0004, +0.0072]$ & 37 \\
MIMIC-CXR & 60--80 & 2791 & 44 & $+0.0034$ & $[-0.0004, +0.0061]$ & 35 \\
MIMIC-CXR & 80+ & 629 & 7 & $+0.0044$ & $[-0.0022, +0.0113]$ & 33 \\
CheXpert & Female & 8009 & 337 & $+0.0063$ & $[+0.0024, +0.0107]$ & 41 \\
CheXpert & Male & 15120 & 601 & $+0.0060$ & $[+0.0013, +0.0088]$ & 38 \\
CheXpert & Asian & 3241 & 139 & $+0.0041$ & $[+0.0012, +0.0094]$ & 38 \\
CheXpert & Black & 1568 & 42 & $+0.0075$ & $[+0.0031, +0.0137]$ & 39 \\
CheXpert & Hispanic & 3151 & 128 & $+0.0080$ & $[+0.0025, +0.0113]$ & 39 \\
CheXpert & White & 15169 & 595 & $+0.0054$ & $[+0.0007, +0.0096]$ & 38 \\
CheXpert & 18--40 & 3780 & 100 & $+0.0068$ & $[+0.0009, +0.0113]$ & 38 \\
CheXpert & 40--60 & 7945 & 268 & $+0.0055$ & $[+0.0014, +0.0091]$ & 39 \\
CheXpert & 60--80 & 8957 & 382 & $+0.0067$ & $[+0.0018, +0.0103]$ & 40 \\
CheXpert & 80+ & 2428 & 116 & $+0.0055$ & $[-0.0000, +0.0108]$ & 36 \\
\bottomrule
\end{tabular}
\end{adjustbox}
\par\smallskip
\begin{minipage}{\textwidth}\footnotesize Every subgroup score averages the same eight diagnostic labels, each with at least one positive and one negative image. Min. positive is the smallest positive image count among those labels. Median and interquartile range describe the 48 configuration differences within each dataset; they are not confidence intervals. Requiring at least ten positive images for every label excludes the MIMIC-CXR Asian, Hispanic and 80+ groups; all remaining subgroup median gains remain positive.\end{minipage}
\endgroup
\end{table}

\begin{table}[!htbp]
\centering
\caption{Patient resampling in the three sparsely sampled MIMIC-CXR groups.}
\label{tab:downstream-uncertainty}
\begingroup\small
\setlength{\tabcolsep}{4pt}
\begin{adjustbox}{max width=\textwidth}
\begin{tabular}{lrrrrrr}
\toprule
Group & Images & Patients & Min. positive & Median gain & 95\% valid-draw range & Valid draws \\
\midrule
Asian & 298 & 155 & 5 & $+0.0058$ & $[+0.0002, +0.0130]$ & 994/1000 \\
Hispanic & 611 & 267 & 2 & $+0.0073$ & $[-0.0006, +0.0138]$ & 863/1000 \\
80+ & 629 & 321 & 7 & $+0.0044$ & $[-0.0009, +0.0113]$ & 999/1000 \\
\bottomrule
\end{tabular}
\end{adjustbox}
\par\smallskip
\begin{minipage}{\textwidth}\footnotesize The point estimate is the median CompDiff minus RoentGen-v2 AUROC gain across the fixed 48 configurations. Min. positive counts distinct patients with a positive image for the least represented label. Each of 1,000 draws resamples patients with replacement and retains all their images and all model predictions. Draws lacking either class for any label cannot yield the specified eight-label AUROC and are excluded. The range gives the 2.5th and 97.5th percentiles among valid draws, conditional on the fixed trained models and valid label coverage. It should not be interpreted as an unconditional 95\% confidence interval, particularly for the Hispanic group, where some labels occur in only two patients.\end{minipage}
\endgroup
\end{table}

\FloatBarrier
\FloatBarrier

\FloatBarrier
\section{Demographic tokens in the prompt: truncation statistics and probe of the encoder}\label{app:clip-sensitivity}

These analyses distinguish prompt truncation from changes within the shared text representation.

The analysis procedure is specified in Section~\ref{subsec:clip-protocol}.

\paragraph{Truncation is negligible.} Tokenising the final training prompts of all $\numAppendixK{}$ training images ($\numAppendixL{}$ unique clinical impressions) with Stable Diffusion 2.1's CLIP tokenizer, no clinical impression alone exceeds the $\numAppendixM{}$-token context limit (median $\numAppendixN{}$ tokens per unique impression, maximum $\numAppendixO{}$), and prepending the demographic prefix, which costs exactly $\numAppendixP{}$ tokens for every (age, sex, race) combination, pushes only $\numAppendixQ{}$ prompts ($<\numAppendixR{}\%$) beyond it. Displacement of clinical tokens past the context limit therefore cannot explain the FID gap in the ablation. The prefix nonetheless accounts for a mean $\numAppendixS{}\%$ (median $\numAppendixT{}\%$; $\numAppendixU{}$th percentile $\numAppendixV{}\%$) of the tokens in a training prompt, because clinical impressions are short. We therefore examined how demographic text changes the encoder's shared representation.

\paragraph{Probe of the encoder.} Two patterns are visible in Supplementary Figure~\ref{fig:clip-sensitivity}. First, the perturbation falls steeply with clinical prompt length and then plateaus (Supplementary Figure~\ref{fig:clip-sensitivity}a): a short impression of \numAppendixW{}--\numAppendixX{} words is perturbed $\numAppendixY{}\times$ as much as a long one of \numAppendixZ{}--\numAppendixAA{} words (Supplementary Figure~\ref{fig:clip-sensitivity}b), so when clinical content is scarce the demographic prefix dominates the representation that conditions the UNet. Second, this shift is not an artefact of demographic information leaking through clinical phrasing: linear probes trained on the embeddings without demographic text recover sex, race, and age only at levels close to chance (balanced accuracy $\numAppendixAB{}$, $\numAppendixAC{}$, $\numAppendixAD{}$ against chance $\numAppendixAE{}$, $\numAppendixAF{}$, $\numAppendixAG{}$). These patterns are consistent with demographic text altering the shared representation. The architectural ablations in Section~\ref{subsec:ablations} assess how separating the conditioning pathways affects image quality and demographic control.

\begin{figure}[htbp]
\centering
\DisplayMark{fig:clip-sensitivity/plot-top}
\begin{adjustbox}{max totalsize={\textwidth}{0.67\textheight},center}
\begin{minipage}{\textwidth}
\centering

\includegraphics[width=\textwidth]{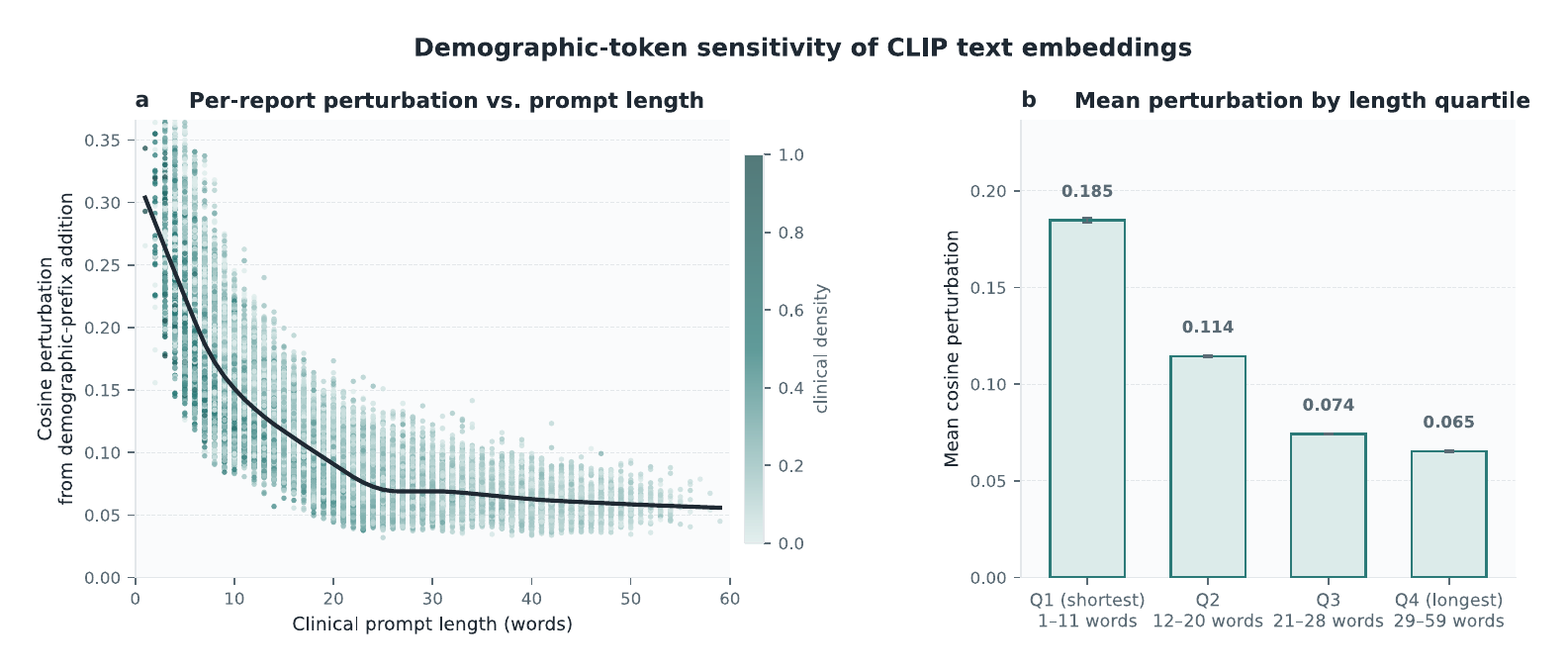}

\end{minipage}
\end{adjustbox}
\par\DisplayMark{fig:clip-sensitivity/caption-top}
\caption{CLIP's text encoder is sensitive to demographic tokens, and this sensitivity falls steeply with clinical prompt length. For each of $\numSupplementFiguresM{}$ MIMIC-CXR clinical impressions, the clinical text and the same text with each of $\numSupplementFiguresN{}$ demographic prefixes (three age descriptors $\times$ two sexes $\times$ four race labels) are passed through Stable Diffusion 2.1's frozen CLIP text encoder; we report the cosine perturbation between embeddings with and without demographic text, averaged over the $\numSupplementFiguresO{}$ demographic variants. \textbf{a}, Perturbation for each report against clinical prompt length on linear axes. Each point is one report, coloured by its clinical density (fraction of words that are recognised clinical entities); the dark line is a LOWESS smoother (span $\numSupplementFiguresP{}$). The same relationship is well described by a power law in log--log space, $y \propto L^{-\numSupplementFiguresQ{}}$ ($R^2_{\log\log}=\numSupplementFiguresR{}$). \textbf{b}, Mean perturbation within quartiles of word count with bootstrap $\numSupplementFiguresS{}\%$ CIs (\numSupplementFiguresT{} resamples); impressions in the shortest quartile are perturbed $\numSupplementFiguresU{}\times$ more than impressions in the longest quartile.}
\label{fig:clip-sensitivity}
\par\DisplayMark{fig:clip-sensitivity/caption-bottom}
\end{figure}

\FloatBarrier

\clearpage